\documentclass[10pt,twocolumn,letterpaper]{article}

\usepackage[pagenumbers]{cvpr} 

\usepackage[dvipsnames]{xcolor}

\usepackage{color, colortbl}
\usepackage{multirow}
\definecolor{Gray}{gray}{0.9}

\definecolor{cvprblue}{rgb}{0.21,0.49,0.74}
\usepackage[pagebackref,breaklinks,colorlinks,citecolor=cvprblue]{hyperref}

\newcommand{\xx}{\mathbf{x}}
\newcommand{\ww}{\mathbf{w}}

\newcommand{\yy}{\mathbf{y}}

\newcommand{\vv}{\mathbf{v}}

\newcommand{\ttt}{\mathbf{t}}
\newcommand{\logit}{\mathbf{l}}
\newcommand{\appensecref}[1]{\textcolor{red}{Supp.~Sec.~\ref{#1}}}
\newcommand{\appencref}[1]{\textcolor{red}{Supp.~\cref{#1}}}
\newcommand{\rms}{\setlength{\thinmuskip}{0mu}\setlength{\medmuskip}{0mu}\setlength{\thickmuskip}{0mu}} 

\def\vrho{{\boldsymbol{\rho}}}
\def\vlambda{{\boldsymbol{\lambda}}}
\newcommand{\real}{\mathbb{R}}
\newcommand{\N}{\mathbb{N}}
\newcommand{\Loss}{\mathcal{L}}

\newcommand{\ent}[1]{\mathcal{H}(#1)}

\def\real{{\mathbb{R}}}

\def\blambda{\boldsymbol{\lambda}}
\def\brho{\boldsymbol{\rho}}
\def\slambda{\lambda^\star}
\def\sblambda{\boldsymbol{\lambda}^\star}

\usepackage{graphicx}
\usepackage{xcolor} 
\usepackage{color} 

\usepackage[euler]{textgreek}
\usepackage{arydshln}
\usepackage{marvosym}

\def\paperID{} 
\def\confName{}
\def\confYear{}

\title{A Closer Look at the Few-Shot Adaptation of Large Vision-Language Models}

\author{Julio Silva-Rodríguez\textsuperscript{\Letter} \qquad Sina Hajimiri \qquad Ismail Ben Ayed \qquad Jose Dolz \\
ÉTS Montreal \\
\Letter {\tt \small julio-jose.silva-rodriguez@etsmtl.ca}
}

\begin{document}

\makeatletter
\let\@oldmaketitle\@maketitle
\renewcommand{\@maketitle}{\@oldmaketitle
	\begin{center}
    \begin{tabular}{ccccc}

        \hspace{-2em} \scriptsize{ (a) \textbf{CLIP-Adapter} \cite{gao2021clip} } & \hspace{-1.5em} \scriptsize{ (b) \textbf{TIP-Adapter(f)} \cite{zhang2021tip} } & \hspace{-1.5em} \scriptsize{ (c) \textbf{TaskRes} \cite{yu2023task} } & \hspace{-1.5em} \scriptsize{ (d) \textbf{CLAP} (\textit{Ours})} & \\
        
         \hspace{-2em} \includegraphics[width=.25\linewidth]{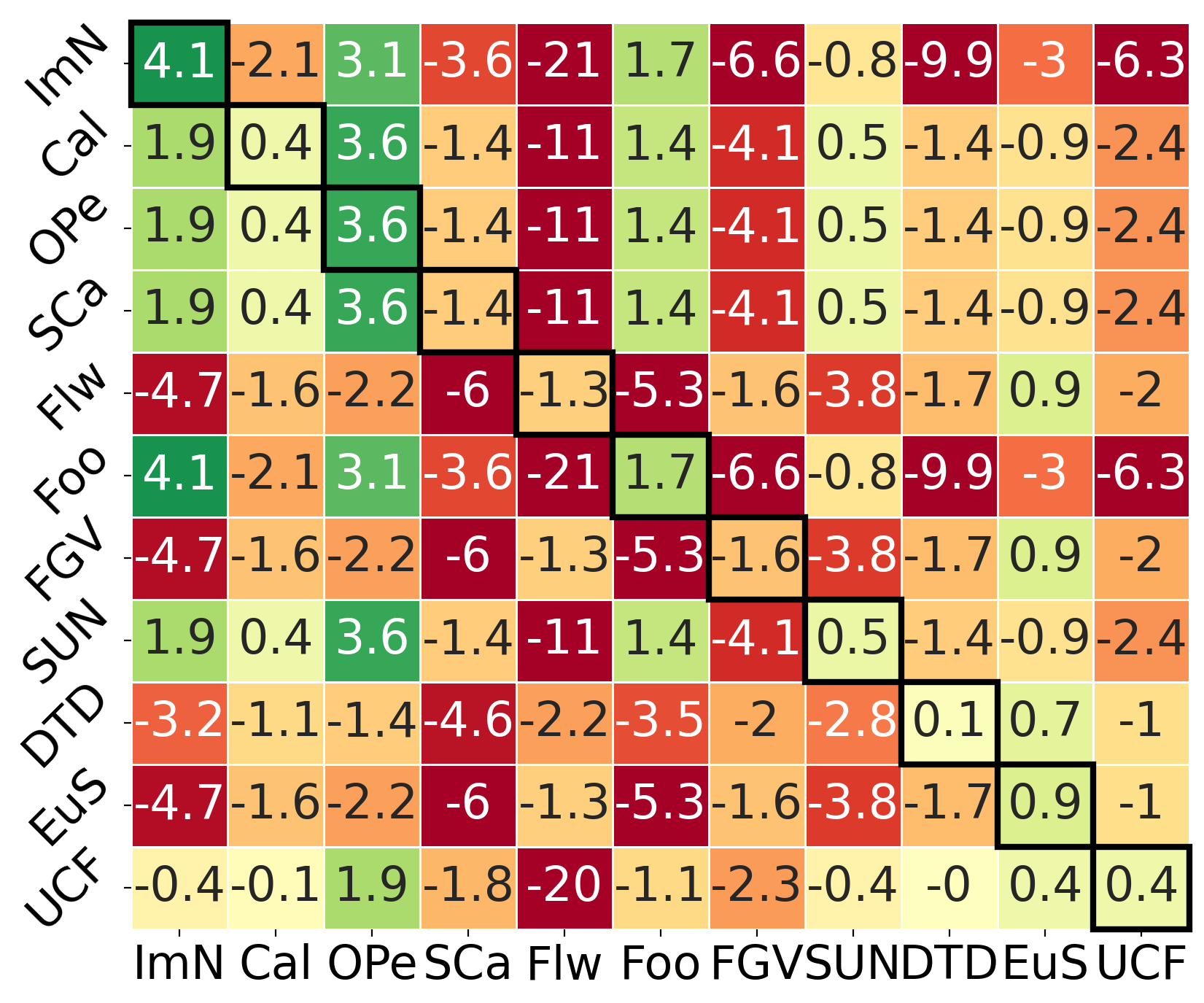} &
         \hspace{-1.5em} \includegraphics[width=.25\linewidth]{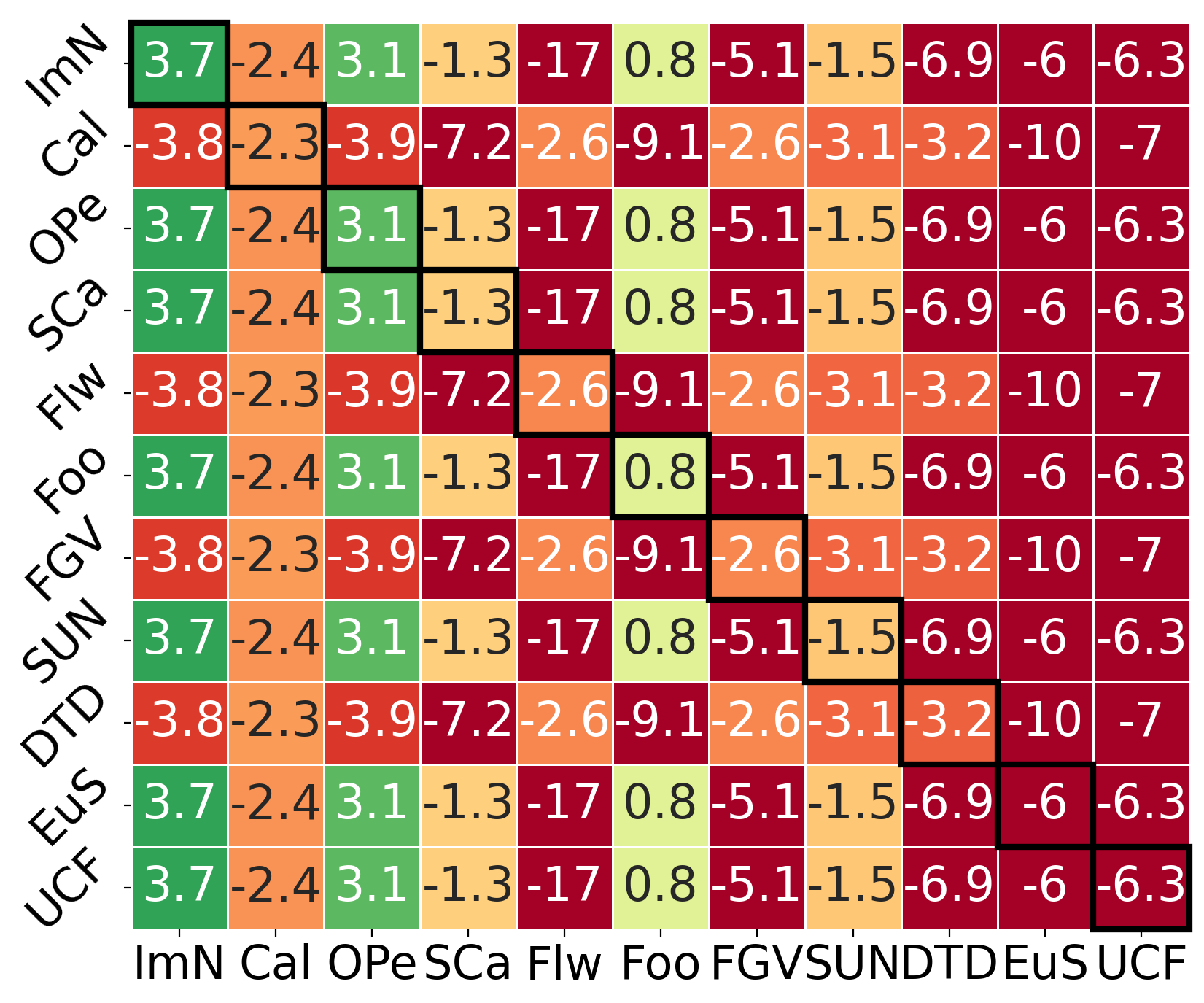} &
         \hspace{-1.5em} \includegraphics[width=.25\linewidth]{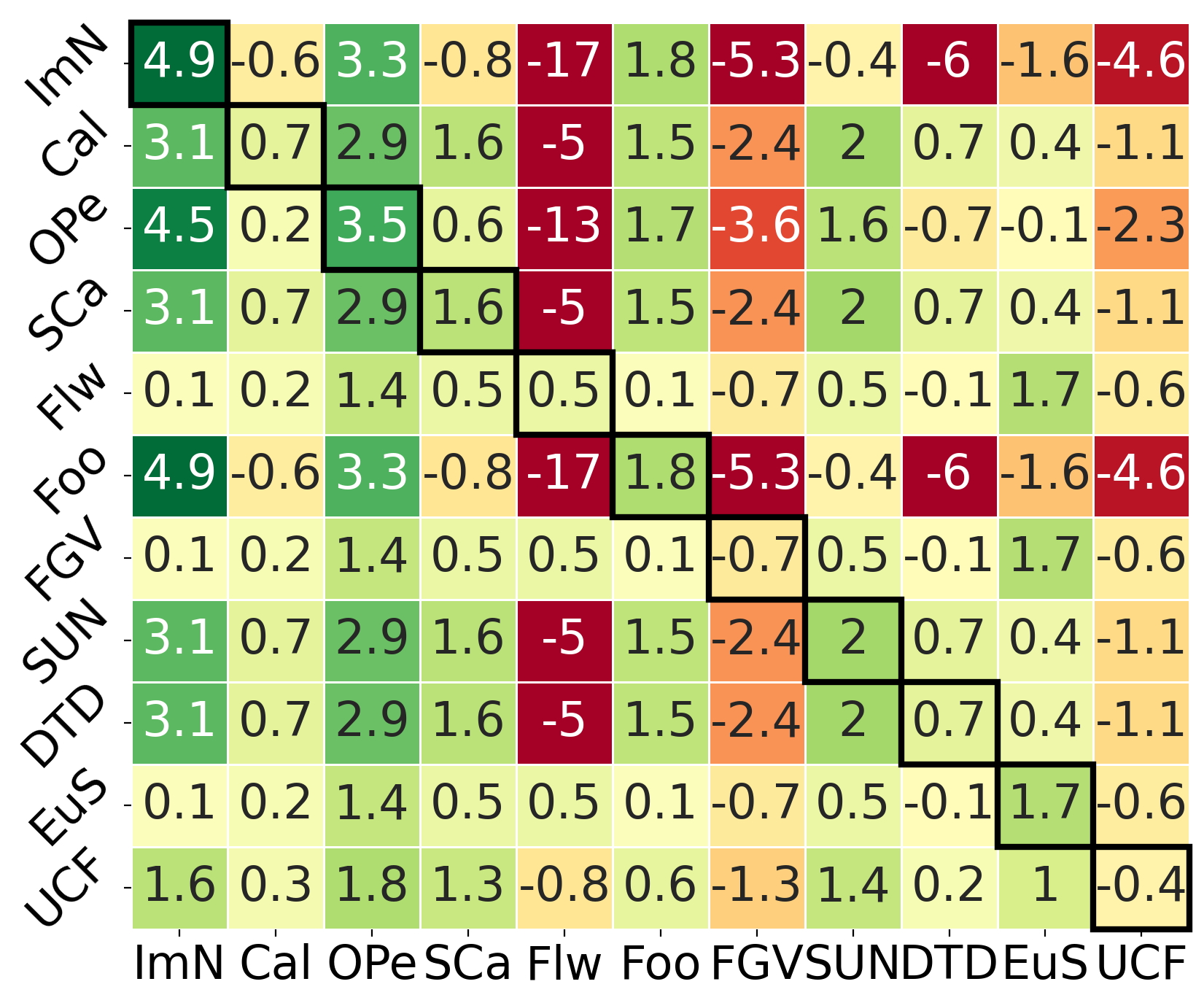} &
         \hspace{-1.5em} \includegraphics[width=.25\linewidth]{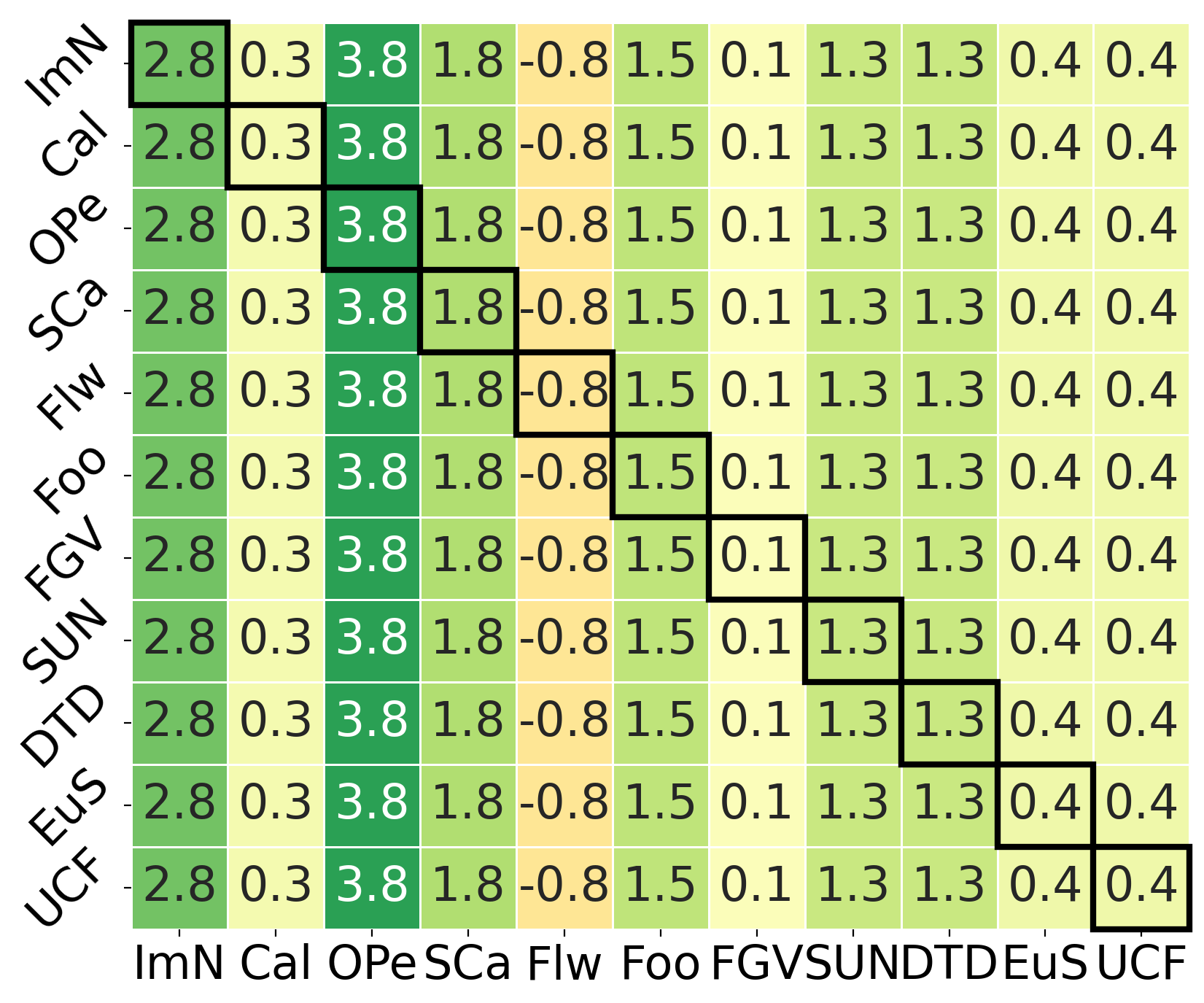} &
         \hspace{-1.2em} \multirow{1}{*}[8em]{\includegraphics[width=.03\linewidth]{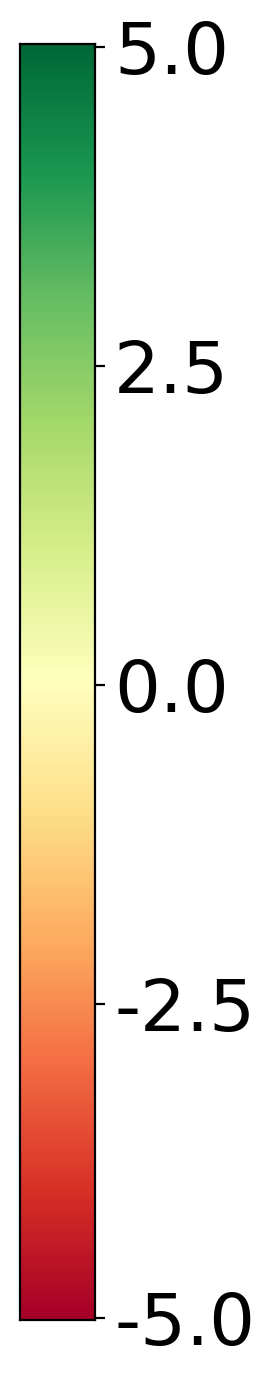}}
         
        \end{tabular}
        \vspace{-2mm}
        \captionsetup{type=figure}
        \captionof{figure}{\textbf{Pitfalls of few-shot adapters due to the absence of a \textit{model selection} strategy.} The cross-shift model selection matrices $(i,j)$ depict the relative improvement w.r.t. a zero-shot initialized Linear Probing when using the optimal hyperparameters for the \mbox{dataset $i$ (\textit{rows})}, for adapting in another task $j$ (\textit{columns}), for each SoTA method (\textit{first three plots}) and our approach (\textit{last plot}).}
        \label{fig:cross-shift}
\end{center}%
}
\makeatother

\maketitle

\thispagestyle{empty}


\begin{abstract}

Efficient transfer learning (ETL) is receiving increasing attention to adapt large pre-trained language-vision models on downstream tasks with a few labeled samples. While significant progress has been made, we reveal that state-of-the-art ETL approaches exhibit strong performance only in narrowly-defined experimental setups, and with a careful adjustment of hyperparameters based on a large corpus of labeled samples. In particular, we make two interesting, and surprising empirical observations. First, to outperform a simple Linear Probing baseline, these methods require to optimize their hyper-parameters on each target task. And second, they typically underperform --sometimes dramatically-- standard zero-shot predictions in the presence of distributional drifts. Motivated by the unrealistic assumptions made in the existing literature, i.e., access to a large validation set and case-specific grid-search for optimal hyperparameters, we propose a novel approach that meets the requirements of real-world scenarios. More concretely, we introduce a CLass-Adaptive linear Probe (CLAP) objective, whose balancing term is optimized via an adaptation of the general Augmented Lagrangian method tailored to this context. We comprehensively evaluate CLAP on a broad span of datasets and scenarios, demonstrating that it consistently outperforms SoTA approaches, while yet being a much more efficient alternative. Code available at \url{https://github.com/jusiro/CLAP} \ .

\end{abstract}



\section{Introduction}
\label{main:section_introduction}

Large vision-language models (VLMs), such as CLIP \cite{radford2021learning}, are reshaping the research landscape with their unprecedented performance. These models undergo training on an extensive dataset consisting of hundreds of millions of image-text pairs, which are leveraged via contrastive learning \cite{radford2021learning}. Once trained, VLMs offer a remarkable zero-shot performance on a wide span of visual recognition problems thanks to the rich learned representations \cite{radford2021learning,Menon2023}. Nevertheless, the extensive hardware and data-driven resources that such training demands \cite{Bommasani2021FoundationModels} suggest that these models can only be trained on singular occasions. Furthermore, the large scale of these networks poses important challenges when it comes to adjusting their parameters on small downstream tasks that involve only a few labeled samples, making the full fine-tuning of the entire model impractical.

An emerging alternative to alleviate this issue consists in fine-tuning VLMs by adding a small set of learnable parameters, whose values are optimized during the adaptation step \cite{zhou2022coop,jia2022visual,zhou2022cocoop,gao2021clip,zhang2021tip}. These tunable weights can be introduced in the input space as visual \cite{jia2022visual} or text prompts \cite{zhou2022coop,zhou2022cocoop}, or added in the form of adapters across the network \cite{gao2021clip,zhang2021tip,yu2023task}. While both families of approaches fit within the Efficient Transfer Learning (ETL) literature, prompt learning still requires backpropagating the gradients through the entire network. Thus, besides introducing a burden on resource reuse, these methods preclude \textit{black-box} adaptation, introducing a potential concern about leaking the source data, which is paramount in privacy-oriented applications. In contrast, strategies based on adapters only need gradients on the extra set of parameters, typically in the last layer, avoiding costly fine-tuning processes and data leakage, yet yielding state-of-the-art performance \cite{yu2023task,lin2023crossmodal}.

Despite the progress observed in adapter-based methods for fine-tuning VLMs under the few-shot learning paradigm, improving the performance on the target task while preserving their generalization capabilities remains still a challenge \cite{zhou2022coop}. We argue that this is likely due to the severe overfitting to the support set samples employed during few-shot adaptation, which significantly deviates the updated class prototypes from the zero-shot prototypes initially provided by the pre-trained model. In fact, popular adapter-based ETL strategies, such as CLIP-Adapter \cite{gao2021clip} and TIP-Adapter \cite{zhang2021tip}, carefully adjust the model-specific hyperparameters, in conjunction with other key hyperparameters related to the learning scheduler, to control the trade-off between initial zero-shot inference and the integration of new information from the support set. Furthermore, recent evidence \cite{lin2023crossmodal} suggests that these works apparently use the large-scale test set to adjust their hyperparameters.

A significant limitation becomes evident in that these hyperparameters, when optimized for one specific task, do not exhibit strong generalizability to other tasks, as illustrated in \cref{fig:cross-shift}. Indeed, state-of-the-art (SoTA) methods \textbf{struggle to find a homogeneous configuration that outperforms a simple well-initialized Linear Probing (LP) adaptation}. Notably, in a realistic adaptation scenario (\cref{fig:cross-shift}), we can observe dramatic performance degradations, up to 21\%, compared to this simple baseline. These practices virtually bias the model selection process, as assuming access to a significantly larger set of labeled samples, and adjusting the model hyperparameters in a case-specific manner, is not only \textit{unrealistic} but also \textit{impractical} (grid-search must be done for each case). Thus, we argue that if an ETL method's model selection strategy is not solely based on the support samples, the method is \textit{incomplete}, and impractical for real-world few-shot adaptation problems.

In this work, we seek to redirect the efforts on few-shot ETL to a more strict, but realistic scenario, in which only the support samples are accessible during training. The absence of an evaluation subset urges novel adapters to include a model selection strategy, robust across a large spectrum of tasks. Interestingly, we empirically observed that a carefully designed Linear Probing (\mbox{ZS-LP}), whose weights are initialized with the zero-shot prototypes from CLIP, is a strong baseline that outperforms more convoluted ETL solutions. 
To further improve the baseline \mbox{ZS-LP} and optimize the trade-off between initial zero-shot representations and updated class prototypes on novel tasks, we propose penalizing large deviations from the original zero-shot prototypes during adaptation. The resulting learning objective, however, presents two major issues. First, the penalty included to control the deviation between original and updated prototypes is a scalar value, uniform across all classes, which can detrimentally affect the model's performance in the presence of harder-to-learn classes. Second, the penalty balancing weight must be set using a validation set, which juxtaposes with our \textit{validation-free} scenario. To address these limitations, we propose CLass-Adaptive linear Probe (CLAP), which is based on an Augmented Lagrangian Multiplier approach. We can summarize our contributions as:

\begin{itemize}
    \item We empirically observe that SoTA few-shot ETL adapters require careful adjustment of a set of key hyperparameters for each task, which is unrealistic and impractical in real-world settings. Surprisingly, if a fixed configuration is adopted across tasks, these methods are likely to substantially underperform a simple Linear Probing strategy initialized with the zero-shot prototypes from CLIP.

    \item We propose a principled solution to tackle the trade-off between original and updated class prototypes in Linear Probing, which integrates a penalty term to penalize large deviations from zero-shot prototypes. To address the underlying challenges from the resulting constrained optimization problem, we present a modified Augmented Lagrangian Multiplier (ALM) method. This alleviates the need of having to fine-tune the penalty balancing weight, which is learned in the outer iteration of the optimization process. In order to adapt ALM to the presented scenario, two critical choices were made: \textit{i)} Leveraging class prototypes, as well as data augmentation, motivate the use of class-wise multipliers, instead of sample and class-wise multipliers as in the original ALM; \textit{ii)} In the presented scenario, there is no access to a validation set, and the only feedback available is from the support samples. Hence, we only perform one outer-step update, which can avoid potential overfitting on the support set. 
    
    \item We provide extensive experiments to assess the performance of CLAP in the proposed scenario, including few-shot adaptation on 11 popular classification benchmarks, domain generalization, comparison to full fine-tuning methods, and ablation studies to validate our choices. As shown in \cref{fig:cross-shift} and in the experimental section, CLAP delivers consistent performance across different tasks with a homogeneous configuration, and largely outperforms SoTA ETL approaches in all scenarios. 
    
\end{itemize}


\section{Related work}
\label{main:section_rw}

\noindent \textbf{Vision-language pre-trained models.} The field of machine learning is in the midst of a paradigm shift with the emerging rise of vision-language models (VLMs). These networks have gained increasing popularity, especially fueled by the significant improvements achieved in computer vision and natural language processing tasks \cite{radford2021learning,brown2020language,jia2021scaling,zhai2022lit}. The prevailing learning paradigm consists of a dual stream of data, which separately encodes images and their text counterparts, leveraging contrastive learning at a large scale to bridge image and text representations in the latent space. Particularly, models such as CLIP \cite{radford2021learning} and ALIGN \cite{jia2021scaling} have successfully mitigated the distribution discrepancy between text and images, and have shown tremendous zero-shot capabilities on visual recognition tasks, primarily in the context of classification.

\noindent \textbf{Full fine-tuning.} A body of work proposes fine-tuning the entire VLMs to adapt to a specific task \cite{LPFT,FLYP,WiSE}. This strategy, however, presents several drawbacks. Concretely, fine-tuning increases the complexity of the model being optimized, makes the optimization process more time-consuming compared to ETL methods, and requires access to the backbone weights, which does not allow a black-box adaptation. Furthermore, full fine-tuning methods typically tend to overfit when trained on small datasets, requiring a large corpus of labeled data for the target task, which may be impractical in many real-world scenarios. 

\noindent \textbf{Efficient transfer leaning} attempts to address these issues by updating a small set of learnable parameters and leveraging a limited amount of annotated samples. Current ETL literature can be categorized into \textit{Prompt Learning} \cite{zhou2022coop,zhou2022cocoop,zhu2023prompt,yao2023visual,khattak2023maple,xing2023dual} and \textit{Adapter-based} \cite{zhang2021tip,gao2021clip,yu2023task} approaches. \textit{Prompt Learning} represents a recent advancement in the realm of natural language processing \cite{lester2021power,zhong2021factual}, which has been recently adopted with success in VLMs. In these methods, only the text tokens provided to the model are optimized. Nevertheless, these techniques require long training steps due to backpropagating the gradient over the entire network, which juxtaposes with the spirit of \textit{efficient} adaptation. Furthermore, black-box adaptation is also not possible in prompt learning. \textit{Adapter-based} methods, in contrast, offer a much lighter alternative as only a small subset of parameters, typically at the latest layers, are adjusted. For example, CLIP-Adapter \cite{gao2021clip} integrates a two-layer MLP to modify the visual embedding generated by CLIP. In TIP-Adapter \cite{zhang2021tip}, the visual prototypes obtained from the few-shot support samples are leveraged to compute the similarity with the visual embedding of the test image, which is later used to modify the CLIP visual embedding.

\section{Preliminaries}
\label{main:section_preliminaries}

\subsection{Contrastive vision-language pre-training}
\label{main:subsection_vlp}

Large-scale VLMs, such as CLIP \cite{radford2021learning}, are trained on large heterogeneous datasets, encouraging image and text representations to correlate in a joint embedding space. Formally, CLIP comprises a vision encoder, $f_\theta(\cdot)$, and a text encoder, $f_\phi(\cdot)$, each aiming at learning a rich representation of their data points. These points are projected in an $\ell_{2}$-normalized shared embedding space, yielding the corresponding visual ${\vv}$ and text ${\ttt}$ embeddings. The whole network is optimized to maximize the similarity between the projected embeddings of paired images and texts, using a contrastive loss.

\subsection{Transferability}
\label{main:subsection_transferability}

\paragraph{Zero-shot inference.} For a particular downstream image classification task, CLIP-based models are able to provide predictions based on the similarity between category prompts, \ie, text descriptions of target classes, and testing images. Given a set of $C$ categories, and an ensemble of $N$ text prompts for each one, $\{\{T_{n,c}\}_{n=1}^{N}\}_{c=1}^{C}$, a common practice is to obtain a zero-shot prototype for each target category by computing the center of the $\ell_2$-normalized text embeddings for each class, $\ttt_{c}=\frac{1}{N}\sum_{n=1}^Nf_\phi(T_{n,c})$. Thus, for a given query image $\xx$, the zero-shot prediction is obtained from the softmax cosine similarity between the vision embedding $\vv=f_\theta(\xx)$, and category prototypes $\ttt_{c}$: 

\begin{align}
\label{eq:zs_prediction}
    \phantom{,}\hat{y}_{c}
    = \frac
    {\exp( \vv \cdot \ttt_{c}^\top / \tau)}
    {\sum_{i=1}^{C} \exp( \vv \cdot \ttt_i^\top / \tau)},
\end{align}

\noindent where $\tau$ is a temperature parameter learned during the pre-training stage, and $\vv \cdot \ttt^\top$ the dot product operator, which is equivalent to cosine similarity, as vectors are $\ell_2$-normalized.

\paragraph{Few-shot learning.} This scenario assumes access to limited supervisory information on the downstream tasks, in the form of a few examples for each target category, so-called shots. Formally, we denote a support set, $\mathcal{S}=\{(\xx^{(m)},\yy^{(m)})\}_{m=1}^{M=K\times C}$, composed of $K$ images for each target category, such that $K$ takes a small value, \eg, $K \in \{1, 2, 4, 8, 16\}$, and where $\yy \in \{0,1\}^C$ is the corresponding one-hot label for a given image $\xx$. The objective is to adapt the pre-trained model using this limited support set.

\subsection{Efficient transfer learning with adapters}
\label{main:subsection_etl}

In their general form, ETL methods based on adapters learn a set of transformations over the pre-trained features ($\vv',\ttt'=f_{\psi}(\vv,\ttt)$), parameterized by the so-called adapter $\psi$, which produces softmax scores for the new tasks following \cref{eq:zs_prediction}. The adapter $\psi$ can be optimized by minimizing the popular cross-entropy (CE) loss, \mbox{$\ent{\yy, \hat{\yy}} = - \sum_{c=1}^C {y}_c \log \hat{y}_c$}, over the support set samples:

\begin{equation}
\label{eq:ceLoss}
    \phantom{.}\min_\psi \; \frac{1}{M} \sum\limits_{m=1}^{M}
    \ent{\yy^{(m)}, \hat{\yy}^{(m)}}.
\end{equation}

\subsection{Pitfalls of existing few-shot ETL methods}
\label{main:subsection_pitfalls_adapters}

Recent ETL methods tailored to VLMs focus on enhancing the supervision provided by the support samples with priors learned by the VLMs at the task at hand. The pre-trained model gathers robust knowledge and is able to align visual and textual concepts. Retaining this prior knowledge can therefore produce more robust adapters, able to generalize beyond the specific bias introduced in the few support samples, to more general concepts. In this context, the zero-shot prototypes from CLIP act as a proxy to initialize the learning procedure into a reliable region. For instance, CLIP-Adapter \cite{gao2021clip} maintains the zero-shot prototypes based inference as in \cref{eq:zs_prediction}, but includes a residual multi-layered perceptron to modify visual features, such as $\mathbf{v'}=\mathbf{v}+\alpha_\mathrm{r}f_{\psi}(\mathbf{v})$. TIP-Adapter \cite{zhang2021tip} includes an additional complexity layer, by combining the similarity of the zero-shot prototypes with a weighted similarity to the support samples, $f_{\psi}(\cdot, \beta)$, controlled by the hyperparameter $\beta$, such that the predicted logits are $\logit_{c} = \alpha_{\mathrm{tipA}} f_{\psi}(\vv, \beta) + \vv \cdot \ttt_{c}^\top / \tau$. Finally, TaskRes \cite{yu2023task} learns a modification of the initial zero-shot prototypes, $\ww_{TR}$, using the support samples. The divergence between the initial and final prototypes is controlled by a residual ratio: ${\mathbf t'} = {\mathbf t} + \alpha_\mathrm{TR} \ww_{TR}$. Nevertheless, these methods lack a \textit{model selection} strategy to set these hyperparameters (See \appensecref{supp:section_adapters} for details).

\section{Proposed approach}
\label{main:section_approach}

\subsection{Revisiting Linear Probing}
\label{main:subsection_lp}

The most straightforward approach used to adapt VLMs is Linear Probing 
\cite{radford2021learning}, which refers to fitting a multiclass logistic regression linear classifier on top of the pre-trained features. Formally, the objective is to learn a set of class-wise prototypes, $\ww_c$, to provide softmax class scores for a given visual embedding $\vv$:

\begin{equation}
\label{eq:lp_pred}
    \phantom{.}\hat{y}_{c} = \frac
    {\exp(\vv \cdot \ww_{c}^\top / \tau)}
    {\sum_{i=1}^{C} \exp(\vv \cdot \ww_i^\top / \tau)}.
\end{equation}

The $\mathbf{w_c}$ prototypes can be trained to minimize the cross-entropy loss on the support samples, as in \cref{eq:ceLoss}, using standard SGD. Besides, a common practice in 
ETL is to regularize the trained weights \cite{radford2021learning,yu2023task,lin2023crossmodal} by minimizing its $\ell_{2}$-norm with an additional term, weighted by an empirically-optimized non-negative balancing term $\lambda_{wd}$. Despite its limited performance shown for few-shot adaptation \cite{radford2021learning,gao2021clip}, we believe that this requires further exploration, as LP is a lightweight adaptation strategy, especially convenient due to its convexity during optimization. In this work, we present an updated view of Linear Probing. First, the class weights are initialized using the CLIP zero-shot prototypes, as SoTA ETL methods do \cite{gao2021clip,zhang2021tip,yu2023task}. Second, we replace the weight decay in the loss function and explicitly perform an $\ell_{2}$-normalization of the prototypes after each update, to exactly meet the pre-training scenario during adaptation, inspired by \cite{FLYP}. Similarly, cosine similarity is also scaled with CLIP's pre-trained temperature $\tau$. Last, we incorporate data augmentation, usually not included in LP. 
We refer to this updated Linear Probing version for vision-language models as \mbox{ZS-LP}\footnote{Although the recent work in \cite{lin2023crossmodal} explores some of these LP improvements, they still resort to a weight-decay regularization of the LP parameters, whose optimum relative weight is found in a few-shot validation set.}. Interestingly, \mbox{ZS-LP} serves as a strong baseline (see \cref{table_main_few_shot_results}), which does not require adjusting specific hyperparameters per task. 

\subsection{Constrained Linear Probing}
\label{main:subsection_const_lp}

Albeit a well-initialized Linear Probing offers a strong baseline for efficient transfer learning, the updated prototypes might deviate from the initial regions offering strong generalization. This is especially the case in the few-shot setting, where the few provided support samples might be under-representative and contain specific biases that produce spurious correlations, hence harming the generalization after adaptation \cite{taori2020measuring,KaiyangDG}. Thus, to retain the strong basis provided by the VLM model, and avoid prototype degradation, we resort to a constrained formulation of the loss in \cref{eq:ceLoss}. 

\paragraph{Retaining prior knowledge.} A direct form to avoid prototype degradation from zero-shot points is to constrain the cross-entropy minimization to enforce the resulting prototypes to remain close to the initial solution (\ie, initial set of prototypes $\mathcal{T}=[\ttt_1, \dots, \ttt_c]$). Specifically, this constrained optimization problem can be defined as follows:

\begin{equation}
\begin{gathered}
\label{eq:constraint_optim}   
    \min_{\mathcal{W}} \quad \frac{1}{M} \sum\limits_{m=1}^{M} \ent{\yy^{(m)},\hat{\yy}^{(m)}} \\
    \text{s.t.} \quad \ww_c=\ttt_c \quad                \forall c \in \{1, \dots ,C\},
\end{gathered}
\end{equation}

\noindent with $\mathcal{W}=[\ww_1,...,\ww_C]$ the set of learnable class prototypes. We can approximate the minimum of the constrained problem in \cref{eq:constraint_optim} by a penalty-based optimization approach, transforming the above formulation into an unconstrained problem, and using an $\ell_{2}$-penalty between the class prototypes and the set of zero-shot anchors:
\begin{equation}
\label{eq:loss_penalty_single}
    \phantom{.}\min_{\mathcal{W}}  \quad 
    \sum\limits_{m=1}^{M} \ent{\yy^{(m)},\hat{\yy}^{(m)}} +
    \lambda \sum_{m=1}^M \sum_{c=1}^C \; ||\ttt_c - \ww_c^{(m)}||_{2}^{2},
\end{equation}

\noindent where $\lambda \in \real_{+}$ is a scalar weight controlling the contribution of the corresponding penalty. Note that $\ww_c^{(m)}$ is the optimal class prototype for the support sample $m$ that minimizes the left term. For clarity in the presentation, we have omitted the normalization by the cardinality of each set.

\paragraph{Sample and class-specific constraints.} The associated constrained problem in \cref{eq:constraint_optim} is approximated by an unconstrained formulation, which uses a single uniform penalty without considering individual data samples or classes. Certainly, all samples and categories within a given dataset may indeed present different intrinsic learning challenges. Thus, the problem in \cref{eq:loss_penalty_single} is not solved accurately. A better alternative would consist in integrating multiple penalty weights $\lambda$, one for each sample and class, producing a set of penalty weights $\mathbf{\Lambda} \in \real_{+}^{M \times C}$. The resulting optimization problem can then be defined as:

\begin{equation}
\label{eq:loss_lagrangian}
\phantom{.}\min_{\mathcal{W}}  \quad 
\sum\limits_{m=1}^{M} \ent{\yy^{(m)},\hat{\yy}^{(m)}} + 
\sum_{m=1}^M \sum_{c=1}^C \mathbf{\Lambda}_{mc} \; ||\ttt_c - \ww_{c}^{(m)}||_{2}^{2}.
\end{equation}

Now, from an optimization standpoint, if we suppose that there exists an optimal set of class-prototypes $\mathcal{W}^*$ for the problem presented in \cref{eq:constraint_optim}, there also exists $\mathbf{\Lambda^*} \in \real_{+}^{M \times C}$ such that $(\mathcal{W}^*,\mathbf{\Lambda}^*)$ represents a saddle point of the Lagrangian associated to \cref{eq:constraint_optim}. In this scenario, $\mathbf{\Lambda^*}$ are the Lagrange multipliers of the presented problem, and is intuitive to consider $\mathbf{\Lambda} = \mathbf{\Lambda^*}$ as the best choice to solve \cref{eq:loss_lagrangian}.

Nevertheless, using the Lagrange multipliers $\mathbf{\Lambda^*}$ as the weights for the penalties in \cref{eq:loss_lagrangian} may not be feasible in practice. In particular, a number of conventional strategies employed to train deep neural networks hinder straightforward minimization. First, the use of mini-batch gradient descent averages the updated prototypes for every single observation into a mean prototype per class, making a sample-wise constraint hard to achieve. Furthermore, performing data augmentation over the support samples may yield distinct penalty weights for the augmented versions, which could be harder or easier to classify than their original counterparts.

To alleviate the aforementioned challenges, we propose to relax the sample-wise penalties, which results in solving: 

\begin{equation}
\label{eq:loss_problem_to_solve}
\phantom{.}\min_{\mathcal{W}}  \quad 
\sum\limits_{m=1}^{M} \ent{\yy^{(m)},\hat{\yy}^{(m)}} +
\sum_{c=1}^C \lambda_{c} \; ||\ttt_c - \ww_{c}||_{2}^{2},
\end{equation}

\noindent where $\boldsymbol{\lambda} \in \real_+^C$ is a set of $C$ class-wise penalty weights. While the problem complexity has been reduced by removing sample-wise penalty weights, we still need to choose $C$ weights for the class-wise penalties. This poses a challenge in the optimization, particularly for datasets that contain a large number of categories, such as ImageNet \cite{deng2009imagenet} ($C=1000$), where properly selecting the penalty weights $\boldsymbol{\lambda} \in \real_+^C$ can be a laborious process. Furthermore, choosing these values ``by hand" juxtaposes with our goal of providing a \textit{validation-free} solution for ETL.

\subsection{Class Adaptive Constraint for Linear Probing}
\label{main:subsection_class_addaptive_lp}

\noindent \textbf{General Augmented Lagrangian.} Augmented Lagrangian Multiplier (ALM) methods present an appealing alternative for learning the penalty weights. These popular methods in optimization, which solve a constrained problem by the interplay of penalties and primal-dual steps, present well-known advantages \cite{dimi96lag,sangalli2021constrained}. Formally, we can define a general constrained optimization problem as:

\begin{equation}
\label{eq:general_ALM}
 \min_x \quad g(x) \quad \text{s.t.} \quad h_i(x)\leq 0, \quad i=1,\dots,n
\end{equation}

\noindent with $g:\real^d\rightarrow \real$ the \textit{objective function} and $h_i:\real^d\rightarrow\real, i=1,\dots,n$ the \textit{set of constraint functions}. This problem is generally tackled by solving a succession of $j\in \N$ unconstrained problems, each solved approximately w.r.t $x$:

\begin{equation}
\label{eq:lagrangian_alm}
    \min_{x,\lambda} \quad \Loss^{(j)}(x) = g(x) + \sum_{i=1}^n P(h_i(x), \rho_i^{(j)}, \lambda_i^{(j)}),
\end{equation}

\noindent with $P:\real \times \real_{++}\times \real_{++} \rightarrow \real$ a \textit{penalty-Lagrangian function}, whose derivative w.r.t. its first variable $P'(z,\rho,\lambda) \equiv \frac{\partial}{\partial z}P(z,\rho,\lambda)$ exists, is positive and continuous for all $z \in \real$ and $(\rho, \lambda) \in (\real_{++})^2$. The set of axioms that any penalty function $P$ must satisfy \cite{birgin2005numerical} are detailed in \appensecref{supp:subsection_penalties_axioms}. 
Furthermore, $\vrho^{(j)}=(\rho^{(j)}_i)_{1\leq i\leq n}\in \real_{++}^n$ and $\vlambda^{(j)}=(\lambda^{(j)}_i)_{1\leq i\leq n}\in\real_{++}^n$ denote the penalty parameters and multipliers associated to the penalty $P$ at the iteration $\textit{j}$. 

The ALM can be split into two iterations: \emph{outer} iterations (indexed by $j$), where the \textit{penalty multipliers} $\vlambda$ and the \textit{penalty parameters} $\vrho$ are updated, and the \emph{inner} iterations, where $\Loss^{(j)}$ (\cref{eq:lagrangian_alm}) is minimized using the previous solution as initialization. In particular, the penalty multipliers $\vlambda^{(j)}$ are updated to the derivative of $P$ w.r.t. to the solution obtained during the last \emph{inner} step:

\begin{equation}
    \label{eq:lambda_update}
    \lambda_i^{(j+1)}=P'(h_i(x), \rho_i^{(j)}, \lambda_i^{(j)}).
\end{equation}

\noindent By doing this, the penalty multipliers increase when the constraint is violated, and decrease otherwise. Thus, this strategy enables an \emph{adaptive} and \emph{learnable} way for determining the penalty weights.

\noindent \textbf{Our solution.} We propose to use an ALM approach to solve the problem in \cref{eq:loss_problem_to_solve}. In particular, we reformulate this problem integrating a penalty function $P$ parameterized by $(\vrho, \vlambda)\in\real_{++}^C\times\real_{++}^C$, formally defined as: 

\begin{equation}
\label{eq:final_problem}
\phantom{.}\min_{\mathcal{W},\vlambda}  \quad 
\sum\limits_{m=1}^{M} \ent{\yy^{(m)},\hat{\yy}^{(m)}}
+ \sum_{c=1}^C P(\ttt_c - \ww_{c}, \rho_c, \lambda_c).
\end{equation}

Following our realistic \textit{validation-free} scenario, the only data from which we can obtain feedback during adaptation is the support set $\mathcal{S}$. Thus, the penalty multiplier for class $c$ at epoch $j+1$ can be defined as:

\begin{equation}
    \label{eq:our_lambda_update}
    \lambda_c^{(j+1)}=\frac{1}{|\mathcal{S}|}\sum_{(\xx,\yy) \in \mathcal{S}}P'(\ttt_c-\ww_c, \rho_c^{(j)}, \lambda_c^{(j)}).
\end{equation}

As suggested by prior work \cite{birgin2005numerical,liu2023class}, we employ the PHR function as penalty $P$, defined as:

\begin{equation}
\label{eq:phr}
    \mathrm{PHR}(z,\rho,\lambda) =
    \begin{cases}
        \lambda z + \frac12 \rho z^2 \quad &\text{if} \quad \lambda + \rho z \geq 0;\\
        -\frac{\lambda^2}{2\rho} \quad &\text{otherwise}.
    \end{cases}
\end{equation}

Nevertheless, as we empirically found in our experiments (\appensecref{supp:subsection_ablation}), 
estimating Lagrange multipliers from the support samples might overfit the training data. As we do not have access to additional data points, we follow a simple strategy, consisting in performing only one iteration of the $\vlambda$ update. For a given target task, we rely on text embeddings as an anchor that offers a generalizable representation of concrete concepts along different visual domains. Thus, we consider the zero-shot prototypes $\ttt_c$ as the initial approximation of the problem in \cref{eq:our_lambda_update} (first \emph{inner} step). Instead of initializing $\vlambda$ randomly, which might hamper the convergence, we compute the penalty weight for a given class as the average of the zero-shot softmax scores for all support samples belonging to that class, such that $\slambda_c = \frac{1}{|\mathcal{B}_c^{+}|}\sum_{i\in\mathcal{B}_c^{+}} \hat{y}_{c}^{(i)}$, with $\mathcal{B}_c^{+}=\{ i|i \in M, {y}_{c}^{(i)} = 1 \}$. Note that these values are obtained by replacing $\ww_c$ with the solution found in the \emph{inner} step ($\ttt_c$) in \cref{eq:lp_pred}, which indeed satisfies the constraint $\ww_c=\ttt_c$, resulting in a zero penalty. Taking now the derivative w.r.t. $z$ of PHR, it is straightforward to see that the \emph{learned} value of $\vlambda$ after one iteration is indeed $\slambda_c$.

\section{Experiments}
\label{main:section_experiments}

\subsection{Setup}
\label{main:subsection_setup}

\noindent \textbf{Datasets: Few-shot adaptation.} We follow prior ETL literature \cite{yu2023task,gao2021clip,zhang2021tip} and benchmark all the methods on 11 datasets: Imagenet \cite{deng2009imagenet}, Caltech101 \cite{caltech}, OxfordPets \cite{oxfordpets}, StanfordCars \cite{stanfordcars}, Flowers102 \cite{flowers102}, Food101 \cite{food101}, FGVCAircraft \cite{aircraft}, SUN397 \cite{sun397}, DTD \cite{dtd}, EuroSAT \cite{eurosat}, and UCF101 \cite{ucf101}. These cover a diverse set of computer vision classification tasks, from general objects to actions or fine-grained categories in specialized applications. To train the few-shot adapters, we randomly retrieve $K$ shots ($K\in \{1, 2, 4, 8, 16\}$) for each class. Last, for evaluation, we used the test sets provided in each dataset, with the same data splits as \cite{zhou2022coop,yu2023task}.
\textbf{Domain generalization capabilities.} We further assess the model's \textit{robustness} to domain shifts by following existing ETL works. We used ImageNet as a source domain for adaptation, and its variants as target tasks, which include: ImageNetV2 \cite{imagenetV2}, ImageNet-Sketch \cite{imagenetSketch}, ImageNet-A \cite{imagenet_a}, and ImageNet-R \cite{imagenet_r}. In this scenario, the model only sees a few labeled samples from the source domain, and target data are used exclusively for testing. In addition, we also employ this setting to motivate the use of efficient adapters \textit{vs} fine-tuning the entire VLM \cite{yu2023task,FLYP,LPFT}. 

\noindent \textbf{Implementation details.} All experiments are based on CLIP \cite{radford2021learning} pre-trained features, using different backbones: ResNet-50 \cite{resnet} and ViT-B/16 \cite{dosovitskiy2020vit} (results for other backbones in \appensecref{supp:subsection_results}). We resort to ResNet-50 as backbone in the ablation studies. For each downstream task we first extract all pre-trained features of the support shots and then run adaptation experiments over those. Data augmentation is applied during the feature extraction stage using random zoom, crops, and flips, following \cite{zhou2022cocoop,yu2023task}. The number of augmentations per support sample is set to 20. We used the same text prompts per dataset as in \cite{zhou2022coop,yu2023task}. Following our claim that using a validation set on few-shot adaptation is unrealistic, we trained \mbox{ZS-LP} and CLAP using the same configuration for all datasets, number of shots, and visual backbones. Concretely, we optimize the adapter for $300$ epochs, using SGD optimizer with Momentum of $0.9$. We use a relatively large initial learning rate of $0.1$ to avoid underfitting on the support set, whose value decreases during training following a cosine decay scheduler. We ran all experiments with three different random seeds, and the results were averaged across runs.

\noindent \textbf{Baselines and adaptation protocol.} We selected adapter-based methods as our main competitors based on the similarity to our approach, including Clip-Adapter \cite{gao2021clip}, TIP-Adapter \cite{zhang2021tip}, TaskRes \cite{yu2023task}, and Cross-Modal \cite{lin2023crossmodal}. It is important to highlight that prior works \cite{gao2021clip,zhang2021tip,yu2023task} apparently leverage either the extensive test set, or an independent additional validation subset, to adjust important hyperparameters for few-shot adaptation, such as the learning rate, training epochs, and particular parameters that control each method \cite{lin2023crossmodal}. Nevertheless, as we exposed in \cref{fig:cross-shift}, their performance dramatically decreases when the set of hyperparameters is not adjusted for the testing scenario. To adhere to real-world requirements, we define a strict few-shot adaptation protocol, in which no validation or test samples are available to find the best case-specific configuration for each method, and hyperparameters remain fixed across tasks (details in \appensecref{supp:subsection_details_adapters}).

\subsection{Results}
\label{main:subsection_results}

\paragraph{Efficient transfer learning.}

We report in \cref{table_main_few_shot_results} the performance of adapter-based approaches averaged across 11 datasets, in the more realistic and practical \textit{validation-free} experimental setting. Furthermore, for prompt-learning-based approaches, we include the results reported in prior literature, for a more comprehensive comparison. From these values, we can make interesting observations. First, a well-initialized Linear Probe, \ie, using the CLIP zero-shot weights, does not show the performance degradation discussed in prior works, and it is indeed a competitive alternative to SoTA approaches. Second, and more surprisingly, more complex approaches such as CLIP-Adapter, or TIP-Adapter, show a significant decline in performance compared to their original results when no validation set is available for model selection. Interestingly, TaskRes(e), which is some sort of two-stage zero-shot initialization Linear Probing with an updated text projection, also offers robust performance. Nevertheless, the absence of a detailed explanation of how the enhanced version is obtained in the original work hampers fair comparisons. Third, constraining the weights update to remain close to the zero-shot knowledge (CLAP) shows consistent improvements across different shots, especially in the very low data regime. This suggests that retaining the previous base knowledge from VLMs is important to avoid diverging because of unrepresentative shots during adaptation. Results per dataset are detailed in \appencref{fig:etl} and \appencref{numerical_etl_comparison}.

\begin{table}[h!]
\caption{\textbf{Comparison to state-of-the-art methods} for few-shot adaptation of CLIP-based models, using ResNet-50 backbone. ETL methods are trained under the same protocol, \ie, absence of a validation set and using a fixed configuration across datasets, and results are averaged across 11 datasets. Prompt-learning methods results are directly extracted from \cite{kgcoop23,chen2023plot}. Best results in bold.}
\label{table_main_few_shot_results}
\vspace{-2mm}
\resizebox{0.47\textwidth}{!}{
\centering
\begin{tabular}{lccccc}
\toprule
\multicolumn{1}{c}{\multirow{1}{*}{Method}} & $\rms K=1$     & $\rms K=2$    & $\rms K=4$    & $\rms K=8$    & $\rms K=16$   \\
\midrule
\multicolumn{6}{l}{\textit{Prompt-learning} methods} \\
\hdashline\noalign{\vskip 0.5ex}
CoOp$_\text{ IJCV'22}$\cite{zhou2022coop}                  & 59.56 & 61.78 & 66.47 & 69.85 & 73.33  \\
ProGrad$_\text{ ICCV'23}$\cite{kgcoop23}                   & 62.61 & 64.90 & 68.45 & 71.41 & 74.28  \\
PLOT$_\text{ ICLR'23}$\cite{chen2023plot}                  & 62.59 & 65.23 & 68.60 & 71.23 & 73.94  \\
\midrule
\multicolumn{6}{l}{Efficient transfer learning - \textit{a.k.a} \textit{Adapters}} \\
\hdashline\noalign{\vskip 0.5ex}
Zero-Shot$_\text{ ICML'21}$\cite{radford2021learning}     & 57.71 & 57.71  & 57.71 & 57.71 & 57.71  \\
Rand. Init LP$_\text{ ICML'21}$\cite{radford2021learning}  & 30.42 & 41.86  & 51.69 & 60.84 & 67.54  \\
CLIP-Adapter$_\text{ IJCV'23}$\cite{gao2021clip}           & 58.43 & 62.46  & 66.18 & 69.87 & 73.35  \\
TIP-Adapter$_\text{ ECCV'22}$\cite{zhang2021tip}           & 58.86 & 60.33  & 61.49 & 63.15 & 64.61  \\
TIP-Adapter(f)$_\text{ ECCV'22}$\cite{zhang2021tip}        & 60.29 & 62.26  & 65.32 & 68.35 & 71.40  \\
CrossModal-LP$_\text{ CVPR'23}$\cite{lin2023crossmodal}    & 62.24 & 64.48  & 66.67 & 70.36 & 73.65  \\
TaskRes(e)$_\text{ CVPR'23}$\cite{yu2023task}              & 61.44 & 65.26  & 68.35 & 71.66 & 74.42  \\
\rowcolor{Gray}ZS-LP                                       & 61.28 & 64.88  & 67.98 & 71.43 & 74.37  \\
\rowcolor{Gray}CLAP                                    & \textbf{62.79} & \textbf{66.07} & \textbf{69.13} & \textbf{72.08} & \textbf{74.57}  \\
\bottomrule
\end{tabular}
}
\vspace{-5mm}
\end{table}

\paragraph{Domain generalization.}

If adaptation is not carefully conducted, the resulting model might distort the pre-trained knowledge and underperform when new data with domain drifts is involved \cite{LPFT}, even below the zero-shot (no adaptation) performance. Thus, evaluating the robustness of novel adapters under this scenario of domain generalization is of special interest. To do so, adapters are optimized on ImageNet using 16 shots per class, and directly evaluated on ImageNet variants. In this setting, we also assume the absence of a validation dataset, and hence all adapters are trained until convergence, using the same configuration across backbones. A summary of the results is reported in \cref{results_generalization_main}, while specific numbers across datasets and additional backbones are included in \appencref{results_generalization_supp}. From these experiments, we make two striking observations. First, \mbox{ZS-LP} is a strong baseline compared to other more complex adapters on the source domain. Even more remarkably, prior SoTA adapters, such as CLIP-Adapter or TIP-Adapter, fail to generalize to unseen domains. Indeed, when using recent vision transformers, which are overtaking convolutional neural networks, \textbf{none of existing adapters-based approaches outperform standard zero-shot prediction in the presence of distributional drifts.} In contrast, CLAP yields the best in-distribution performance and also shows consistent improvements under domain shifts across all backbones.

\begin{table}[h!]
\caption{\textbf{Robustness to domain shifts.} Adapters are adjusted on ImageNet and evaluated at out-of-distribution generalization on 4 ImageNet shifts. Bold indicates best performance. Differences with respect to no adaptation (\textit{a.k.a} zero-shot) are highlighted.}
\label{results_generalization_main}
\vspace{-2mm}
\centering
\scriptsize
\begin{tabular}{clcc}
\toprule
 & \multicolumn{1}{c}{Method} & \multicolumn{1}{c}{Source (Imagenet)} & \multicolumn{1}{c}{Target (Average)} \\ 
\midrule
\multicolumn{1}{c}{\multirow{8}{*}{\rotatebox{90}{ResNet-50}}} & Zero-Shot$_\text{ ICML'21}$\cite{radford2021learning}       & 60.35      & 40.61 \\
 &  Rand. Init LP$_\text{ ICML'21}$\cite{radford2021learning}  &    52.24$_{(-8.11)}$\textcolor{red}{$\downarrow$}        & 24.61$_{(-16.00)}$\textcolor{red}{$\downarrow$} \\
 &  CLIP-Adapter$_\text{ IJCV'23}$\cite{gao2021clip}           &          59.02$_{(-1.33)}$\textcolor{red}{$\downarrow$}    &    31.21$_{(-9.40)}$\textcolor{red}{$\downarrow$} \\
 & TIP-Adapter$_\text{ ECCV'22}$\cite{zhang2021tip}           &   57.81$_{(-2.54)}$\textcolor{red}{$\downarrow$}      & 40.69$_{(+0.08)}$\textcolor{blue}{$\uparrow$} \\
 & TIP-Adapter(f)$_\text{ ECCV'22}$\cite{zhang2021tip}        &     62.27$_{(+1.92)}$\textcolor{blue}{$\uparrow$}          & 41.36$_{(+0.75)}$\textcolor{blue}{$\uparrow$} \\
 & TaskRes(e)$_\text{ CVPR'23}$\cite{yu2023task}              &             60.85$_{(+0.50)}$\textcolor{blue}{$\uparrow$}          &       41.28$_{(+0.67)}$\textcolor{blue}{$\uparrow$} \\
 & \cellcolor{Gray}ZS-LP                                       &        \cellcolor{Gray}61.00$_{(+0.65)}$\textcolor{blue}{$\uparrow$}               & \cellcolor{Gray}36.58$_{(-4.03)}$\textcolor{red}{$\downarrow$} \\
 & \cellcolor{Gray}CLAP                                         &      \cellcolor{Gray}\textbf{65.02}$_{(+4.67)}$\textcolor{blue}{$\uparrow$} &  \cellcolor{Gray}\textbf{42.91}$_{(+2.30)}$\textcolor{blue}{$\uparrow$} \\
\midrule
\multicolumn{1}{c}{\multirow{8}{*}{\rotatebox{90}{ViT-B/16}}} & Zero-Shot$_\text{ ICML'21}$\cite{radford2021learning}      &    68.71     & 57.17 \\
 & Rand. Init LP$_\text{ ICML'21}$\cite{radford2021learning}  &       62.95$_{(-5.76)}$\textcolor{red}{$\downarrow$}          & 40.41$_{(-16.76)}$\textcolor{red}{$\downarrow$} \\ 
 & CLIP-Adapter$_\text{ IJCV'23}$\cite{gao2021clip}           &                      68.46$_{(-0.25)}$\textcolor{red}{$\downarrow$}           & 50.72$_{(-6.45})$\textcolor{red}{$\downarrow$} \\
 & TIP-Adapter$_\text{ ECCV'22}$\cite{zhang2021tip}           &         53.81$_{(-14.90)}$\textcolor{red}{$\downarrow$}         &       41.55$_{(-15.62)}$\textcolor{red}{$\downarrow$} \\ 
 & TIP-Adapter(f)$_\text{ ECCV'22}$\cite{zhang2021tip}        &                    51.71$_{(-17.00)}$\textcolor{red}{$\downarrow$}           & 35.58$_{(-21.6)}$\textcolor{red}{$\downarrow$} \\  
 & TaskRes(e)$_\text{ CVPR'23}$\cite{yu2023task}              &            70.84$_{(+2.13)}$\textcolor{blue}{$\uparrow$}           &      55.35$_{(-1.82)}$\textcolor{red}{$\downarrow$} \\
 & \cellcolor{Gray}ZS-LP                                       &         \cellcolor{Gray}69.73$_{(+1.02)}$\textcolor{blue}{$\uparrow$}           &         \cellcolor{Gray}53.65$_{(-3.52)}$\textcolor{red}{$\downarrow$} \\ 
 & \cellcolor{Gray}CLAP                                     &        \cellcolor{Gray}\textbf{73.38}$_{(+4.67)}$\textcolor{blue}{$\uparrow$}  & \cellcolor{Gray}\textbf{60.04}$_{(+2.87)}$\textcolor{blue}{$\uparrow$} \\ 
\bottomrule
\end{tabular}
\vspace{-5mm}
\end{table}

\begin{table}[h!]
\caption{\textbf{Fine-tuning (FT) vs. efficient transfer learning (ETL).} 
A benchmark for the low data regime, \ie, 8 shots for each class. For the sake of fairness, FT methods (above the dashed line) are trained with 4 shots and early-stopped using a validation set containing 4 shots. On the other hand, ETL methods (below the dashed line) are trained using 8 shots and rely solely on the support set. All methods use ViT-B/16 as CLIP backbone.}
\label{ft_vs_etl_main}
\vspace{-2mm}
\resizebox{0.47\textwidth}{!}{
\centering
\begin{tabular}{lcccccc}
\toprule
\multicolumn{1}{c}{\multirow{2}{*}{Method}} & \multicolumn{1}{c}{Source} & \multicolumn{5}{c}{Target} \\ \cline{3-7} 
\multicolumn{1}{c}{} & \multicolumn{1}{c}{Imagenet} & \multicolumn{1}{c}{-V2} & \multicolumn{1}{c}{-Sketch} & \multicolumn{1}{c}{-A} & \multicolumn{1}{c}{-R} & \multicolumn{1}{c}{Avg.} \\ 
\midrule
Fine-tuning (FT)                                                         & 69.88 & 62.44 & 47.07 & 47.52 & 76.08 & 58.28 \\
LP-FT$_\text{ ICLR'23}$ \cite{LPFT}                         & 71.29 & 64.04 & 48.50 & 49.49 & 77.63 & 59.92 \\
WiSE$_\text{ CVPR'22}$ \cite{WiSE}                       & 71.17 & 63.81 & 49.38 & 50.59 & 78.56 & 60.59 \\
FLYP$_\text{ CVPR'23}$ \cite{FLYP}                          & 71.51 & 64.59 & 49.50 & 51.32 & 78.52 & \textbf{60.98} \\
\hdashline\noalign{\vskip 0.5ex}
Zero-Shot                                 & 68.71 & 60.76 & 46.18 & 47.76 & 73.98 & 57.17 \\
Rand. Init LP                                        & 56.58 & 47.17 & 25.82 & 27.03 & 47.05 & 36.77 \\
\rowcolor{Gray}ZS-LP                                     & 68.49 & 60.07 & 42.77 & 42.39 & 71.73 & 54.24 \\
\rowcolor{Gray}CLAP                    & \textbf{71.75} & 64.06 & 47.66 & 48.40 & 76.70 & 59.21 \\
\bottomrule
\multicolumn{7}{l}{*Specific numbers for FT, LP-FT, WiSE-FT, and FLYP are retrieved from \cite{FLYP}.}\\
\end{tabular}
}
\vspace{-8mm}
\end{table}

\paragraph{Is it worth optimizing the entire model?}

We now compare CLAP to end-to-end full fine-tuning (FT) approaches: LP-FT \cite{LPFT}, WiSE-FT \cite{WiSE}, and FLYP \cite{FLYP}. The former two methods require a validation set for early stopping, and the latter two use it for both early stopping and tuning the \textit{mixing coefficient} hyperparameter $\alpha$. Therefore, for a $K$-shot problem, these methods actually require $2K$ shots for each class, $K$ for training, and $K$ for validation. As the balancing penalty term in CLAP is optimized with the support set, and does not require a validation set, a fair comparison would be to evaluate the $K$-shot performance of fine-tuning methods against our method's $2K$-shot results. Thus, \cref{ft_vs_etl_main} includes the performance of all the models when 8 labeled images are available for each class overall. Analyzing the results, we can conclude that in the low data regime, full fine-tuning is not necessarily superior to ETL when compared properly. More specifically, our approach outperforms fine-tuning methods in in-distribution performance and performs reasonably well on OOD datasets, while having a fraction of the optimizable parameters of fine-tuning methods. 

\subsection{Ablation experiments}
\label{main:subsection_ablation}

\paragraph{On the need for model selection strategies.} Relevant methods (\eg, CLIP-Adapter \cite{gao2021clip}, TIP-Adapter \cite{zhang2021tip}, or TaskRes \cite{yu2023task}) include different hyperparameters that directly control their performance. Nevertheless, these methods are \textit{incomplete}, since they do not include any strategy for adjusting these parameters, typically referred to as \textit{model selection}. In contrast, and as previously stressed, there is evidence that these works use a large evaluation subset to adapt their settings to each scenario \cite{lin2023crossmodal}. To investigate this observation, we evaluate these methods in cross-dataset model selection experiments. The best hyperparameters values for a task (\ie, dataset), which are found in an Oracle scenario using the entire test subset, are used during adaptation to another dataset. The matrices showing the relative improvements over a zero-shot initialized Linear Probing (\mbox{ZS-LP}) are depicted in \cref{fig:cross-shift}. These results show empirically that the hyperparameters values are highly task-dependent, and \textbf{that SoTA methods must adjust their hyperparameters on the target task to outperform this simple baseline,} which is \textit{unrealistic} in practice. In contrast, the proposed CLAP is more robust, showing consistent results across all datasets, even in the worst degradation case, as it does not require particular modifications per task. 

\begin{table}[h!]
\caption{\textbf{Improving Linear Probing.} Using as baseline the proposed \mbox{ZS-LP} configuration detailed in \cref{main:subsection_lp}, we isolate the effect of removing different parts of the model, while keeping the rest static. Results are averaged across 11 datasets.}
\vspace{-2mm}
\label{linear_probing}
\resizebox{0.47\textwidth}{!}{
\centering
\begin{tabular}{lccc}
\toprule
\multicolumn{1}{c}{\multirow{1}{*}{Method}} & $\rms K=1$     & $\rms K=2$    & $\rms K=4$ \\
\midrule
\rowcolor{Gray}ZS-LP             & 61.28 & 64.88 & 67.98 \\
\hdashline\noalign{\vskip 0.5ex}
w/o DA                           & 57.72$_{(-3.5)}$\textcolor{red}{$\downarrow$}  & 61.94$_{(-2.9)}$\textcolor{red}{$\downarrow$}  & 65.41$_{(-2.5)}$\textcolor{red}{$\downarrow$} \\
w/o Temp. Scaling ($\tau$)       & 58.33$_{(-2.9)}$\textcolor{red}{$\downarrow$} & 59.85$_{(-5.0)}$\textcolor{red}{$\downarrow$} & 59.91$_{(-8.0)}$\textcolor{red}{$\downarrow$} \\
w/o $L^2$-norm                      & 48.67$_{(-12.6)}$\textcolor{red}{$\downarrow$} & 55.29$_{(-9.6)}$\textcolor{red}{$\downarrow$} & 61.16$_{(-6.8)}$\textcolor{red}{$\downarrow$} \\
Rand. Init.                      & 30.42$_{(-30.8)}$\textcolor{red}{$\downarrow$} & 41.86$_{(-23.0)}$\textcolor{red}{$\downarrow$} & 51.69$_{(-16.2)}$\textcolor{red}{$\downarrow$} \\
\bottomrule
\end{tabular}
}
\vspace{-5mm}
\end{table}

\paragraph{Details in Linear Probing matter.} As described earlier in \cref{main:subsection_lp}, LP has been discouraged in the prior literature due to its limited performance in few-shot adaptation \cite{radford2021learning,gao2021clip}. Nevertheless, we argue that this behavior stems from the original way in which LP was introduced in \cite{radford2021learning}, inspired by prior self-supervised learning methods. Indeed, a strategy tailored to contrastive VLMs alleviates the performance drop of LP observed in prior works. In particular, using zero-shot initialization, the same temperature scaling as pre-training, and explicit $\ell_{2}$-normalization of the class prototypes, considerably improves the generalization of few-shot adaptation (\cref{linear_probing}). This aligns with relevant literature on other topics such as FT \cite{FLYP}, which suggests that the adaptation conditions should match the pre-training setting. Also, including other heuristics such as data augmentation (DA), usually omitted in LP \cite{zhang2021tip,yu2023task}, is of special relevance.

\paragraph{Using a few-shot validation set.} Cross-Modal adapter \cite{lin2023crossmodal} uses a validation set composed of ($min(K, 4)$) samples to adjust the experimental setting and early stopping. Even though this setting is more appropriate, it still requires an additional number of shots for model selection. Nevertheless, for the sake of fairness, the performance comparison to methods that do not require a validation set should be carried out by training the latter methods using $K+min(K, 4)$ shots. When this fair benchmark is established (see \cref{validation_set}), simple \mbox{ZS-LP} excels again as a strong baseline, outperforming more complex methods on the low-shot regime. Only when using a large number of shots ($K>8$) partial fine-tuning and ETL methods marginally benefit from validation samples. However, model selection using a validation set increases the computational workload and processing times during adaptation due to its grid search nature.

\begin{table}[h!]
\caption{\textbf{Using a few-shot validation set.} Results for priors works on this setting are obtained from \cite{lin2023crossmodal}. Average across 11 datasets.}
\label{validation_set}
\vspace{-2mm}
\centering
\scriptsize
\begin{tabular}{lccccc}
\toprule
\multicolumn{1}{c}{\multirow{1}{*}{Method}} & $\rms K=1$     & $\rms K=2$    & $\rms K=4$    & $\rms K=8$    & $\rms K=16$   \\
\midrule
\multicolumn{6}{l}{Protocol in \cite{lin2023crossmodal}: $K$-shots for train + $min(K, 4)$ for validation} \vspace{0.5mm}                                                   \\
\hdashline\noalign{\vskip 0.5ex}
TIP-Adapter \cite{zhang2021tip}                                & 63.3 & 65.9 & 69.0 & 72.2 & 75.1       \\
CrossModal LP \cite{lin2023crossmodal}                         & 64.1 & 67.0 & 70.3 & 73.0 & 76.0       \\
CrossModal Adapter \cite{lin2023crossmodal}               & 64.4 & 67.6 & 70.8 & 73.4 & 75.9       \\
CrossModal PartialFT \cite{lin2023crossmodal}                  & 64.7 & 67.2 & 70.5 & \textbf{73.6} & \textbf{77.1}       \\
\midrule
\multicolumn{6}{l}{Ours: using $K+min(K, 4)$ shots for training}  \vspace{0.5mm} \\
\hdashline\noalign{\vskip 0.5ex}
\rowcolor{Gray}ZS-LP                                        & 64.9   & 68.0 & 71.4 & 73.1 & 75.0      \\
\rowcolor{Gray}CLAP                      & \textbf{66.1}   & \textbf{69.1} & \textbf{72.1} & \textbf{73.5} & 75.1      \\
\bottomrule
\end{tabular}
\vspace{-5mm}
\end{table}

\section{Limitations}
\label{main:section_limitations}

In this work, we have introduced a CLass-Adaptive linear Probe (CLAP) objective, based on an adaptation of the general Augmented Lagrangian method, for efficient adaptation of large vision-language models in realistic scenarios. Despite its superiority, our empirical validation suggests that the benefits of our approach diminish as the number of shots increases, indicating that other strategies might be privileged if the number of adaptation samples is large.

\section*{Acknowledgments}

This work is supported by the National Science and Engineering Research Council of Canada (NSERC) and Fonds de recherche du Québec (FRQNT). We also thank Calcul Quebec and Compute Canada.

{\small
\bibliographystyle{ieeenat_fullname}
\bibliography{refs}
}

\clearpage
\appendix

\setcounter{section}{0}
\renewcommand{\thesection}{\Alph{section}}

\maketitlesupplementary

\section{A closer look to the pitfalls of previous vision-language adapters}
\label{supp:section_adapters}

\begin{figure*}[h!]
    \begin{center}
        \begin{tabular}{cccc}

        (a) \textbf{CLIP-Adapter} \cite{gao2021clip} & (b) \textbf{TIP-Adapter} \cite{zhang2021tip} & (c) \textbf{TIP-Adaper(f)} \cite{zhang2021tip} & \\
        
         \includegraphics[width=.31\linewidth]{arXiv/figures/cross_shift_clipA.png} &
         \includegraphics[width=.31\linewidth]{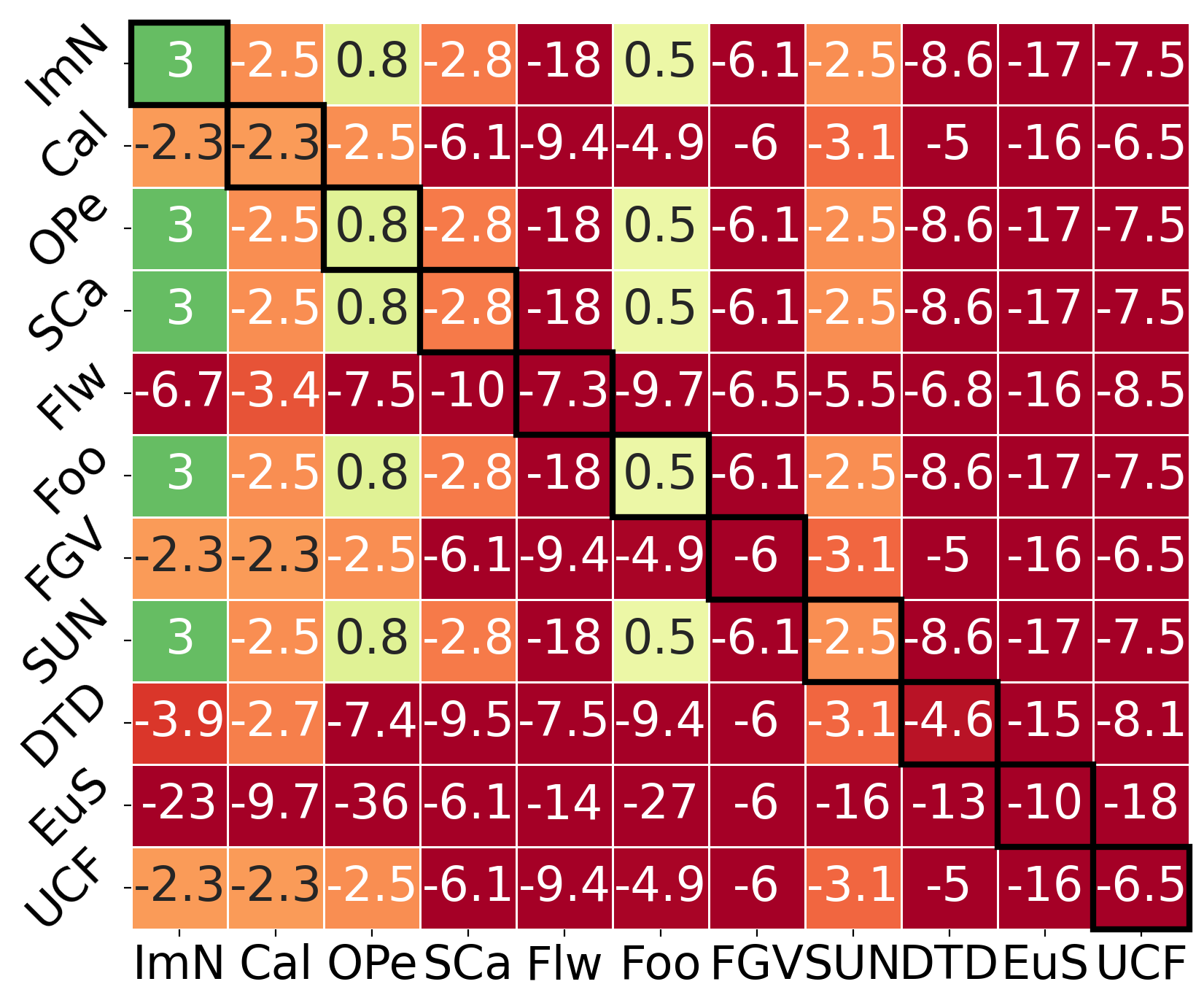} &
         \includegraphics[width=.31\linewidth]{arXiv/figures/cross_tipAf.png} & \\

        (d) \textbf{TaskRes} \cite{yu2023task} & (e) \textbf{CLAP \textit{(Ours)}} & & \\
         
         \includegraphics[width=.31\linewidth]{arXiv/figures/cross_taskres.png} &  
          \includegraphics[width=.31\linewidth]{arXiv/figures/cross_shift_zslpc.png} 
          & \hspace{-50mm} \multirow{7}{*}[10.75em]{\includegraphics[width=.04\linewidth]{arXiv/figures/cross_shift_colobar.png}} & \\
         
        \end{tabular}
        \caption{\textbf{Pitfalls of few-shot adapters due to the absence of a \textit{model selection} strategy - Additional methods.} The cross-shift model selection matrices $(i,j)$ depict the relative improvement w.r.t. a zero-shot initialized Linear Probing when using the optimal hyperparameters for the dataset $i$, for adapting in another task $j$, for each SoTA method (\textit{first four plots}) and our approach (\textit{last plot}). This is an extended version of \cref{fig:cross-shift} in the main manuscript.}
        \label{fig:cross-shift_supplementary}
    \end{center}\vspace{-5mm}
\end{figure*}

In this work, we provide a closer view of the pitfalls of current literature on few-shot vision-language adapters of large vision-language models. In particular, we observe that recently proposed adapters rely on a large test subset to adjust important hyperparameters per dataset, and thus become \textit{impractical} in real-world few-shot scenarios. This limitation becomes evident when fixing the hyperparameters on a given scenario and testing the model on other tasks, SoTA methods typically see their performance degrade compared to aa well-initialized Linear Probing (see \appencref{fig:cross-shift_supplementary}). In the following section, we aim to depict a detailed view of these methods and the reasons that underlay their promising reported performance.

\subsection{What are SoTA adapters doing?}
\label{supp:subsection_adapters_method}

We observe two concurrent phenomena on current SoTA vision-language adapters: (i) they use a good initialization, based on the zero-shot prototypes; and (ii) they introduce a set of empirically-fixed hyperparameters that control the divergence from the initial set of initial zero-shot prototypes.

\noindent \textbf{CLIP-Adapter} \cite{gao2021clip}. The inference relies on the zero-shot inference as in \cref{eq:zs_prediction}, using the same prototypes, which remain static. During training, CLIP-Adapter trains a residual MLP block to refine the visual features, such that $\mathbf{v'}=\mathbf{v}+\alpha_\mathrm{r}f_{\psi}(\mathbf{v})$. This method explicitly keeps the class prototypes close to the zero-shot initialization, while modifies the input visual features. This modification can be controlled with the residual ratio, $\alpha_\mathrm{r}$, together with the used learning rate and early stopping at specific epochs.

\noindent \textbf{TIP-Adapter} \cite{zhang2021tip}. Training-free CLIP proposes a multimodal combination of logits, using two terms: (i) a weighted similarity to the support sample, and (ii), the similarity of the zero-sot prototypes. This dual formulation can be expressed in the following formula:

 \begin{align}
\label{eq:tipA}
\begin{split}
\logit_{c} = \underbrace{\alpha_{\mathrm{tipA}} f_{\psi}(\vv, \beta)}_{\text{vision logits}} + \underbrace{\tau \; \vv \cdot \ttt_{c}^\top}_{\text{zero-shot logits}}
\end{split}
\end{align}

\noindent where $\alpha_{\mathrm{tipA}}$ and $\beta$ are control hyperparameters, which are empirically fixed.

There are two versions of TIP-Adapter. First, a training-free version, in which the vision logits are obtained by the post-processed version of the average cosine similarity between the vision embedding of the target and the support samples per class. Second, a trainable version in which, additionally, the vision embeddings from the support set are tunned, which dramatically increases the number of trainable parameters with the number of shots.

Since details matter, it is worth mentioning that in the combined logits depicted in \appencref{eq:tipA}, temperature scaling is only applied on the cosine similarity of text prototypes. The $\tau$ value is learned during training and usually converges to large values, which are clipped at a maximum value of $100$. This scaling makes the logits obtained from the zero-shot weights \textit{dominate} in the combined formulation if $\alpha$ is not properly fixed to large values. As we previously stated, this results in an initialization close to the zero-shot prototypes, and the deviation from this solution is carefully controlled with an $\alpha$ scaling per dataset.

\noindent \textbf{Task Residual Learning} \cite{yu2023task}. TaskRes uses a linear classifier to obtain class prototypes, following \cref{eq:lp_pred}. In particular, the authors propose to train a \textit{``prior-independent task residual"}, which follows a re-parametrization of the learned prototypes $\ww$, such that $\ww=t+\alpha \ww_r$, where $t$ is the language prototypes for the target classes (zero-shot weights), and $w_r$ is a learnable matrix that modifies them. This modification is controlled by the hyperparameter $\alpha$, which is empirically fixed for each dataset. Since $\ww_r$ is initialized to a matrix filled with zeros, the re-parametrization is equivalent to using a zero-shot initialization at the first iteration. In addition, given a feature vector $\mathbf{v}$ of a support sample, and optimizing $\ww_r$ via gradient descent using \cref{eq:ceLoss}, it is straightforward to derive that this term simply introduces a scaling factor on a given learning rate $\eta$, and no additional information is introduced to a simple Linear Probing:

\begin{align}
\label{eq:taskres}
\begin{split}
\ww_{r}^{t} = \ww_{r}^{t-1} - \eta \frac{\partial \ent{\yy,\hat{\yy}}}{\partial \ww_{r}} = \ww_{r}^{t-1} - \eta \frac{\partial \ent{\yy,\hat{\yy}}}{\partial \ww} \frac{\partial \ww}{\partial \ww_{r}} = \\
\ww_{r}^{t-1} - \underbrace{(\eta \ \alpha)}_{\text{learning rate}} \ \mathbf{v} \ (\hat{\yy} - \yy) \ .
\end{split}
\end{align}

\noindent \textbf{Other related literature}. Albeit proposed in the context of full fine-tuning of vision-language models in the large data regime, WiSE \cite{WiSE} approach also introduces some insights on efficient adaptation using a simple linear classifier. In particular, the authors study the benefit of linear interpolation between fine-tuned and zero-shot (initial) weights. In the case of adjusting uniquely a linear classifier, this method would be equivalent to balancing the text embeddings and the trained prototypes, such that $w=\alpha \; w_\mathrm{LP} + (1-\alpha) \; t$. Concretely, the ratio $\alpha$ is fixed using a validation subset after training, The idea that underlays this method aligns with the observations derived from the few-shot adapters.

\subsection{Remaining close to the initial zero-shot prototypes}
\label{supp:subsection_zs_init}

\begin{figure*}[h!]
    \begin{center}
        \begin{tabular}{cc}

        {\textbf{ImageNet}} &  {\textbf{Flowers102}} \\
        
         \includegraphics[width=0.40\linewidth]{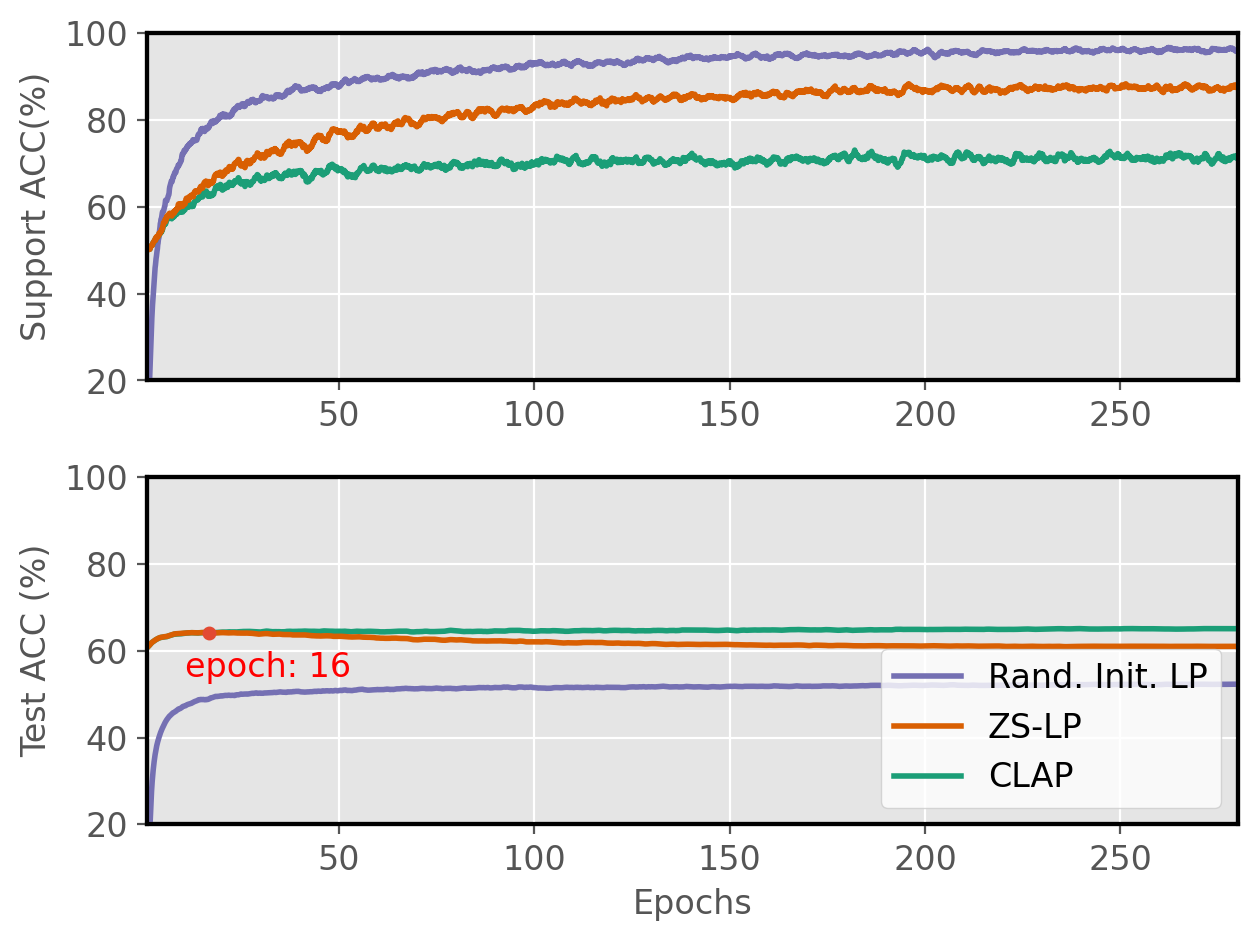} &
         \includegraphics[width=0.40\linewidth]{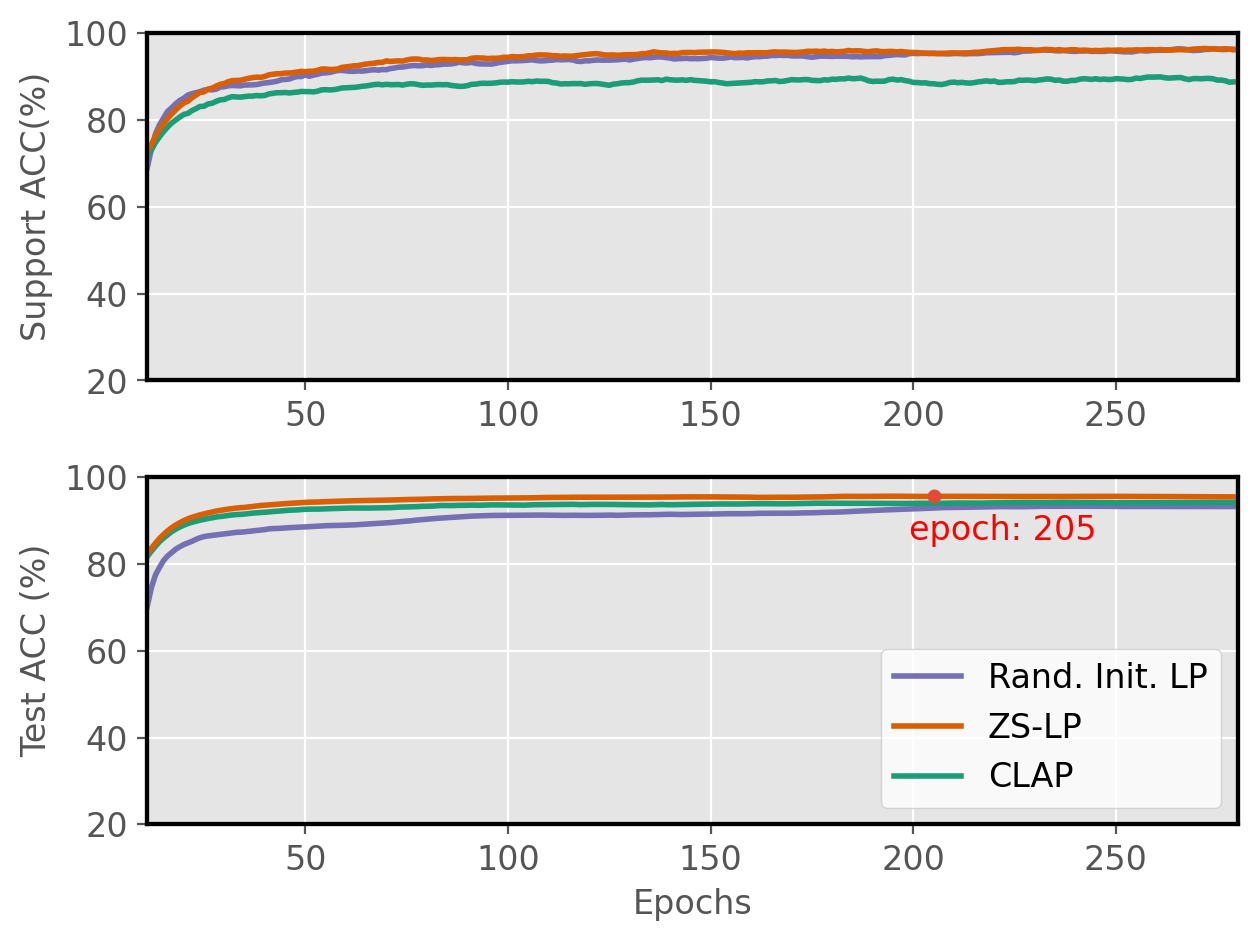} \\

        {\textbf{Food101}} &  {\textbf{SUN397}} \\

         \includegraphics[width=0.40\linewidth]{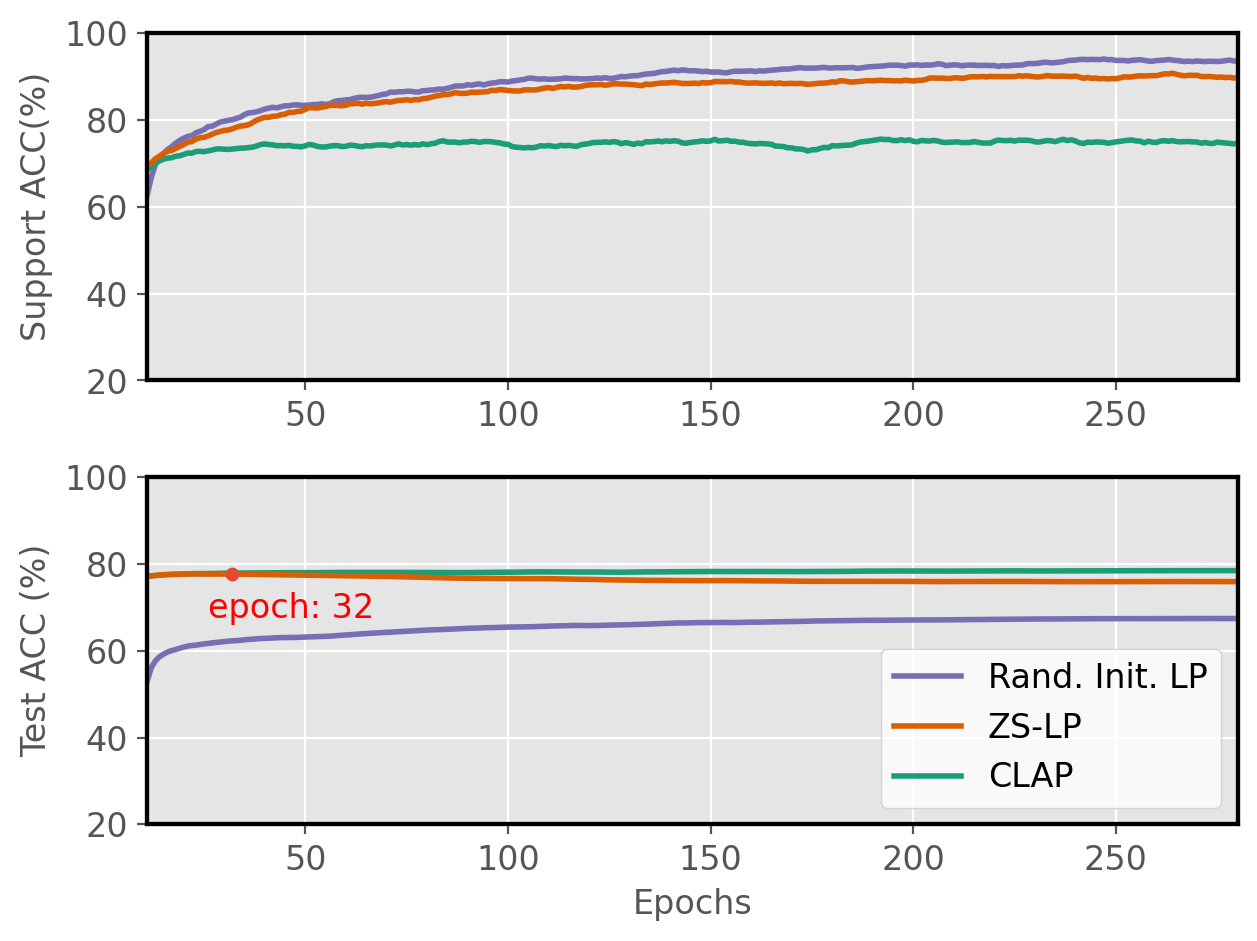} &
         \includegraphics[width=0.40\linewidth]{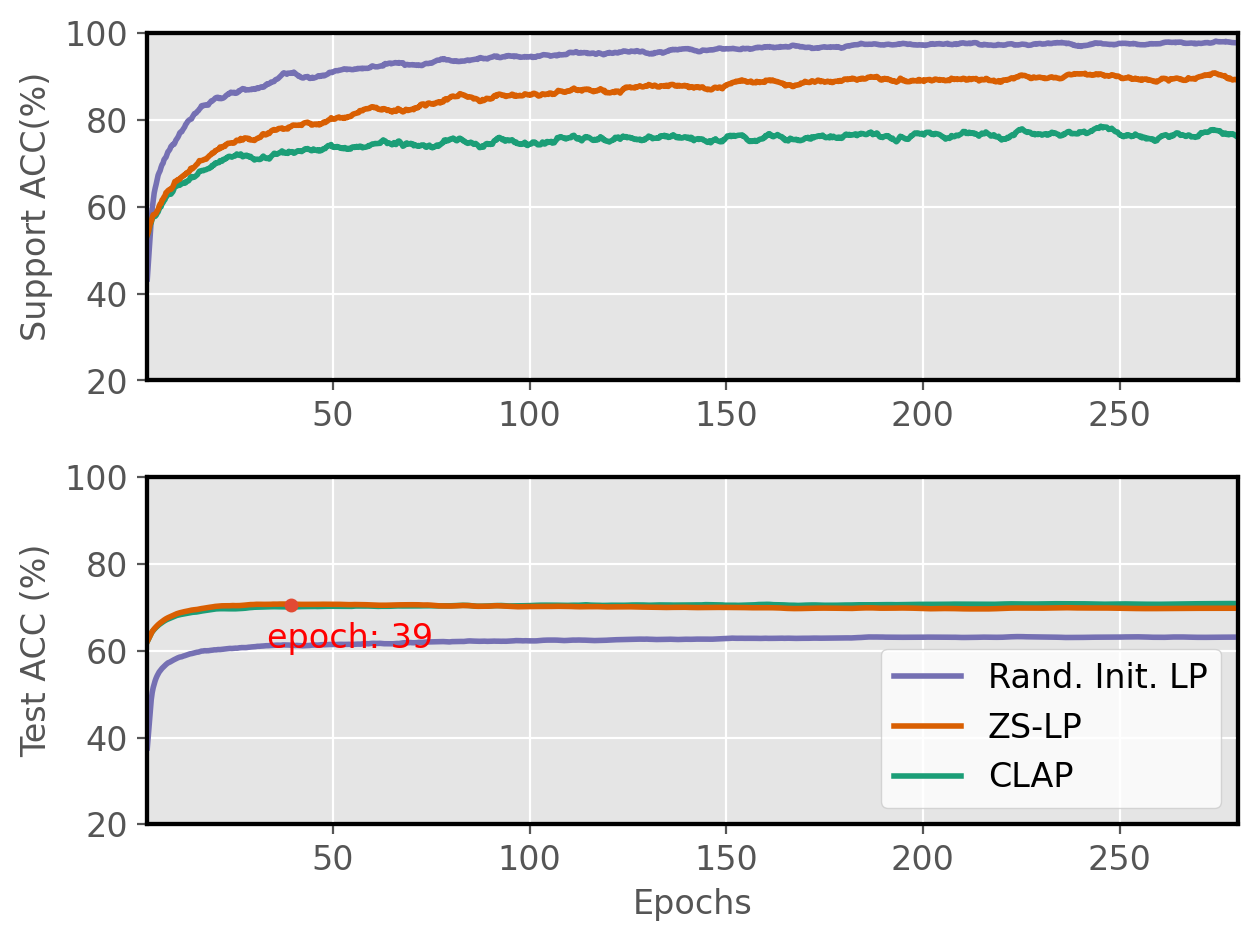} \\

        \end{tabular}
        \caption{\textbf{Linear Probing learning curves}. Results of Linear Probing-based methods when adapted to ImageNet using ResNet-50 as a backbone, $16$ shots per class as a support set, and a training scheduler using SGD. During training, both support set accuracy (\textit{top}) and the performance on the test subset (\textit{bottom}) are monitored, and the maximum test accuracy is highlighted in the curves. The training scheduler is described in \cref{main:subsection_setup}.}
        \label{fig:lc}
    \end{center}\vspace{-5mm}
\end{figure*}

As we stated previously, we observed that recently proposed few-shot adapters benefit from a good initialization, which is obtained from robust class-wise text embeddings. Also, in the main manuscript, we introduce a revisited Linear Probing baseline tailored for vision-language models (\mbox{ZS-LP} in \cref{main:subsection_lp}). This method benefits from this good initialization, together with other training heuristics. Indeed, we demonstrate empirically in the experimental section that it serves as a strong baseline for VLMs adaptation. We now study the convergence of this method during adaptation (curves in \appencref{fig:lc}), to shed light on the benefits of using a good set of initial prototypes. Furthermore, our goal is to expose that in different datasets, deviating much from initial prototypes may, or may not, be beneficial. We stress that, as a reminder, the more iterations are performed during adaptation, the more the model predictions deviate from initial zero-shot representations, which can also be controlled with the step size, \textit{a.k.a.}, learning rate. First, we can observe that, the zero-shot CLIP initialized Linear Probe (\textit{orange line}) achieves a maximum in performance over test samples at different epochs, which do not correspond to the convergence on the support set. Indeed, letting the adaptation converge typically yields performance degradation in ZS-LP. Even though this solution (i.e., maximum performance on test samples) could be reached using a large validation subset, which can be used for tuning the hyperparameters and early stopping, its presence is unrealistic on a strict few-shot protocol. In contrast, it is worth mentioning that the proposed learnable class-adaptative Linear Probing (CLAP, see \cref{main:subsection_class_addaptive_lp}) prevents this degradation, and does not require access to any additional data. Last, we would like to highlight an interesting observation from the convergence points seen in these curves. In particular, and interestingly, the range of values for searching the corresponding hyper-parameters in methods such as TIP-Adapter, varies with the convergence scenario for the best test performance (more details in the next section).

\begin{figure}[h!]
    \begin{center}
        \begin{tabular}{cc}

        \scriptsize{\textbf{ZS-LP - OxfordPets}} &  \scriptsize{\textbf{ZS-LP - Flowers102}} \\
        
         \includegraphics[width=0.47\linewidth]{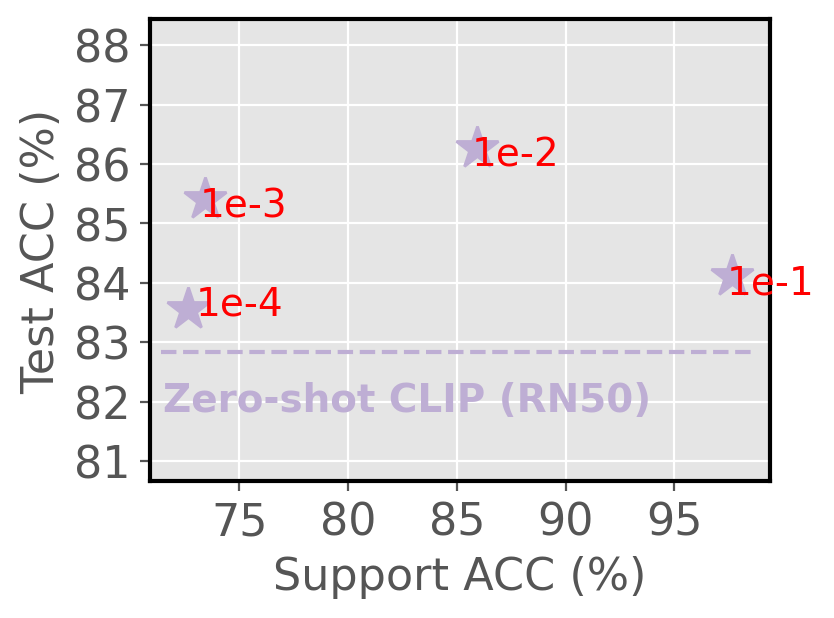} &
         \includegraphics[width=0.47\linewidth]{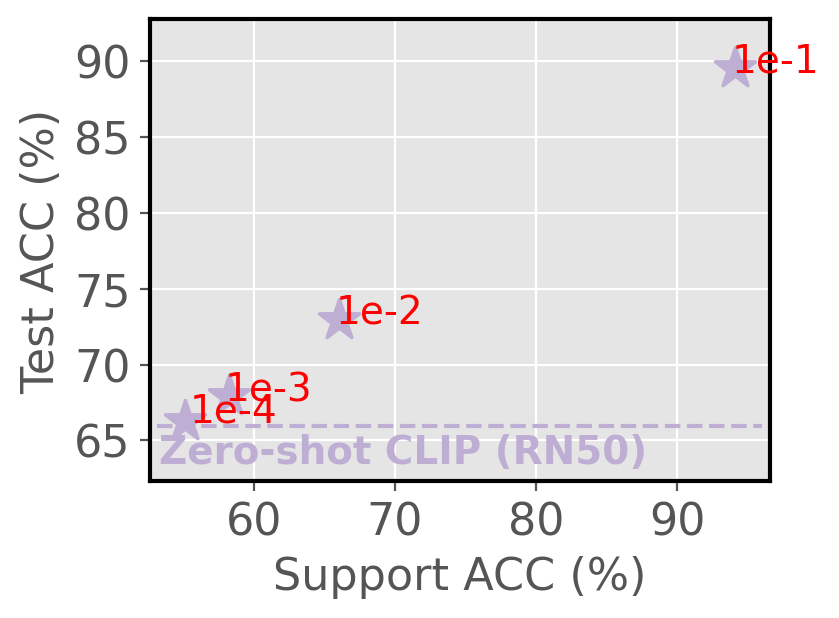} \\

        \scriptsize{\textbf{CLAP - OxfordPets}} &  \scriptsize{\textbf{CLAP - Flowers102}} \\

         \includegraphics[width=0.47\linewidth]{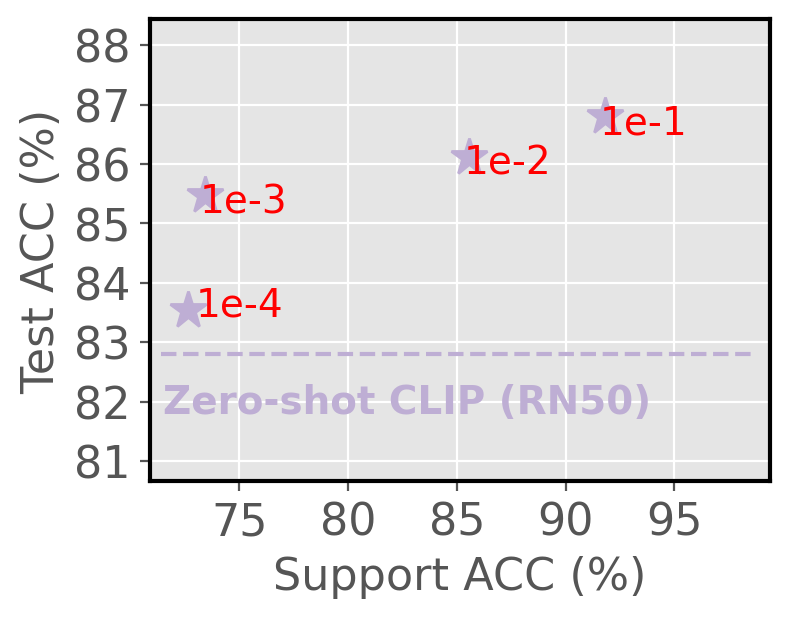} &
         \includegraphics[width=0.47\linewidth]{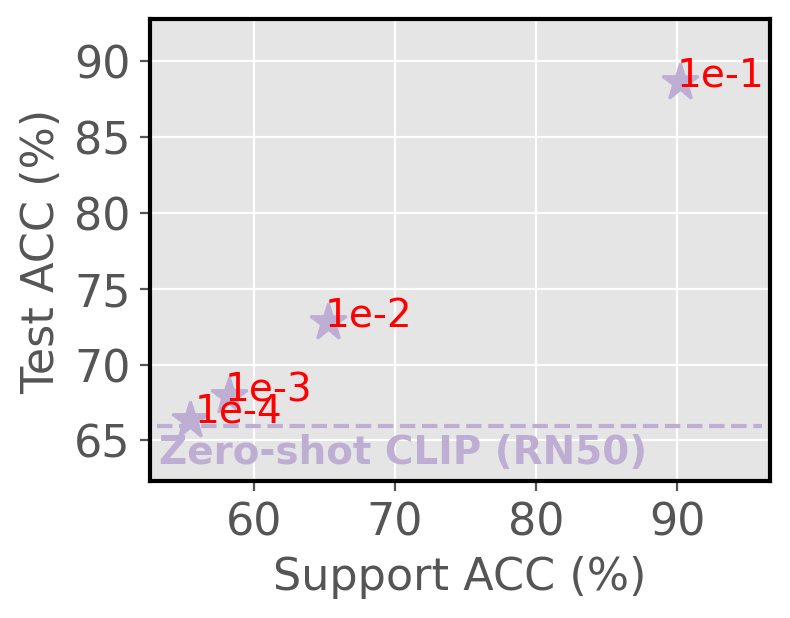} \\
        \end{tabular}
        \caption{\textbf{The trade-off between convergence on support set and generalization for zero-shot initialized adapters}. We depict the performance on the support and test subsets (after training) of zero-shot initialized Linear Probing adapters. Red numbers indicate the initial learning rate used, on the fixed scheduler described in \cref{main:subsection_setup}. Two methods are presented: zero-shot initialized Linear Probe (ZS-LP, \textit{top}, see \cref{main:subsection_lp}), and class adaptive Linear Probe (CLAP, \textit{bottom}, see \cref{main:subsection_class_addaptive_lp}).}
        \label{fig:lr_effect}
    \end{center}\vspace{-5mm}
\end{figure}

To provide further empirical evidence, we now study in \appencref{fig:lr_effect} the performance obtained by a zero-shot initialized Linear Probing (ZS-LP) with a fixed scheduler (see \cref{main:subsection_setup} for details), just varying the initial learning rate. Larger learning rates might produce solutions farther from the initial data points, and vice-versa. In particular, we focus on two popular datasets used for adaptation: OxfordPets \cite{oxfordpets} and Flowers102 \cite{flowers102}. The experimental results show that even for Linear Probing, adjusting the training specifications per task leads to better test generalization. For some datasets, such as OxfordPets, it is beneficial to underfit on the support set (see \appencref{fig:lr_effect}~{top-left}), and thus using smaller learning rates is beneficial. In other cases, such as FLowers102 (see \appencref{fig:lr_effect}~{top-right}), the degradation from fitting to the support set is not observed. Thus, each adaptation task presents its specific behavior. In a few-shot setting, however, only the support set information is available and model selection for a given adapter should rely only on this data. It is worth mentioning that the proposed class-adaptive solution (CLAP) is able to keep robust performance in both cases, using the same training setting across datasets.

\subsection{SoTA methods: is it all about playing with hyperparameters?}
\label{supp:subsection_adapters_hyperparams}

We previously introduced the methodological basis of SoTA adapters in \appensecref{supp:subsection_adapters_method}, and the different hyperparameters they use for model selection. Also, we have introduced in \appensecref{supp:subsection_zs_init} that each adaptation dataset might present different characteristics, and thus the optimum solution might be closer or farther to the zero-shot CLIP initialization.
For instance, the Flowers101 dataset presents a particular behavior, which differs from other datasets (see \appencref{fig:lr_effect}). Interestingly, for this dataset, our cross-shift dataset experiments unveil that large performance drops are experienced in SoTA methods when using the optimum hyperparameters found for other tasks (see \appencref{fig:cross-shift_supplementary}). In the following, we provide observational evidence that these methods adjust specific hyperparameter values per dataset, using prior knowledge from the test subset, which is unrealistic in practice.

\noindent \textbf{CLIP-Adapter} \cite{gao2021clip}. While an official implementation of the training code is not available, authors explicitly claim in the paper that: ``\textit{We perform hyperparameter searching over different value selections of $\alpha$ for each dataset and report the best performance among all searching spaces.}" In addition, we only could replicate their results when directly adjusting the learning rate (swept among $\{10^{-1}, 10^{-2}, 10^{-3}\}$) and residual ratio (searching values are $\alpha_\mathrm{r}=$$\{0.2, 0.4, 0.6, 0.8, 1\}$) on a grid search at the test subset.

\noindent \textbf{TIP-Adapter} \cite{zhang2021tip}. The absence of any details in the original publication regarding model selection strategies or the use of validation subsets leaves the GitHub repository as the only available documentation of the official implementation. In this repository, the authors claim (see Issue $\#13$)\footnote{\label{note1}Recommendation provided in the official project repository: \url{https://github.com/gaopengcuhk/Tip-Adapter/issues/13}.}: ``\textit{The alpha and beta are both set to 1 as the tuning baseline. The alpha weighs the importance of CLIP-pre-trained and few-shot knowledge. If the few-shot domain has a large gap to pre-trained data (general images, just like ImageNet), alpha is better to be larger than 1.}" This suggests no explicit strategy for model selection exists. In addition, the official implementation contains a hyperparameter search function that takes as input the test subset in the case of ImageNet and a large validation subset for other tasks. It is worth mentioning that the grid search boundaries per hyperparameter also depend on each specific task. For instance, the $\alpha_{\mathrm{tipA}}$ parameter for ImageNet is searched between $[1.17, 7]$, and for Flowers102 dataset the target range is $[10, 50]$, not presenting an overlapping at all. Interestingly, $\alpha_{\mathrm{tipA}}$ controls the relative importance of the vision logits, and larger weight values are searched on Flowers102, a dataset which, as we show in \appencref{fig:lr_effect}, benefits from diverging from the zero-shot initialization. These details suggest that the hyperparameters for TIP-Adapter methods are fixed assuming prior knowledge of the test subset for each particular task.

\noindent \textbf{Task Residual Learning} \cite{zhang2021tip}. As previously described, TaskRes is equivalent to a zero-shot initialized Linear Probing, and contains an $\alpha$ parameter that regulates the learning rate per dataset. It is worth mentioning that the implementation details describe the use of different learning rates for ImageNet adaptation ($\eta=2\cdot10^{-4}$), and for other tasks ($\eta=2\cdot10^{-3}$), as well as different epochs depending on the number of shots. In addition, it is stated that ``\textit{By default, the scaling factor $\alpha$ is set to 0.5 for all datasets except for Flowers102 using 1}". This detail is especially relevant, as it suggests the access to prior knowledge to test performance. Again, a larger adaptation to the support samples is used for the Flowers102 task, which aligns with the low transferability of the hyperparameters set on this task to other datasets in \cref{fig:cross-shift_supplementary}, as well as with the longer convergence on test performance observed in this dataset (\appencref{fig:lc}).

\subsection{Choosing hyperparameters for a validation-free benchmark}
\label{supp:subsection_details_adapters}

In this work, we seek to provide a realistic protocol for comparing few-shot vision-language adapters. In this setting, we assume access to only the available support samples, and no additional validation examples are used. Next, we describe the implementation details of the different baselines and the motivation for the use of particular hyperparameter values. 

For CLIP-Adpater \cite{gao2021clip}, we set the hyperparameter $\alpha$ to $0.2$ for all datasets, as it is the best value found on ImageNet evaluation in the original paper. The TIP-Adapter \cite{zhang2021tip} umbrella gathers two methods: training-free, and a trainable version, \ie, TIP-Adapter(f), in which the support samples embeddings are updated. For both methods, we set $\beta$ and $\alpha$ to $1$, as recommended in the official repository (see \appencref{note1}) . For TaskRes \cite{yu2023task}, we only used as baseline its enhanced version, referred to as TaskRes(e)\footnote{The training code for TaskRes(e) base is not provided in the official implementation (\url{https://github.com/geekyutao/TaskRes}) and might contain specific tuning that indirectly resorts to the test set. Authors uniquely share the enhanced weights, and the lack of specific implementation details might produce unfair comparisons. The only information available in the manuscript is: \textit{``... enhanced base classifier obtained by tuning the text projection layer of CLIP on the target task before starting our task residual tuning ... The aforementioned enhanced base classifier is tuned for 50 epochs".}}, which updates the projection layer of the text encoder. The reason for not using the base version of TaskRes is motivated by our findings that suggest that this method is equivalent to a Linear Probe tuning with zero-shot initialization, and a specific learning rate scaling for each dataset (see \appensecref{supp:subsection_adapters_method}). We set $\alpha$ to $0.5$ in TaskRes(e) since this is the value used in the majority of the tasks in the original publication. Finally, we included Cross-Modal adapters \cite{lin2023crossmodal}, in particular the Linear Probing version, which does not require special hyperparameter tuning. To avoid using an empirical grid search for weight decay, we implicitly applied an $\ell_2$-normalization over the weights during training, which provided a better performance on our ablation experiments (see \cref{main:subsection_ablation}). All methods are trained using the same general optimizer and scheduler as our proposed methods, which showed proper convergence on the support set, and all baselines employ the same text prompts for each dataset.

\subsection{Trainable parameters}
\label{supp:subsection_parameters}

\begin{figure}[h!]
    \begin{center}
         
         \includegraphics[width=0.95\linewidth]{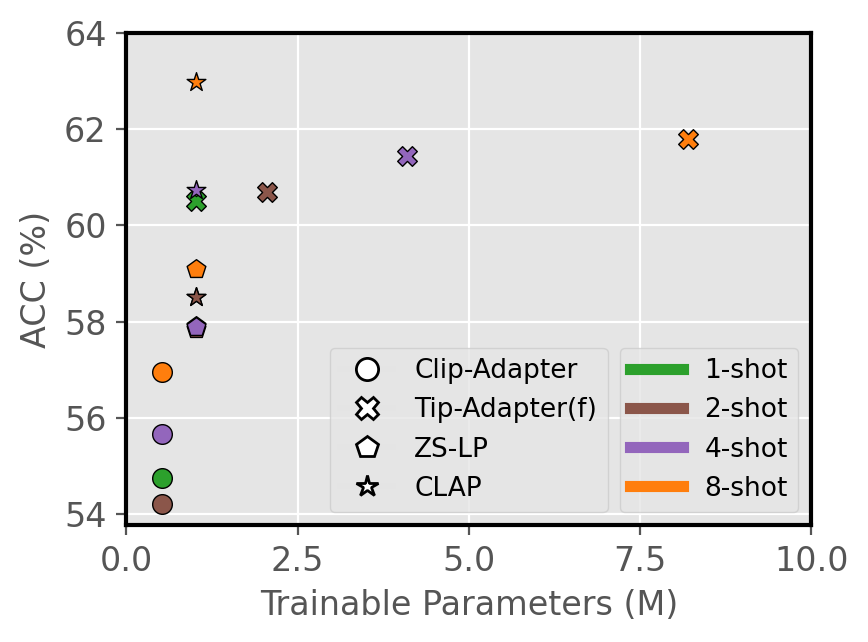}
        \caption{\textbf{Trade-off between number of shots, trainable parameters, and adaptation performance.} The test accuracy is presented with respect to the number of trainable parameters for CLIP-Adapter \cite{gao2021clip}, TIP-Adapter(f) \cite{zhang2021tip}, and the two proposed solutions in this work: a revisited Linear Probing (\mbox{ZS-LP}, see \cref{main:subsection_lp}), and a class-adaptive Linear Probing (CLAP,  see \cref{main:subsection_class_addaptive_lp}). Results were obtained for 1 to 8 shots in the ImageNet dataset.}
        \label{fig:parameters}
    \end{center}
\end{figure}

Efficient transfer learning ought to exploit limited supervision during adaptation while being efficient in the number of trainable parameters. We depict in \appencref{fig:parameters}, a visualization of the trade-off between the number of trainable parameters and test performance of relevant prior methods, and the proposed class adaptive Linear Probing (CLAP, see \cref{main:subsection_class_addaptive_lp}). All results are obtained in the validation-free protocol, using the implementation details described in \cref{main:subsection_setup}. While CLIP-Adapter is a specially lightweight solution, the obtained performance is limited with respect to the proposed method (CLAP), and even a well-initialized Linear Probing (ZS-LP). On the other hand, TIP-Adapter largely increases the number of tunable weights with the number of shots, which questions its transferability to other tasks, such as dense image segmentation, in which each pixel prototype would constitute an individual parameter. In contrast, CLAP just introduces a negligible set of additional trainable multipliers - one per class - over a Linear Probing solution, which considerably enhances its performance.

\section{Penalty functions for ALM: axioms}
\label{supp:subsection_penalties_axioms}

In this section, we provide the requirements for a penalty function in the Augmented Lagrangian Multiplier (ALM) method, detailed in \cref{main:subsection_class_addaptive_lp}.

A function $P : \real \times \real_{++} \times \real_{++} \rightarrow \real $ is a Penalty-Lagrangian function such that $P'(z, \rho, \lambda) \equiv \frac{\partial}{\partial z}P(z, \rho, \lambda)$ exists and is continuous for all $z \in \real$, $\rho \in \real_{++}$ and $\lambda \in \real_{++}$. In addition, a penalty function $P$ should satisfy the following four axioms \cite{birgin2005numerical}:
\begin{itemize}[leftmargin=*,label={}]
    \item {\bf Axiom 1:} $P'(z, \rho, \lambda) \geq 0 \quad \forall z\in\real,  \rho \in \real_{++}, \lambda \in \real_{++}$
    
    \item {\bf Axiom 2:} $P'(0, \rho, \lambda) = \lambda  \quad \forall \rho \in \real_{++}, \lambda \in \real_{++}$
    
    \item {\bf Axiom 3:} If, for all $j\in\N, \; \lambda^{(j)} \in [\lambda_\text{min},\lambda_\text{max}]$, where $0 < \lambda_\text{min} \leq \lambda_\text{max} < \infty$, then:
    
    \qquad $\lim\limits_{j\rightarrow\infty}\rho^{(j)}=\infty$ and $\lim\limits_{j\rightarrow\infty}y^{(j)}>0$ imply that $\lim\limits_{j\rightarrow\infty}P'(y^{(j)}, \rho^{(j)}, \lambda^{(j)})=\infty$
    
    \item {\bf Axiom 4:} If, for all $j\in\N, \; \lambda^{(j)} \in [\lambda_\text{min},\lambda_\text{max}]$, where $0 < \lambda_\text{min} \leq \lambda_\text{max} < \infty$, then: 
    
    \qquad$\lim\limits_{j\rightarrow\infty}\rho^{(j)}=\infty$ and $\lim\limits_{j\rightarrow\infty}y^{(j)}<0$ imply that $\lim\limits_{j\rightarrow\infty}P'(y^{(j)}, \rho^{(j)}, \lambda^{(j)})=0$.
\end{itemize} 

\noindent The first two axioms guarantee that the derivative of the Penalty-Lagrangian function $P$ \wrt $z$ is positive and equals to $\lambda$ when $z=0$. The last two axioms guarantee that the derivative tends to infinity when the constraint is not satisfied, and zero otherwise. 

\section{Supplementary experimental details}
\label{supp:section_experiments}

\subsection{Additional setup information}
\label{supp:subsection_datasets}

\begin{table*}[h!]
\caption{\textbf{Summary of datasets details.} Detailed description of the 11 datasets used to validate the SoTA few-shot adapters of VLMs, and 4 ImageNet shifts employed to evaluate the generalization capabilities of those. Also, handcrafted prompts used to obtain the zero-shot predictions and prototypes are detailed. These are the same ones used in relevant prior literature on this topic \cite{yu2023task,zhou2022coop,lin2023crossmodal}.}
\label{datasets}
\centering
\scriptsize
\begin{tabular}{lccll}
\toprule
\multicolumn{1}{c}{Dataset} &
  Classes &
  \begin{tabular}[c]{@{}c@{}}Splits\\ Train / Val / Test\end{tabular} &
  \multicolumn{1}{c}{Task} &
  \multicolumn{1}{c}{\begin{tabular}[c]{@{}c@{}}Prompt\\  Templates\end{tabular}}\\
\midrule
ImageNet \cite{deng2009imagenet}        & 1000 & 1.28M / - / 50,000      & Natural objects recognition               & \multirow{5}{*}{\begin{tabular}[l]{@{}l@{}} [``itap of a [CLS]" , ``a bad photo of a [CLS]" , \\  ``a origami of [CLS]" , ``a photo of the large [CLS]" , \\ ``a [CLS] in a video game" , ``art of the [CLS]" , \\ ``a photo of the small [CLS]" , ``a photo of a [CLS]"] \end{tabular}} \\
ImageNet-V2 \cite{imagenetV2}           & 1000 & - / - / 10,000          & Natural objects recognition               &  \\
ImageNet-Sketch \cite{imagenetSketch}   & 1000 & - / - / 50,889          & Sketch-style image classification         &  \\
ImageNet-A \cite{imagenet_a}            & 200  & - / - / 7,500           & Natural objects recognition               &  \\
ImageNet-R \cite{imagenet_r}            & 200  & - / - / 30,000          & Natural objects recognition               &  \\
\midrule
Caltech101 \cite{caltech}               & 100 & 4,128 / 1,649 / 2,465      & Natural objects classification          & [``a photo of a [CLS]"] \\
OxfordPets \cite{oxfordpets}            & 37  & 2,944 / 736 / 3,669        & Pets classification (fine-grained)      & [``a photo of a [CLS], a type of a pet"] \\
StanfordCars \cite{stanfordcars}        & 196 & 6,509 / 1,635 / 8,041      & Cars classification (fine-grained)      & [``a photo of a [CLS]"] \\
Flowers102 \cite{flowers102}            & 102 & 4,093 / 1,633 / 2,463      & Flowers classification (fine-grained)   & [``a photo of a [CLS], a type of flower" ]\\
Food101 \cite{food101}                  & 101 & 50,500 / 20,200 / 30,300   & Foods classification (fine-grained)     & [``a photo of a [CLS], a type of food"] \\
FGVCAircraft \cite{aircraft}            & 100 & 3,334 / 3,333 / 3,333      & Aircrafts classification (fine-grained) & [``a photo of a [CLS], a type of aircraft"] \\
SUN397 \cite{sun397}                    & 397 & 15,880 / 3,970 / 19,850    & Scenes classification                   & [``a photo of a [CLS]"] \\
DTD \cite{dtd}                          & 47  & 2,820 / 1,128 / 1,692      & Textures classification                 & [``[CLS] texture"] \\
EuroSAT \cite{eurosat}                  & 10  & 13,500 / 5,400 / 8,100     & Satellite image classification          & [``a centered satellite photo of [CLS]"] \\
UCF101 \cite{ucf101}                    & 101 & 7,639 / 1,898 / 3,783      & Recognition of actions                  & [``a photo of a person doing [CLS]"] \\
\bottomrule
\end{tabular}
\end{table*}

\paragraph{Datasets details.} In our main text, we introduce the datasets employed to evaluate the proposed methods and establish comparisons with relevant literature on the few-shot adaptation of CLIP-based models. In \appencref{datasets}, we introduce the specific details of each dataset, including the number of categories, test partition size, and particular tasks.

\paragraph{Text prompt templates.} We followed the same hand-crafted templates as relevant prior literature of efficient transfer learning for the 11 datasets. Concretely, we followed CoOp \cite{zhou2022coop}, TIP-Adapter \cite{zhang2021tip} TaskRes \cite{yu2023task}, and CrossModal hand-crafted version \cite{lin2023crossmodal}. These prompts are composed of an ensemble of 8 different templates for Imagenet-like datasets, and 1 template for the others, which are depicted in \appencref{datasets}. It is worth mentioning that fine-tuning methods used for the benchmark in \cref{main:subsection_results} (LP-FT \cite{LPFT}, FLYP \cite{FLYP}) use a larger set of 80 prompt templates for ImageNet-like datasets, although these are not usually used in the efficient transfer learning literature.

\subsection{Results: supplementary details}
\label{supp:subsection_results}

\begin{figure*}[h!]
    \begin{center}

         \includegraphics[width=.31\linewidth]{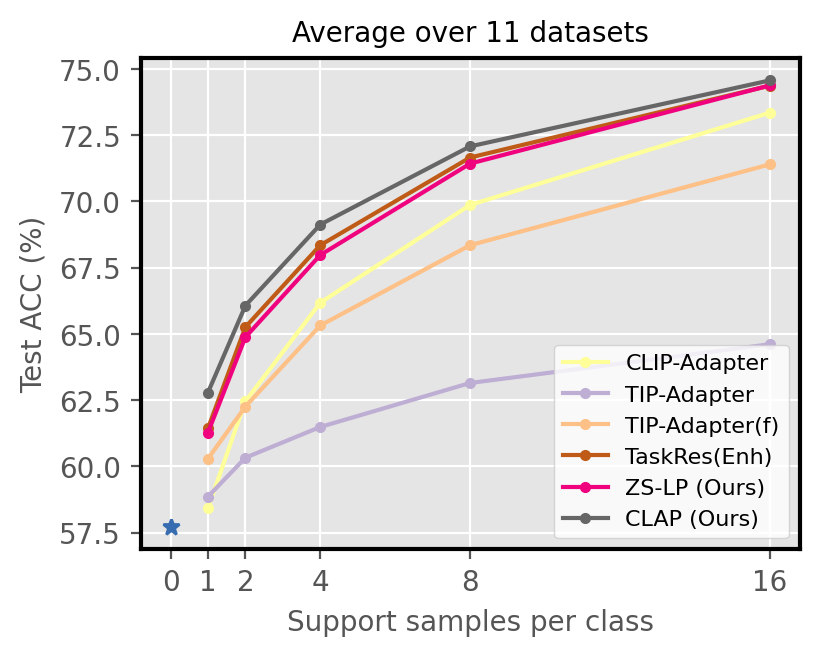}
         \includegraphics[width=.31\linewidth]{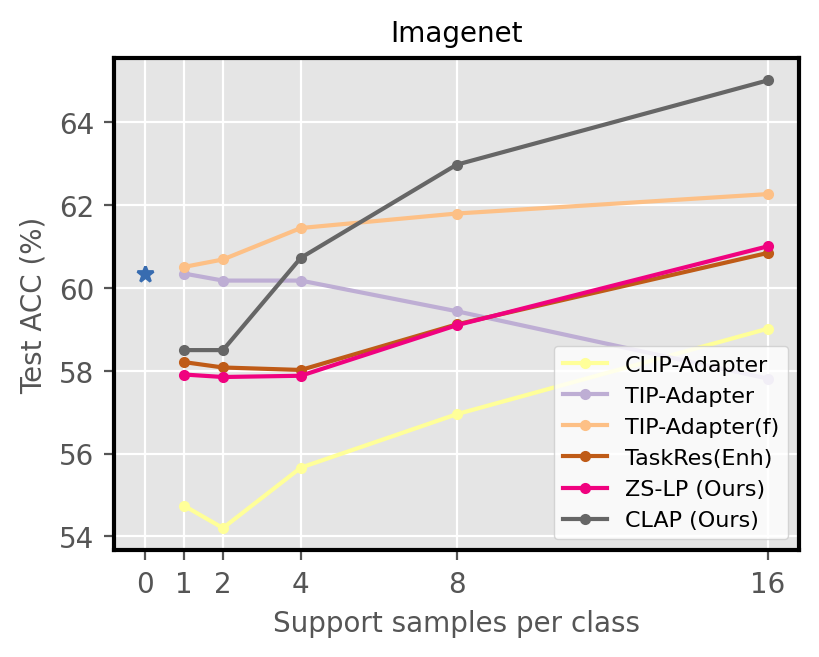}
         \includegraphics[width=.31\linewidth]{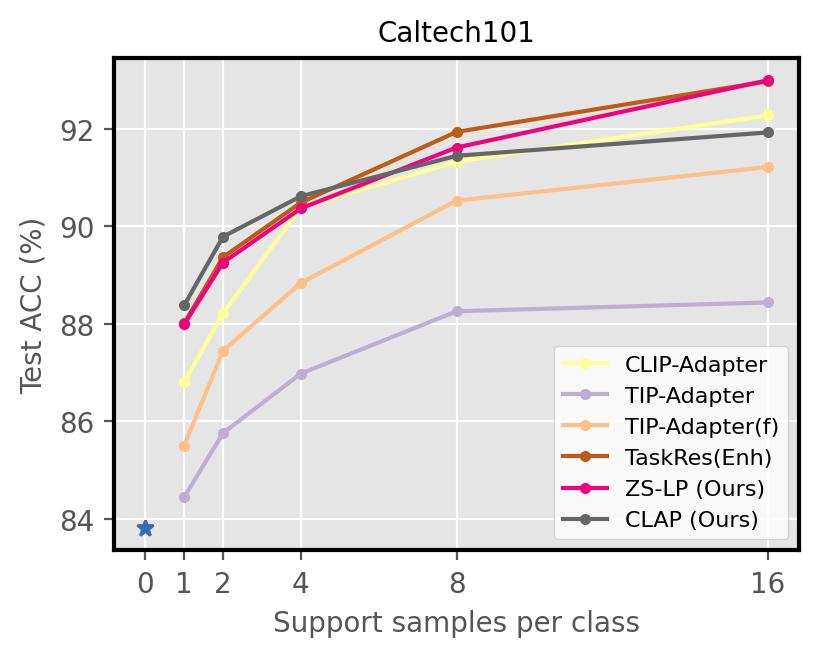}

         \includegraphics[width=.31\linewidth]{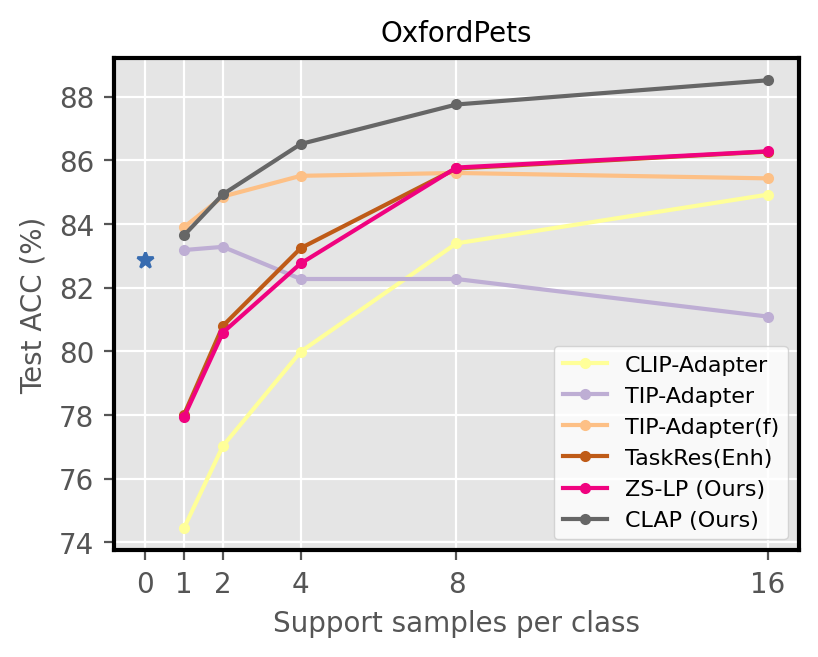}
         \includegraphics[width=.31\linewidth]{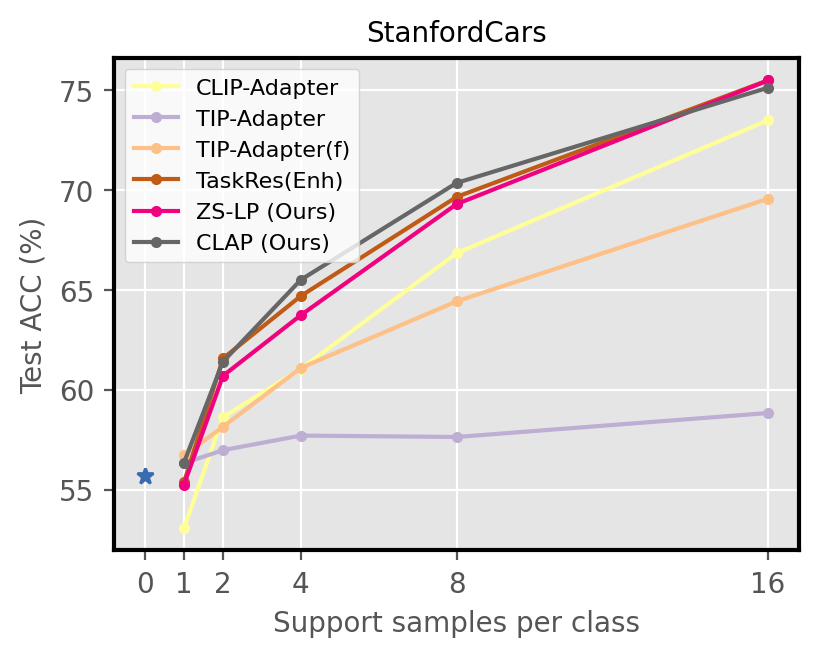}
         \includegraphics[width=.31\linewidth]{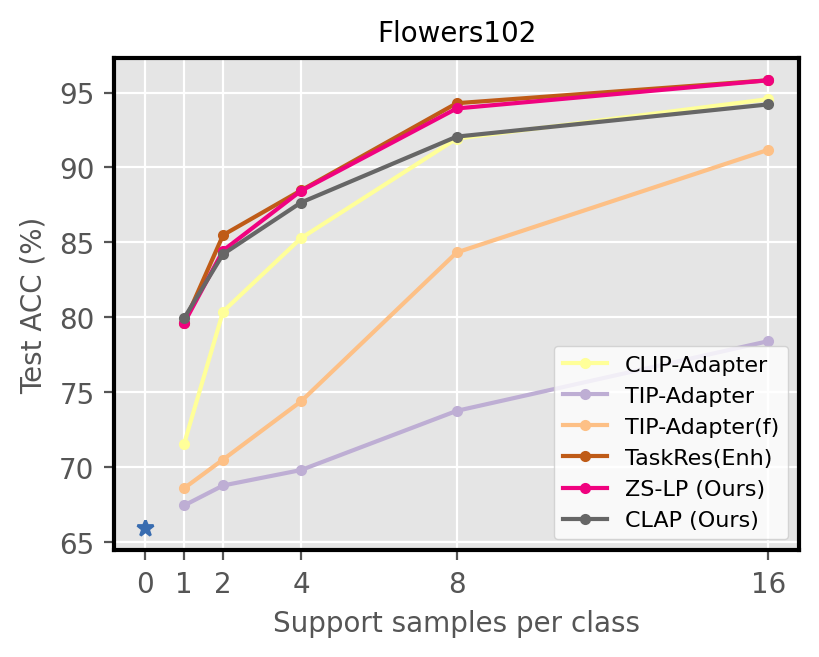}

         \includegraphics[width=.31\linewidth]{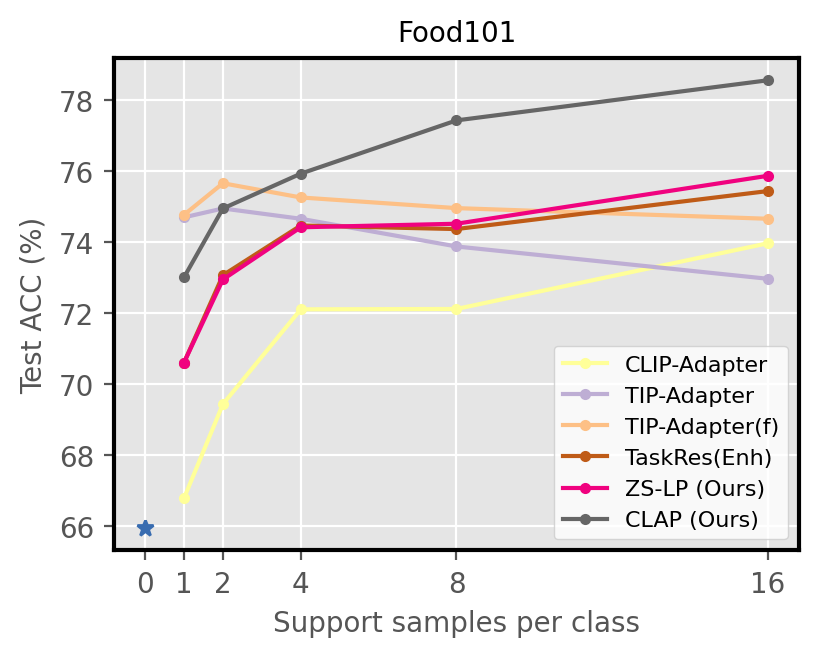}
         \includegraphics[width=.31\linewidth]{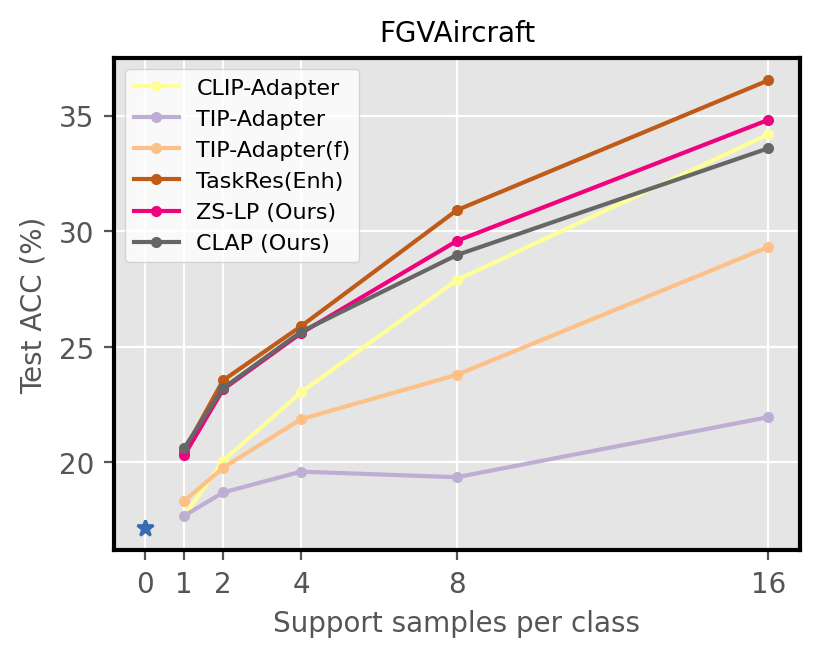}
         \includegraphics[width=.31\linewidth]{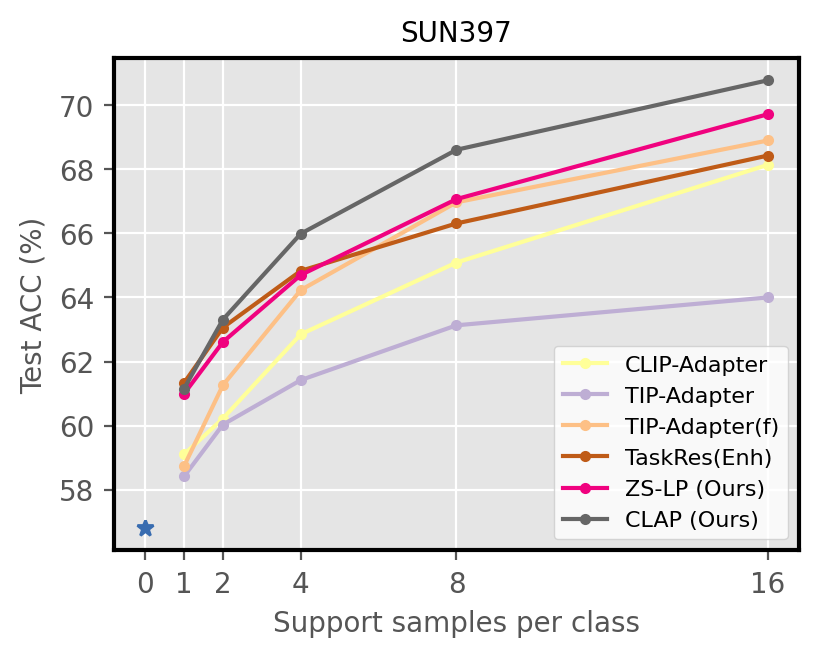}

         \includegraphics[width=.31\linewidth]{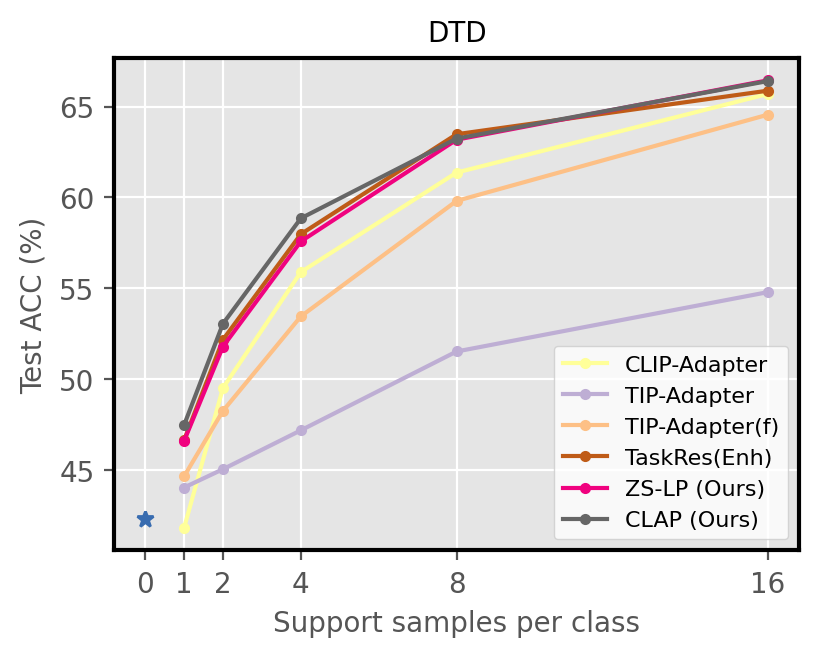}
         \includegraphics[width=.31\linewidth]{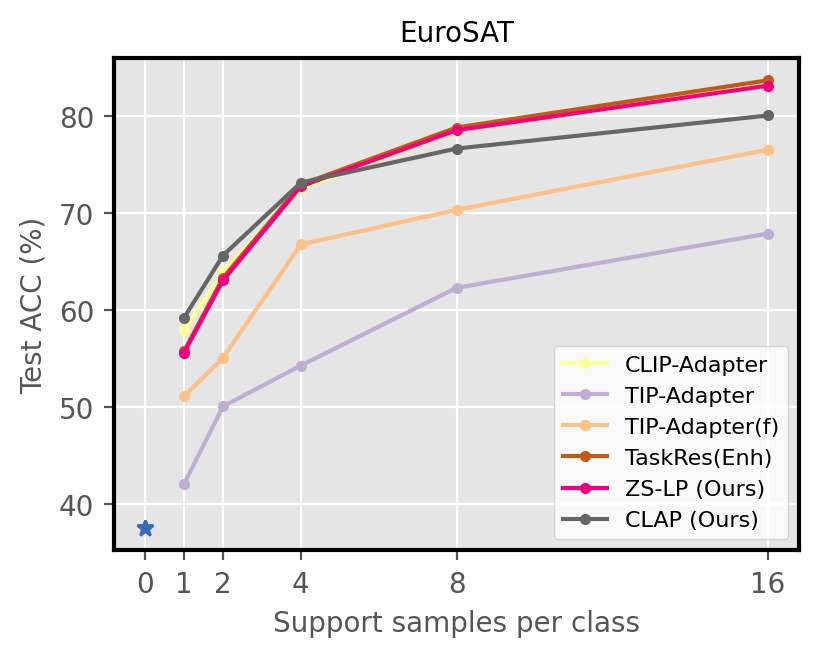}
         \includegraphics[width=.31\linewidth]{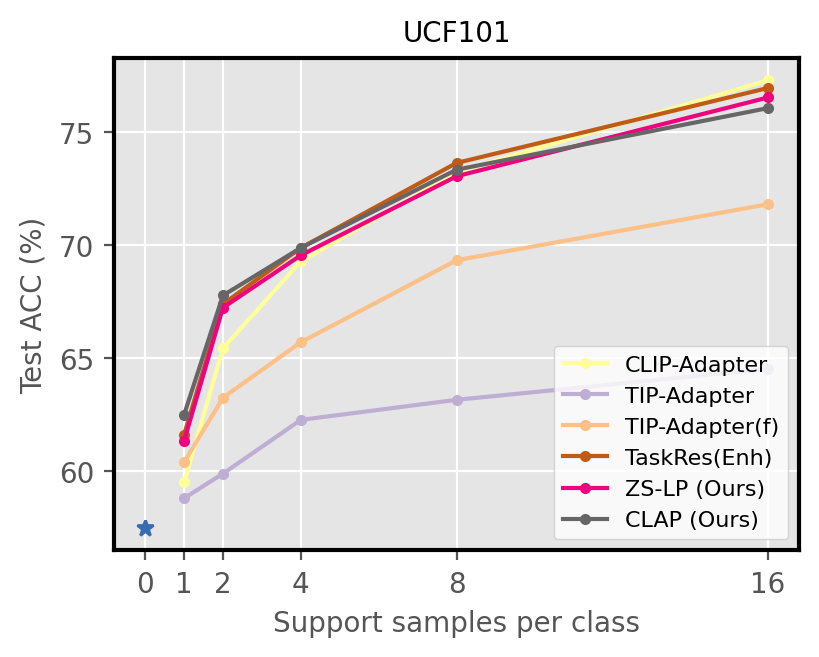}

        \caption{\textbf{Efficient transfer learning Results}. Performance comparison of relevant literature, and the proposed methods for few-shot efficient transfer learning from ResNet-50 CLIP to 11 downstream datasets, using from 1 to 16 shots per class. Average results are depicted in te top-left corner. Full numerical results are introduced in \appencref{numerical_etl_comparison}.}
        \label{fig:etl}
    \end{center}\vspace{-5mm}
\end{figure*}

\paragraph{Efficient transfer learning.} We provide detailed numerical results for the few-shot adaptation experiments using relevant baselines and the proposed methods in \appencref{numerical_etl_comparison}, which extend the values reported in \cref{table_main_few_shot_results}. Furthermore, we also depict visual curves of the performance with respect to the number of shots employed by each method in \appencref{fig:etl}.

\paragraph{Domain generalization.} We show in the main manuscript the domain generalization results using adapters adjusted to ImageNet and evaluated on out-of-distribution shifts (\ie, ImageNet variants). These results are obtained using ResNet-50 and ViT-B/16 CLIP backbones. In the following, we introduce detailed results per dataset, and two additional backbones: ResNet-101 and ViT-B/32, whose results are reported in \cref{results_generalization_supp}. We can observe that the results using these additional backbones hold the conclusions elucidated in the main manuscript. In particular, relevant prior methods such as CLIP-Adapter \cite{gao2021clip} and TIP-Adapter \cite{zhang2021tip} struggle to generalize properly when their hyperparameter setting is held on different backbones than the one used for development, ResNet-50. This is especially the case for Transformer backbones, such as ViT-B/32, which suggests again that existing adapter methods need special care for model selection across each dataset.

\paragraph{Finetuning (FT) vs. efficient transfer learning (ETL), beyond few-shots.} Fine-tuning a whole vision encoder to downstream tasks using a few-shot training subset has been historically less favored compared to efficient transfer learning strategies, due to the tendency of FT methods to overfit to the new data, and thus generalizing poorly. Nevertheless, a relevant core of recent literature for VLMs adaptation \cite{LPFT,FLYP,WiSE} is showing promising results on this task. As stated in the main body of the paper, this is due to (i) using a few-shot validation dataset, with which they early-stop the training, and (ii) employing small learning rates to not deviate from a good initialization. Nevertheless, if compared properly in the low data regime, \ie, using 4 shots for training and 4 samples per class for validation, and allowing ETL methods that do not require a validation set to use all samples for training, then ETL still seems to yield competitive performance, being a much more computationally-efficient solution. The results previously presented in the main body of the manuscript (see \cref{ft_vs_etl_main}) support these observations. 

\begin{figure}[h!]
    \begin{center}
         
         \includegraphics[width=0.95\linewidth]{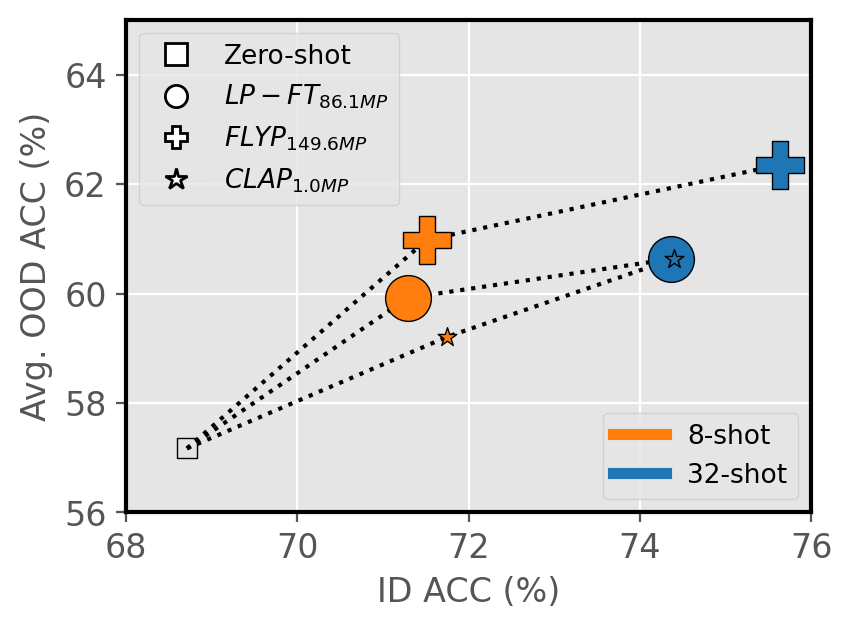}
        \caption{\textbf{Finetuning (FT) vs. efficient transfer learning (ETL), performance and trainable parameters}. We compare the generalization performance of relevant full fine-tuning methods, \ie, LP-FT \cite{LPFT} and FLYP \cite{FLYP}, and the proposed efficient transfer learning method CLAP (see \cref{main:subsection_class_addaptive_lp}), trained on ImageNet and evaluated on OOD shifts. Relative point size is illustrated as the log ratio of the number of tunable parameters of each method with respect to CLAP. MP: millions of parameters.}
        \label{fig:parameters_ft}
    \end{center}
\end{figure}

We now extend this comparison to a scenario in which more data is available. Concretely, a 32-shot scenario, where FT methods use half of it for validation. It is worth mentioning that this experimental setting on ImageNet required 32,000 images (16,000 for validation), which might be hardly considered a few-shot learning protocol. We introduce specific results using 32-shot for ImageNet and its distributional shifts for relevant baselines and the proposed methods in \appencref{ft_etl_results_supp}. It is worth mentioning that FT methods use 16-shots for training and another 16-shots for validation. In addition, we present in \appencref{fig:parameters_ft} a study of the performance evolution with respect to the number of shots of relevant FT methods with an increasing number of parameters. More concretely, we include LP-FT \cite{LPFT}, which completely fine-tunes the CLIP's vision backbone, and FLYP \cite{FLYP}, which trains both vision and text encoders. We compare these results to the proposed class adaptive Linear Probing (CLAP), which only adjusts the classifier head, and thus brings a negligible computational overhead compared to LP-FT and FLYP. In the 32-shot setting, CLAP shows competitive performance compared to methods that adjust the vision encoder entirely, such as FT, LP-FT \cite{LPFT}, and WiSE-FT \cite{WiSE}, for both in-distribution and out-of-distribution datasets, while adjusting only the linear classification head. Only FLYP \cite{FLYP}, which requires fine-tuning both vision and text encoders, outperforms CLAP using 32,000 images, and by a small margin: $1.2\%$ in ID, and $1.6\%$ in OOD. Nevertheless, this comes at the cost of adjusting the entire CLIP model, which entails a non-negligible computational overhead, making this method an inefficient approach in low-resource scenarios. Note that CLAP is two orders of magnitude lighter than FLYP. In addition, CLAP does not exhibit signs of performance saturation (as LP-FT, for example) with an increase in the number of shots (see \appencref{fig:parameters_ft}).

\begin{table}[h!]
\caption{\textbf{Finetuning (FT) vs. efficient transfer learning (ETL), beyond few-shots.} Benchmark for the not-so-low data regime, \ie, 32 shots for each class. FT methods (above the dashed line) are trained with 16 shots and early-stopped using a validation set containing 16 shots. WiSE-FT and FLYP use weight ensembling as proposed in \cite{WiSE}, and therefore, find the best \textit{mixing coefficient} $\alpha$ using the validation set. On the other hand, ETL methods (below the dashed line) are trained using all the 32 shots given. All methods use ViT-B/16 as CLIP backbone.}
\label{ft_etl_results_supp}
\centering
\scriptsize
\begin{tabular}{lccccccc}
\toprule
\multicolumn{1}{c}{\multirow{2}{*}{Method}} & \multicolumn{1}{c}{Source} & \multicolumn{5}{c}{Target} \\ \cline{3-8} 
\multicolumn{1}{c}{} & \multicolumn{1}{c}{Imagenet} & \multicolumn{1}{c}{-V2} & \multicolumn{1}{c}{-Sketch} & \multicolumn{1}{c}{-A} & \multicolumn{1}{c}{-R} & \multicolumn{1}{c}{Avg.} \\ 
\midrule
FT                                        & 71.86 &	64.15 &	47.97 & 48.23 &	75.96 & 59.08 \\
LP-FT \cite{LPFT}                         & 74.36 &	66.43 &	49.35 &	49.84 &	76.89 &	60.63 \\
WiSE-FT \cite{WiSE}                       & 73.06 & 65.70 & 50.03 & 51.04 & 78.22 & 61.25 \\
FLYP \cite{FLYP}                          & \textbf{75.63} & 68.17 & 50.66 & 52.09 & 78.49 & \textbf{62.35} \\
\hdashline\noalign{\vskip 0.5ex}
Zero-Shot                                 & 68.71 & 60.76 & 46.18 & 47.76 & 73.98 & 57.17 \\
LP                                        & 67.40 & 56.43 & 31.71 & 31.92 & 51.04 & 42.71 \\
\rowcolor{Gray}ZS-LP                                     & 71.53 & 56.59 & 40.84 & 41.41 & 67.98 & 51.71 \\
\rowcolor{Gray}CLAP                    & 74.40 & 66.05 & 49.16 & 49.82 & 77.52 & 60.64 \\
\bottomrule
\multicolumn{8}{l}{*Specific numbers for FT, LP-FT, WiSE-FT, and FLYP are retrieved from \cite{FLYP}.}\\
\end{tabular}
\end{table}

\subsection{Supplementary ablation experiments}
\label{supp:subsection_ablation}

\begin{figure}[h!]
    \begin{center}
         
         \includegraphics[width=0.98\linewidth]{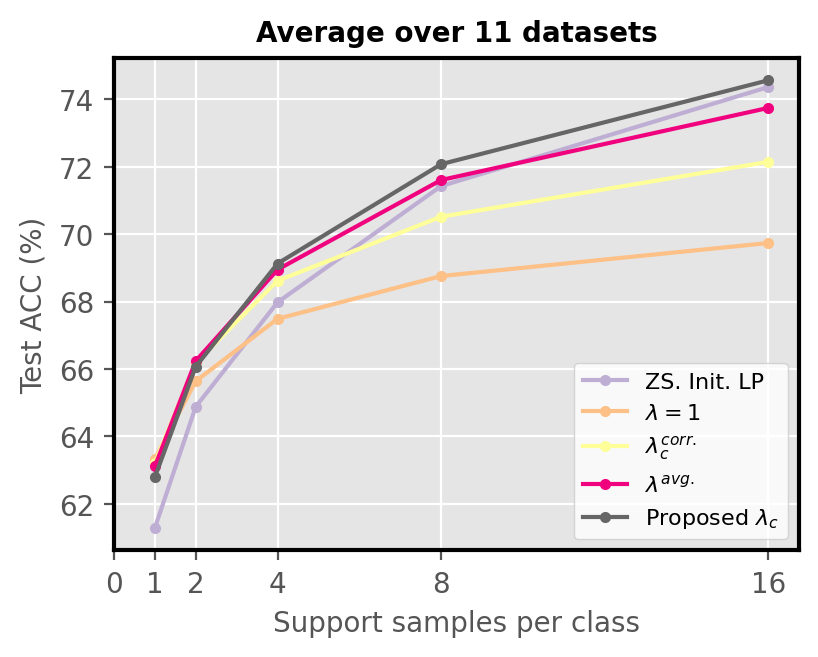}

        \caption{\textbf{Ablation study on different alternatives to compute initial $\blambda$ in \cref{eq:loss_problem_to_solve}.} The average performance over 11 datasets is reported.}
        \label{fig:ablation_lambda}
    \end{center}
\vspace{-3mm}
\end{figure}

\paragraph{Distilling reliable knowledge.} We now study the effect of resorting to a class-adaptive constrained formulation in the proposed CLAP. In particular, we further assess the benefits of using our class-dependent adaptive scaling ($\sblambda$) of the imposed constraint, which is initialized on the performance of the zero-shot CLIP prototypes on the support set. To do so, we take as baseline different empirically-set multipliers baselines. First, we explore a homogeneous weight for all classes, such that $\blambda=\boldsymbol{1}$. Furthermore, we aim to disentangle two different effects that $\sblambda$ might have: (i) changing the overall relative importance of the constraint term with respect to the cross-entropy term in \cref{eq:loss_problem_to_solve}; and, (ii) providing the capability of capturing class dependent prior knowledge from the pre-trained model. Thus, we further include two alternative ways of computing $\blambda$ in our ablation study: (i) a constant version of the constraint formulation, in which all multipliers are set to a constant $\lambda^{avg}=\frac{1}{C}\sum_{c}\slambda_{c}$, \ie, the average importance of the constraint; and, (ii) an importance corrected version of the constrained formulation, $\blambda^{corr}=\sblambda / \lambda^{avg}$, such that $\frac{1}{C} \sum_{i}\lambda^{corr}_{i} = 1$. The average performance over 11 datasets for the few-shot data regime is shown in \appencref{fig:ablation_lambda}, whereas the full numerical results per dataset are presented in \appencref{numerical_ablation_constraint}.

In the special case of 1-shot or 2-shot, adaptive computation of $\blambda$ may lead to a slight overfitting to the information provided by the few samples, even though the performance gap compared to other strategies is minimal. The more shots provided, the less noisy the information will be and computing class-dependent adaptive importance will prove more useful. Looking at the plots, except for the 1-shot setting, using $\blambda = \boldsymbol{1}$ is always suboptimal. In the 2-shot scenario, using $\lambda^{avg.}$ for all classes yields the best results, although close to the performance of the proposed $\sblambda$. Generally, when the number of labeled samples per class is particularly low, the proposed $\sblambda$ is not the best-performing option. However, it proves superior to other alternatives when 4 or more shots are given, and the gap widens as the number of shots increases. It is worth noting that the non-constrained (\mbox{ZS-LP}) version's performance approaches the constrained version in the 16-shot setting, but it underperforms in the lower data regimes, which demonstrates the effectiveness of our formulation across the different regimes.

\begin{table}[h!]
\caption{\textbf{Augmented Lagrangian multiplier optimization.} We present ablation experiments that motivate the use of only one outer iteration in \cref{eq:lagrangian_alm}, which prevents overfitting on the support samples, due to the absence of a validation subset in the realistic few-shot scenario. $K$ denotes the number of shots.}
\vspace{-2mm}
\label{update_multipliers}
\resizebox{0.47\textwidth}{!}{
\centering
\begin{tabular}{lccc}
\toprule
\multicolumn{1}{c}{\multirow{1}{*}{Method}} & $\rms K=1$     & $\rms K=2$    & $\rms K=4$ \\
\midrule
\rowcolor{Gray}ZS-LP             & 61.28 & 64.88 & 67.98 \\
\hdashline\noalign{\vskip 0.5ex}
CLAP - Full outer loop                              & 61.97$_{(+0.7)}$\textcolor{blue}{$\uparrow$}   & 63.25$_{(-1.6)}$\textcolor{red}{$\downarrow$}  & 63.83$_{(-4.2)}$\textcolor{red}{$\downarrow$} \\
\rowcolor{Gray}CLAP - 1 outer loop (\textit{Ours})  & 62.79$_{(+1.5)}$\textcolor{blue}{$\uparrow$}   & 66.07$_{(+1,2)}$\textcolor{blue}{$\uparrow$}   & 69.13$_{(+1.2)}$\textcolor{blue}{$\uparrow$}  \\
\bottomrule
\end{tabular}
}
\vspace{-5mm}
\end{table}

\paragraph{Updating the Lagrangian multipliers beyond the first iteration.} The proposed class adaptive Linear Probing (CLAP) is based on an adaptation of the general Augmented Lagrangian Multiplier method, which learns the multiplier weights per class accounting for the particular difficulty of each category. Given the strict few-shot setting used, we propose to use the support samples to validate the satisfaction of the constrained problem. We hypothesized, however, that doing this could increase the risk of overfiting and proposed to stop after one single iteration of the outer optimization iteration in \cref{eq:lagrangian_alm} (i.e., the step where the penalty multiplies $\blambda$ are updated).
In this section, we provide the empirical evidence for this hypothesis, which validates our choice. In particular, we also perform the adaptation using both inner and outer iterations during the whole adaptation process. To do so, at each epoch, the Lagrangian multipliers are updated following \cref{eq:lambda_update}. Regarding the penalty multipliers $\brho$, these are initialized per class to the initial penalty value, and for each epoch, their value is fixed after updating the penalty multipliers to the resulting penalty after the inner iteration. Results are reported in \cref{update_multipliers}, which show that updating these parameters continuously, based on the support set, results in overfitting and thus provides worse generalization.

\begin{table*}[h!]
\caption{\textbf{Efficient transfer learning performance.} Full numerical performance comparison on the few-shot setting, using ResNet-50 as the backbone. All experiments are run with a fixed configuration, and training is done until full convergence on the support set. Results are averaged across $3$ random seeds. Results for CoOp and PLOT are directly extracted from \cite{chen2023plot}. }
\label{numerical_etl_comparison}
\centering
\resizebox{1\textwidth}{!}{
\begin{tabular}{lccccccccccccc}
\toprule
Method  & Setting & ImageNet & Caltech101 & OxfordPets & StanfordCars & Flowers102 & Food101 & FGVCAAircraft & SUN397 & DTD & EuroSAT & UCF101 & Average \\
\midrule

CoOp$_\text{ IJCV'22}$\cite{zhou2022coop}          &         & 56.99$_{\pm1.03}$ & 87.51$_{\pm1.02}$ & 85.99$_{\pm0.28}$ & 55.81$_{\pm1.67}$ & 67.98$_{\pm1.98}$ & 74.25$_{\pm1.52}$ &  8.59$_{\pm5.79}$  & 60.12$_{\pm0.82}$ & 43.62$_{\pm1.96}$ & 52.12$_{\pm5.46}$ & 62.13$_{\pm1.14}$  & 59.56$_{\pm2.06}$ \\
PLOT$_\text{ ICLR'23}$\cite{chen2023plot}          &         & 59.54$_{\pm0.16}$ & 89.83$_{\pm0.33}$ & 87.49$_{\pm0.57}$ & 56.60$_{\pm0.36}$ & 71.72$_{\pm0.97}$ & 77.74$_{\pm0.47}$ &  17.90$_{\pm0.09}$ & 62.47$_{\pm0.43}$ & 46.55$_{\pm2.62}$ & 54.05$_{\pm5.95}$ & 64.53$_{\pm0.70}$  & 62.59$_{\pm1.13}$ \\
\hdashline\noalign{\vskip 0.5ex}
Zero-Shot$_\text{ ICML'21}$\cite{radford2021learning} & \multirow{9}{*}{1-shot}  & 60.35$_{\pm0.00}$ & 83.81$_{\pm0.00}$ & 82.86$_{\pm0.00}$ & 55.69$_{\pm0.00}$ & 65.94$_{\pm0.00}$ & 74.85$_{\pm0.00}$ & 17.16$_{\pm0.00}$ & 56.80$_{\pm0.00}$ & 42.32$_{\pm0.00}$ & 37.53$_{\pm0.00}$ & 57.47$_{\pm0.00}$  & 57.71$_{\pm0.00}$ \\
Rand. Init LP$_\text{ ICML'21}$\cite{radford2021learning} &  & 17.62$_{\pm0.01}$ & 56.82$_{\pm1.65}$ & 26.61$_{\pm0.50}$ & 18.41$_{\pm0.95}$ & 53.42$_{\pm2.84}$ & 23.92$_{\pm0.86}$ & 12.14$_{\pm0.35}$ & 26.66$_{\pm0.49}$ & 26.69$_{\pm1.22}$ & 40.69$_{\pm4.58}$ & 31.69$_{\pm0.39}$  & 30.42$_{\pm1.26}$ \\
CLIP-Adapter$_\text{ IJCV'23}$\cite{gao2021clip}   &         & 54.74$_{\pm0.19}$ & 86.80$_{\pm0.10}$ & 74.45$_{\pm1.57}$ & 53.07$_{\pm0.37}$ & 71.57$_{\pm0.78}$ & 66.78$_{\pm0.76}$ & 17.01$_{\pm0.06}$ & 59.10$_{\pm0.17}$ & 41.80$_{\pm1.84}$ & 57.92$_{\pm1.68}$ & 59.49$_{\pm0.33}$  & 58.43$_{\pm0.71}$ \\
TIP-Adapter$_\text{ ECCV'22}$\cite{zhang2021tip}   &         & 60.35$_{\pm0.08}$ & 84.44$_{\pm0.35}$ & 83.18$_{\pm1.01}$ & 56.32$_{\pm0.47}$ & 67.45$_{\pm0.16}$ & 74.69$_{\pm0.15}$ & 17.69$_{\pm0.26}$ & 58.41$_{\pm0.04}$ & 44.03$_{\pm0.30}$ & 42.08$_{\pm3.08}$ & 58.79$_{\pm0.06}$  & 58.86$_{\pm0.54}$ \\
TIP-Adapter(f)$_\text{ ECCV'22}$\cite{zhang2021tip}   &      & 60.51$_{\pm0.06}$ & 85.50$_{\pm0.46}$ & 83.90$_{\pm0.92}$ & 56.71$_{\pm0.50}$ & 68.60$_{\pm0.70}$ & 74.76$_{\pm0.15}$ & 18.33$_{\pm0.57}$ & 58.73$_{\pm0.03}$ & 44.64$_{\pm0.29}$ & 51.11$_{\pm2.24}$ & 60.38$_{\pm0.33}$  & 60.29$_{\pm0.57}$ \\
TaskRes(r)$_\text{ CVPR'23}$\cite{yu2023task}  &             & 57.91$_{\pm0.25}$ & 87.99$_{\pm0.11}$ & 77.94$_{\pm2.58}$ & 55.25$_{\pm0.49}$ & 79.62$_{\pm0.45}$ & 70.60$_{\pm0.35}$ & 20.30$_{\pm0.81}$ & 60.99$_{\pm0.19}$ & 46.59$_{\pm1.51}$ & 55.60$_{\pm1.70}$ & 61.32$_{\pm0.75}$  & 61.28$_{\pm0.84}$ \\
TaskRes(e)$_\text{ CVPR'23}$\cite{yu2023task}   &            & 58.21$_{\pm0.19}$ & 88.01$_{\pm0.18}$ & 78.01$_{\pm2.55}$ & 55.36$_{\pm0.48}$ & 79.83$_{\pm0.54}$ & 70.60$_{\pm0.37}$ & 20.50$_{\pm0.82}$ & 61.33$_{\pm0.09}$ & 46.63$_{\pm1.49}$ & 55.75$_{\pm1.67}$ & 61.59$_{\pm0.80}$  & 61.44$_{\pm0.83}$ \\
CrossModal-LP$_\text{ CVPR'23}$\cite{lin2023crossmodal}   &  & 57.40$_{\pm0.11}$ & 88.06$_{\pm0.54}$ & 80.09$_{\pm1.41}$ & 57.43$_{\pm0.45}$ & 78.51$_{\pm0.54}$ & 73.00$_{\pm0.18}$ & 20.72$_{\pm0.29}$ & 61.32$_{\pm0.11}$ & 47.81$_{\pm1.52}$ & 56.85$_{\pm3.36}$ & 63.49$_{\pm0.23}$  & 62.24$_{\pm0.79}$ \\
\rowcolor{Gray}ZS-LP              &                          & 57.91$_{\pm0.25}$ & 87.98$_{\pm0.13}$ & 77.96$_{\pm2.57}$ & 55.24$_{\pm0.46}$ & 79.62$_{\pm0.47}$ & 70.60$_{\pm0.35}$ & 20.30$_{\pm0.81}$ & 61.00$_{\pm0.19}$ & 46.59$_{\pm1.51}$ & 55.57$_{\pm1.71}$ & 61.34$_{\pm0.73}$  & 61.28$_{\pm0.83}$ \\
\rowcolor{Gray}CLAP            &                          & 58.50$_{\pm0.24}$ & 88.38$_{\pm0.25}$ & 83.64$_{\pm1.18}$ & 56.35$_{\pm0.40}$ & 79.90$_{\pm0.46}$ & 73.00$_{\pm0.14}$ & 20.62$_{\pm0.55}$ & 61.15$_{\pm0.18}$ & 47.46$_{\pm1.15}$ & 59.21$_{\pm0.82}$ & 62.48$_{\pm0.99}$  & 62.79$_{\pm0.58}$ \\
\midrule
CoOp$_\text{ IJCV'22}$\cite{zhou2022coop}          &         & 56.40$_{\pm0.87}$ & 87.84$_{\pm1.10}$ & 82.22$_{\pm2.15}$ & 58.41$_{\pm0.43}$ & 77.58$_{\pm1.46}$ & 72.61$_{\pm1.33}$ & 16.52$_{\pm2.38}$ & 59.60$_{\pm0.76}$ & 45.35$_{\pm0.31}$ & 59.00$_{\pm3.48}$ & 64.05$_{\pm0.99}$  & 61.78 $_{\pm1.39}$ \\
PLOT$_\text{ ICLR'23}$\cite{chen2023plot}          &         & 60.64$_{\pm0.06}$ & 90.67$_{\pm0.21}$ & 86.64$_{\pm0.63}$ & 57.52$_{\pm0.71}$ & 81.19$_{\pm0.79}$ & 77.70$_{\pm0.02}$ & 18.94$_{\pm0.44}$ & 61.71$_{\pm0.65}$ & 51.24$_{\pm1.95}$ & 64.21$_{\pm1.90}$ & 66.83$_{\pm0.43}$  & 65.23$_{\pm0.72}$ \\
\hdashline\noalign{\vskip 0.5ex}
Zero-Shot$_\text{ ICML'21}$\cite{radford2021learning}    & \multirow{9}{*}{2-shot}  & 60.35$_{\pm0.00}$ & 83.81$_{\pm0.00}$ & 82.86$_{\pm0.00}$ & 55.69$_{\pm0.00}$ & 65.94$_{\pm0.00}$ & 74.85$_{\pm0.00}$ & 17.16$_{\pm0.00}$ & 56.80$_{\pm0.00}$ & 42.32$_{\pm0.00}$ & 37.53$_{\pm0.00}$ & 57.47$_{\pm0.00}$  & 57.71$_{\pm0.00}$ \\
Rand. Init LP$_\text{ ICML'21}$\cite{radford2021learning} &  & 26.91$_{\pm0.36}$ & 69.29$_{\pm3.92}$ & 38.88$_{\pm2.99}$ & 31.62$_{\pm1.20}$ & 66.38$_{\pm0.52}$ & 37.99$_{\pm1.24}$ & 16.61$_{\pm0.75}$ & 38.97$_{\pm0.75}$ & 36.98$_{\pm0.19}$ & 51.73$_{\pm1.22}$ & 45.15$_{\pm0.39}$  & 41.86$_{\pm1.26}$ \\
CLIP-Adapter$_\text{ IJCV'23}$\cite{gao2021clip}   &         & 54.20$_{\pm0.31}$ & 88.22$_{\pm0.67}$ & 77.03$_{\pm2.53}$ & 58.62$_{\pm0.37}$ & 80.39$_{\pm0.66}$ & 69.43$_{\pm0.57}$ & 20.07$_{\pm0.65}$ & 60.22$_{\pm0.65}$ & 49.51$_{\pm0.21}$ & 63.95$_{\pm1.84}$ & 65.43$_{\pm0.33}$  & 62.46$_{\pm0.71}$ \\
TIP-Adapter$_\text{ ECCV'22}$\cite{zhang2021tip}   &         & 60.18$_{\pm0.15}$ & 85.76$_{\pm0.64}$ & 83.28$_{\pm0.70}$ & 56.97$_{\pm0.06}$ & 68.78$_{\pm0.14}$ & 74.94$_{\pm0.04}$ & 18.70$_{\pm0.18}$ & 60.03$_{\pm0.18}$ & 45.04$_{\pm0.10}$ & 50.07$_{\pm0.30}$ & 59.88$_{\pm0.06}$  & 60.33$_{\pm0.54}$ \\
TIP-Adapter(f)$_\text{ ECCV'22}$\cite{zhang2021tip}   &      & 60.69$_{\pm0.14}$ & 87.45$_{\pm0.36}$ & 84.86$_{\pm0.47}$ & 58.14$_{\pm0.05}$ & 70.51$_{\pm0.05}$ & 75.65$_{\pm0.25}$ & 19.77$_{\pm0.26}$ & 61.26$_{\pm0.26}$ & 48.25$_{\pm0.13}$ & 55.08$_{\pm0.29}$ & 63.24$_{\pm0.33}$  & 62.26$_{\pm0.57}$ \\
TaskRes(r)$_\text{ CVPR'23}$\cite{yu2023task}  &             & 57.86$_{\pm0.05}$ & 89.26$_{\pm0.21}$ & 80.59$_{\pm1.57}$ & 60.69$_{\pm0.41}$ & 84.48$_{\pm0.20}$ & 72.94$_{\pm0.36}$ & 23.16$_{\pm0.36}$ & 62.61$_{\pm0.35}$ & 51.79$_{\pm0.20}$ & 63.06$_{\pm1.51}$ & 67.26$_{\pm0.75}$  & 64.88$_{\pm0.84}$ \\
TaskRes(e)$_\text{ CVPR'23}$\cite{yu2023task}   &            & 58.08$_{\pm0.12}$ & 89.37$_{\pm0.33}$ & 80.80$_{\pm1.54}$ & 61.56$_{\pm0.84}$ & 85.49$_{\pm0.68}$ & 73.06$_{\pm0.49}$ & 23.55$_{\pm0.49}$ & 63.06$_{\pm0.65}$ & 52.17$_{\pm0.39}$ & 63.33$_{\pm1.49}$ & 67.39$_{\pm0.80}$  & 65.26$_{\pm0.83}$ \\
CrossModal-LP$_\text{ CVPR'23}$\cite{lin2023crossmodal}   &  & 49.13$_{\pm0.20}$ & 89.55$_{\pm0.36}$ & 81.72$_{\pm0.72}$ & 61.76$_{\pm0.27}$ & 82.30$_{\pm0.55}$ & 74.31$_{\pm0.32}$ & 22.46$_{\pm0.30}$ & 63.91$_{\pm0.24}$ & 53.09$_{\pm1.44}$ & 62.91$_{\pm1.41}$ & 68.13$_{\pm0.67}$  & 64.48$_{\pm0.59}$ \\
\rowcolor{Gray}ZS-LP              &                          & 57.85$_{\pm0.04}$ & 89.26$_{\pm0.21}$ & 80.56$_{\pm1.58}$ & 60.69$_{\pm0.42}$ & 84.46$_{\pm0.19}$ & 72.94$_{\pm0.35}$ & 23.18$_{\pm0.35}$ & 62.61$_{\pm0.36}$ & 51.79$_{\pm0.20}$ & 63.06$_{\pm1.51}$ & 67.23$_{\pm0.73}$  & 64.88$_{\pm0.83}$ \\
\rowcolor{Gray}CLAP            &                          & 58.50$_{\pm0.24}$ & 89.79$_{\pm0.15}$ & 84.93$_{\pm0.66}$ & 61.40$_{\pm0.38}$ & 84.22$_{\pm0.35}$ & 74.94$_{\pm0.24}$ & 23.21$_{\pm0.24}$ & 63.31$_{\pm0.32}$ & 53.05$_{\pm0.13}$ & 65.63$_{\pm1.15}$ & 67.77$_{\pm0.99}$  & 66.07$_{\pm0.58}$ \\
\midrule
CoOp$_\text{ IJCV'22}$\cite{zhou2022coop}          &         & 58.48$_{\pm0.47}$ & 89.52$_{\pm0.80}$ & 86.65$_{\pm0.97}$ & 62.74$_{\pm0.16}$ & 86.10$_{\pm1.05}$ & 73.49$_{\pm2.03}$ & 20.63$_{\pm2.46}$ & 63.24$_{\pm0.63}$ & 53.94$_{\pm1.37}$ & 68.61$_{\pm3.54}$ & 67.79$_{\pm0.71}$  & 66.47$_{\pm1.29}$ \\
PLOT$_\text{ ICLR'23}$\cite{chen2023plot}          &         & 61.49$_{\pm0.23}$ & 90.80$_{\pm0.20}$ & 88.63$_{\pm0.26}$ & 63.41$_{\pm0.29}$ & 87.82$_{\pm0.20}$ & 77.21$_{\pm0.43}$ & 22.36$_{\pm0.42}$ & 65.09$_{\pm0.43}$ & 56.03$_{\pm0.43}$ & 72.36$_{\pm2.29}$ & 69.60$_{\pm0.67}$  & 68.60$_{\pm0.52}$ \\
\hdashline\noalign{\vskip 0.5ex}
Zero-Shot$_\text{ ICML'21}$\cite{radford2021learning}    & \multirow{9}{*}{4-shot}  & 60.35$_{\pm0.00}$ & 83.81$_{\pm0.00}$ & 82.86$_{\pm0.00}$ & 55.69$_{\pm0.00}$ & 65.94$_{\pm0.00}$ & 74.85$_{\pm0.00}$ & 17.16$_{\pm0.00}$ & 56.80$_{\pm0.00}$ & 42.32$_{\pm0.00}$ & 37.53$_{\pm0.00}$ & 57.47$_{\pm0.00}$  & 57.71$_{\pm0.00}$ \\
Rand. Init LP$_\text{ ICML'21}$\cite{radford2021learning} &  & 36.98$_{\pm0.57}$ & 78.11$_{\pm2.90}$ & 50.00$_{\pm2.03}$ & 44.75$_{\pm0.28}$ & 77.21$_{\pm1.15}$ & 50.10$_{\pm0.96}$ & 21.08$_{\pm0.79}$ & 49.63$_{\pm0.69}$ & 47.97$_{\pm0.40}$ & 58.51$_{\pm4.06}$ & 54.26$_{\pm0.46}$  & 51.69$_{\pm1.30}$ \\
CLIP-Adapter$_\text{ IJCV'23}$\cite{gao2021clip}   &         & 55.66$_{\pm0.31}$ & 90.39$_{\pm0.26}$ & 79.99$_{\pm1.69}$ & 61.04$_{\pm0.59}$ & 85.28$_{\pm0.57}$ & 72.10$_{\pm0.12}$ & 23.03$_{\pm0.08}$ & 62.85$_{\pm0.32}$ & 55.89$_{\pm0.88}$ & 72.49$_{\pm3.30}$ & 69.28$_{\pm0.16}$  & 66.18$_{\pm0.75}$ \\
TIP-Adapter$_\text{ ECCV'22}$\cite{zhang2021tip}   &         & 60.18$_{\pm0.08}$ & 86.98$_{\pm0.40}$ & 82.27$_{\pm1.21}$ & 57.70$_{\pm0.76}$ & 69.81$_{\pm0.62}$ & 74.65$_{\pm0.24}$ & 19.60$_{\pm0.46}$ & 61.42$_{\pm0.41}$ & 47.18$_{\pm0.50}$ & 54.29$_{\pm3.66}$ & 62.26$_{\pm0.26}$  & 61.49$_{\pm0.78}$ \\
TIP-Adapter(f)$_\text{ ECCV'22}$\cite{zhang2021tip}   &      & 61.45$_{\pm0.05}$ & 88.84$_{\pm0.78}$ & 85.51$_{\pm0.57}$ & 61.09$_{\pm0.50}$ & 74.39$_{\pm0.08}$ & 75.25$_{\pm0.20}$ & 21.87$_{\pm0.69}$ & 64.23$_{\pm0.16}$ & 53.45$_{\pm0.27}$ & 66.77$_{\pm3.38}$ & 65.70$_{\pm0.29}$  & 65.32$_{\pm0.63}$ \\
TaskRes(r)$_\text{ CVPR'23}$\cite{yu2023task}  &             & 57.88$_{\pm0.18}$ & 90.37$_{\pm0.34}$ & 82.76$_{\pm1.02}$ & 63.73$_{\pm0.38}$ & 88.44$_{\pm0.65}$ & 74.41$_{\pm0.33}$ & 25.59$_{\pm0.43}$ & 64.70$_{\pm0.38}$ & 57.58$_{\pm0.20}$ & 72.77$_{\pm3.42}$ & 69.58$_{\pm0.25}$  & 67.98$_{\pm0.69}$ \\
TaskRes(e)$_\text{ CVPR'23}$\cite{yu2023task}   &            & 58.02$_{\pm0.26}$ & 90.49$_{\pm0.41}$ & 83.24$_{\pm1.08}$ & 64.69$_{\pm0.53}$ & 89.38$_{\pm0.81}$ & 74.46$_{\pm0.15}$ & 25.89$_{\pm0.47}$ & 64.83$_{\pm0.10}$ & 57.98$_{\pm0.22}$ & 72.95$_{\pm3.35}$ & 69.88$_{\pm0.43}$  & 68.35$_{\pm0.71}$ \\
CrossModal-LP$_\text{ CVPR'23}$\cite{lin2023crossmodal}   &  & 42.15$_{\pm0.21}$ & 90.40$_{\pm0.25}$ & 84.56$_{\pm0.94}$ & 65.18$_{\pm0.14}$ & 85.00$_{\pm0.59}$ & 75.66$_{\pm0.36}$ & 24.44$_{\pm0.41}$ & 66.09$_{\pm0.42}$ & 58.41$_{\pm0.20}$ & 71.72$_{\pm2.42}$ & 69.74$_{\pm0.55}$  & 66.67$_{\pm0.59}$ \\
\rowcolor{Gray}ZS-LP              &                          & 57.88$_{\pm0.18}$ & 90.36$_{\pm0.35}$ & 82.76$_{\pm1.02}$ & 63.73$_{\pm0.37}$ & 88.47$_{\pm0.65}$ & 74.41$_{\pm0.33}$ & 25.57$_{\pm0.45}$ & 64.70$_{\pm0.38}$ & 57.58$_{\pm0.20}$ & 72.78$_{\pm3.44}$ & 69.55$_{\pm0.26}$  & 67.98$_{\pm0.69}$ \\
\rowcolor{Gray}CLAP            &                          & 60.73$_{\pm0.20}$ & 90.62$_{\pm0.46}$ & 86.51$_{\pm0.32}$ & 65.50$_{\pm0.26}$ & 87.66$_{\pm0.85}$ & 75.92$_{\pm0.16}$ & 25.65$_{\pm0.67}$ & 65.99$_{\pm0.31}$ & 58.85$_{\pm0.06}$ & 73.15$_{\pm2.34}$ & 69.88$_{\pm0.26}$  & 69.13$_{\pm0.54}$ \\
\midrule
CoOp$_\text{ IJCV'22}$\cite{zhou2022coop}          &         & 60.39$_{\pm0.57}$ & 90.28$_{\pm0.42}$ & 85.36$_{\pm1.00}$ & 67.64$_{\pm0.06}$ & 91.27$_{\pm0.83}$ & 71.58$_{\pm0.79}$ & 26.63$_{\pm0.86}$ & 65.77$_{\pm0.02}$ & 59.69$_{\pm0.13}$ & 77.08$_{\pm2.42}$ & 72.71$_{\pm0.50}$  & 69.85$_{\pm0.69}$ \\
PLOT$_\text{ ICLR'23}$\cite{chen2023plot}          &         & 61.92$_{\pm0.09}$ & 91.54$_{\pm0.33}$ & 87.39$_{\pm0.74}$ & 67.03$_{\pm0.50}$ & 92.43$_{\pm0.25}$ & 75.31$_{\pm0.30}$ & 26.17$_{\pm0.29}$ & 67.48$_{\pm0.04}$ & 61.70$_{\pm0.35}$ & 78.15$_{\pm0.00}$ & 74.45$_{\pm0.50}$  & 71.23$_{\pm0.51}$ \\
\hdashline\noalign{\vskip 0.5ex}
Zero-Shot$_\text{ ICML'21}$\cite{radford2021learning}    & \multirow{9}{*}{8-shot}  & 60.35$_{\pm0.00}$ & 83.81$_{\pm0.00}$ & 82.86$_{\pm0.00}$ & 55.69$_{\pm0.00}$ & 65.94$_{\pm0.00}$ & 74.85$_{\pm0.00}$ & 17.16$_{\pm0.00}$ & 56.80$_{\pm0.00}$ & 42.32$_{\pm0.00}$ & 37.53$_{\pm0.00}$ & 57.47$_{\pm0.00}$  & 57.71$_{\pm0.00}$ \\
Rand. Init LP$_\text{ ICML'21}$\cite{radford2021learning} &  & 45.06$_{\pm0.42}$ & 84.00$_{\pm2.38}$ & 61.65$_{\pm1.37}$ & 58.06$_{\pm0.23}$ & 87.47$_{\pm0.55}$ & 59.65$_{\pm0.14}$ & 27.99$_{\pm0.49}$ & 57.18$_{\pm0.38}$ & 55.24$_{\pm1.13}$ & 67.34$_{\pm4.76}$ & 65.58$_{\pm0.53}$  & 60.84$_{\pm1.13}$ \\
CLIP-Adapter$_\text{ IJCV'23}$\cite{gao2021clip}   &         & 56.95$_{\pm0.24}$ & 91.33$_{\pm0.24}$ & 83.39$_{\pm0.51}$ & 66.83$_{\pm0.80}$ & 91.93$_{\pm0.40}$ & 72.11$_{\pm0.19}$ & 27.89$_{\pm0.65}$ & 65.09$_{\pm0.21}$ & 61.37$_{\pm1.25}$ & 78.49$_{\pm1.67}$ & 73.23$_{\pm1.46}$  & 69.87$_{\pm0.69}$ \\
TIP-Adapter$_\text{ ECCV'22}$\cite{zhang2021tip}   &         & 59.44$_{\pm0.14}$ & 88.26$_{\pm0.33}$ & 82.27$_{\pm1.21}$ & 57.63$_{\pm0.51}$ & 73.76$_{\pm0.31}$ & 73.87$_{\pm0.34}$ & 19.36$_{\pm0.41}$ & 63.13$_{\pm0.25}$ & 51.52$_{\pm0.27}$ & 62.30$_{\pm1.19}$ & 63.15$_{\pm1.02}$  & 63.15$_{\pm0.54}$ \\
TIP-Adapter(f)$_\text{ ECCV'22}$\cite{zhang2021tip}   &      & 61.80$_{\pm0.05}$ & 90.53$_{\pm0.28}$ & 85.60$_{\pm0.35}$ & 64.42$_{\pm0.06}$ & 84.33$_{\pm0.23}$ & 74.95$_{\pm0.66}$ & 23.79$_{\pm0.48}$ & 66.97$_{\pm0.09}$ & 59.81$_{\pm0.46}$ & 70.34$_{\pm4.31}$ & 69.33$_{\pm1.04}$  & 68.35$_{\pm0.73}$ \\
TaskRes(r)$_\text{ CVPR'23}$\cite{yu2023task}  &             & 59.10$_{\pm0.19}$ & 91.62$_{\pm0.29}$ & 85.77$_{\pm0.39}$ & 69.29$_{\pm0.10}$ & 93.94$_{\pm0.31}$ & 74.52$_{\pm0.29}$ & 29.58$_{\pm0.81}$ & 67.07$_{\pm0.04}$ & 63.18$_{\pm0.99}$ & 78.55$_{\pm3.05}$ & 73.06$_{\pm0.85}$  & 71.43$_{\pm0.66}$ \\
TaskRes(e)$_\text{ CVPR'23}$\cite{yu2023task}   &            & 59.12$_{\pm0.15}$ & 91.94$_{\pm0.24}$ & 85.74$_{\pm0.35}$ & 69.65$_{\pm0.47}$ & 94.29$_{\pm0.37}$ & 74.36$_{\pm0.26}$ & 30.91$_{\pm0.60}$ & 66.31$_{\pm0.25}$ & 63.48$_{\pm0.51}$ & 78.83$_{\pm2.89}$ & 73.64$_{\pm0.35}$  & 71.66$_{\pm0.59}$ \\
CrossModal-LP$_\text{ CVPR'23}$\cite{lin2023crossmodal}   &  & 46.81$_{\pm0.11}$ & 91.76$_{\pm0.06}$ & 86.74$_{\pm0.45}$ & 69.34$_{\pm0.52}$ & 92.87$_{\pm0.24}$ & 76.12$_{\pm0.22}$ & 28.27$_{\pm0.79}$ & 68.20$_{\pm0.29}$ & 62.61$_{\pm0.82}$ & 77.73$_{\pm2.72}$ & 73.55$_{\pm0.53}$  & 70.36$_{\pm0.61}$ \\
\rowcolor{Gray}ZS-LP              &                          & 59.10$_{\pm0.19}$ & 91.62$_{\pm0.29}$ & 85.80$_{\pm0.40}$ & 69.29$_{\pm0.12}$ & 93.94$_{\pm0.29}$ & 74.51$_{\pm0.28}$ & 29.59$_{\pm0.82}$ & 67.08$_{\pm0.04}$ & 63.18$_{\pm0.99}$ & 78.55$_{\pm3.04}$ & 73.05$_{\pm0.88}$  & 71.43$_{\pm0.67}$ \\
\rowcolor{Gray}CLAP            &                          & 62.98$_{\pm0.13}$ & 91.45$_{\pm0.05}$ & 87.75$_{\pm0.40}$ & 70.35$_{\pm0.30}$ & 92.06$_{\pm0.43}$ & 77.42$_{\pm0.31}$ & 28.97$_{\pm0.89}$ & 68.61$_{\pm0.20}$ & 63.24$_{\pm0.65}$ & 76.66$_{\pm2.78}$ & 73.34$_{\pm0.49}$  & 72.08$_{\pm0.60}$ \\
\midrule
CoOp$_\text{ IJCV'22}$\cite{zhou2022coop}          &         & 61.91$_{\pm0.17}$ & 91.99$_{\pm0.31}$ & 87.02$_{\pm0.89}$ & 73.60$_{\pm0.19}$ & 94.49$_{\pm0.40}$ & 74.48$_{\pm0.15}$ & 31.43$_{\pm0.96}$ & 68.36$_{\pm0.66}$ & 62.51$_{\pm0.25}$ & 83.69$_{\pm0.47}$ & 76.90$_{\pm0.50}$  & 73.33$_{\pm0.42}$ \\
PLOT$_\text{ ICLR'23}$\cite{chen2023plot}          &         & 63.01$_{\pm0.13}$ & 92.24$_{\pm0.38}$ & 87.21$_{\pm0.40}$ & 72.80$_{\pm0.75}$ & 94.76$_{\pm0.34}$ & 77.09$_{\pm0.18}$ & 31.49$_{\pm0.89}$ & 69.96$_{\pm0.24}$ & 65.60$_{\pm0.82}$ & 82.23$_{\pm0.91}$ & 77.26$_{\pm0.64}$  & 73.94$_{\pm0.54}$ \\
\hdashline\noalign{\vskip 0.5ex}
Zero-Shot$_\text{ ICML'21}$\cite{radford2021learning}    & \multirow{9}{*}{16-shot} & 60.35$_{\pm0.00}$ & 83.81$_{\pm0.00}$ & 82.86$_{\pm0.00}$ & 55.69$_{\pm0.00}$ & 65.94$_{\pm0.00}$ & 74.85$_{\pm0.00}$ & 17.16$_{\pm0.00}$ & 56.80$_{\pm0.00}$ & 42.32$_{\pm0.00}$ & 37.53$_{\pm0.00}$ & 57.47$_{\pm0.00}$  & 57.71$_{\pm0.00}$ \\
Rand. Init LP$_\text{ ICML'21}$\cite{radford2021learning} &  & 52.24$_{\pm0.10}$ & 87.55$_{\pm2.32}$ & 71.63$_{\pm0.95}$ & 69.20$_{\pm0.62}$ & 92.73$_{\pm0.58}$ & 66.92$_{\pm0.47}$ & 34.63$_{\pm0.89}$ & 63.07$_{\pm0.10}$ & 60.60$_{\pm1.08}$ & 73.38$_{\pm1.38}$ & 70.94$_{\pm0.44}$  & 67.54$_{\pm0.81}$ \\
CLIP-Adapter$_\text{ IJCV'23}$\cite{gao2021clip}   &         & 59.02$_{\pm0.15}$ & 92.28$_{\pm0.21}$ & 84.92$_{\pm0.74}$ & 73.49$_{\pm0.57}$ & 94.56$_{\pm0.30}$ & 73.96$_{\pm0.18}$ & 34.19$_{\pm0.65}$ & 68.14$_{\pm0.20}$ & 65.70$_{\pm0.24}$ & 83.24$_{\pm0.56}$ & 77.30$_{\pm0.37}$  & 73.35$_{\pm0.38}$ \\
TIP-Adapter$_\text{ ECCV'22}$\cite{zhang2021tip}   &         & 57.81$_{\pm0.18}$ & 88.44$_{\pm0.37}$ & 81.09$_{\pm1.89}$ & 58.83$_{\pm0.38}$ & 78.41$_{\pm0.53}$ & 72.96$_{\pm0.42}$ & 21.96$_{\pm0.56}$ & 64.00$_{\pm0.26}$ & 54.79$_{\pm0.66}$ & 67.90$_{\pm2.32}$ & 64.52$_{\pm0.97}$  & 64.61$_{\pm0.78}$ \\
TIP-Adapter(f)$_\text{ ECCV'22}$\cite{zhang2021tip}   &      & 62.27$_{\pm0.13}$ & 91.22$_{\pm0.43}$ & 85.43$_{\pm0.59}$ & 69.56$_{\pm0.36}$ & 91.18$_{\pm0.27}$ & 74.65$_{\pm0.34}$ & 29.32$_{\pm0.64}$ & 68.90$_{\pm0.05}$ & 64.56$_{\pm0.32}$ & 76.55$_{\pm1.44}$ & 71.81$_{\pm0.70}$  & 71.40$_{\pm0.48}$ \\
TaskRes(r)$_\text{ CVPR'23}$\cite{yu2023task}  &             & 61.01$_{\pm0.11}$ & 93.00$_{\pm0.07}$ & 86.28$_{\pm0.35}$ & 75.48$_{\pm0.10}$ & 95.82$_{\pm0.29}$ & 75.86$_{\pm0.04}$ & 34.82$_{\pm0.89}$ & 69.72$_{\pm0.03}$ & 66.45$_{\pm0.33}$ & 83.15$_{\pm0.14}$ & 76.54$_{\pm0.42}$  & 74.38$_{\pm0.25}$ \\
TaskRes(e)$_\text{ CVPR'23}$\cite{yu2023task}   &            & 60.85$_{\pm0.12}$ & 93.09$_{\pm0.17}$ & 86.28$_{\pm0.21}$ & 75.38$_{\pm0.03}$ & 96.14$_{\pm0.37}$ & 75.43$_{\pm0.07}$ & 36.53$_{\pm0.31}$ & 68.43$_{\pm0.11}$ & 65.88$_{\pm0.28}$ & 83.70$_{\pm0.38}$ & 76.96$_{\pm0.07}$  & 74.42$_{\pm0.19}$ \\
CrossModal-LP$_\text{ CVPR'23}$\cite{lin2023crossmodal}   &  & 52.90$_{\pm0.10}$ & 92.77$_{\pm0.04}$ & 87.48$_{\pm0.14}$ & 75.44$_{\pm0.33}$ & 95.20$_{\pm0.23}$ & 77.14$_{\pm0.16}$ & 33.30$_{\pm0.26}$ & 70.56$_{\pm0.12}$ & 66.92$_{\pm0.56}$ & 82.03$_{\pm0.99}$ & 76.40$_{\pm0.28}$  & 73.65$_{\pm0.29}$ \\
\rowcolor{Gray}ZS-LP              &                          & 61.00$_{\pm0.11}$ & 92.98$_{\pm0.09}$ & 86.27$_{\pm0.33}$ & 75.49$_{\pm0.09}$ & 95.82$_{\pm0.29}$ & 75.86$_{\pm0.04}$ & 34.82$_{\pm0.86}$ & 69.72$_{\pm0.04}$ & 66.43$_{\pm0.30}$ & 83.16$_{\pm0.13}$ & 76.54$_{\pm0.42}$  & 74.37$_{\pm0.25}$ \\
\rowcolor{Gray}CLAP            &                          & 65.02$_{\pm0.06}$ & 91.93$_{\pm0.18}$ & 88.51$_{\pm0.16}$ & 75.12$_{\pm0.21}$ & 94.21$_{\pm0.13}$ & 78.55$_{\pm0.07}$ & 33.59$_{\pm0.86}$ & 70.78$_{\pm0.05}$ & 66.41$_{\pm0.74}$ & 80.07$_{\pm0.39}$ & 76.29$_{\pm0.21}$  & 74.57$_{\pm0.28}$ \\
\bottomrule
\end{tabular}
}
\end{table*}

\begin{table*}[h!]
\caption{\textbf{Domain generalization results.} Adapters are adjusted on ImageNet using 16 shots per class, and evaluated at out-of-distribution generalization on 4 ImageNet shifts with multiple CLIP visual backbones. Bold indicates best performance. Relative improvements are obtained for each adapter with respect to no adaptation, \ie, zero-shot prediction.}
\label{results_generalization_supp}
\centering
\scriptsize
\begin{tabular}{lcccccccc}
\toprule
\multicolumn{1}{c}{\multirow{2}{*}{Method}} &
  \multirow{2}{*}{Visual Backbone} &
  \multicolumn{1}{c}{Source} &
  \multicolumn{5}{c}{Target} \\ \cline{4-8} 
\multicolumn{1}{c}{} &
   &
  \multicolumn{1}{c}{Imagenet} & 
  \multicolumn{1}{c}{-V2} &
  \multicolumn{1}{c}{-Sketch} &
  \multicolumn{1}{c}{-A} &
  \multicolumn{1}{c}{-R} &
  \multicolumn{1}{c}{Avg.} \\ 
\midrule
Zero-Shot$_\text{ ICML'21}$\cite{radford2021learning}      & \multirow{7}{*}{ResNet-50}    & 60.35                                                  & 51.49           & 33.33          & \textbf{21.67} & 55.93          & 40.61 \\
Rand. Init LP$_\text{ ICML'21}$\cite{radford2021learning}  &                               & 52.24$_{(-8.11)}$\textcolor{red}{$\downarrow$}         & 41.85           & 15.93          & 10.72          & 29.95          & 24.61$_{(-16.00)}$\textcolor{red}{$\downarrow$} \\
CLIP-Adapter$_\text{ IJCV'23}$\cite{gao2021clip}           &                               & 59.02$_{(-1.33)}$\textcolor{red}{$\downarrow$}         & 48.15           & 14.63          & 15.75          & 46.29          & 31.21$_{(-9.40)}$\textcolor{red}{$\downarrow$} \\
TIP-Adapter$_\text{ ECCV'22}$\cite{zhang2021tip}           &                               & 57.81$_{(-2.54)}$\textcolor{red}{$\downarrow$}         & 50.32           & 33.59          & 21.88          & 56.98          & 40.69$_{(+0.08)}$\textcolor{blue}{$\uparrow$} \\
TIP-Adapter(f)$_\text{ ECCV'22}$\cite{zhang2021tip}        &                               & 62.27$_{(+1.92)}$\textcolor{blue}{$\uparrow$}          & 53.99           & 33.75          & 20.48          & 57.22          & 41.36$_{(+0.75)}$\textcolor{blue}{$\uparrow$} \\
TaskRes(e)$_\text{ CVPR'23}$\cite{yu2023task}              &                               & 60.85$_{(+0.50)}$\textcolor{blue}{$\uparrow$}          & \textbf{56.47}  & 32.80          & 19.90          & 55.93          & 41.28$_{(+0.67)}$\textcolor{blue}{$\uparrow$} \\
\rowcolor{Gray}ZS-LP                                       &                               & 61.00$_{(+0.65)}$\textcolor{blue}{$\uparrow$}          & 51.09           & 27.90          & 16.95          & 50.37          & 36.58$_{(-4.03)}$\textcolor{red}{$\downarrow$} \\
\rowcolor{Gray}CLAP                                        &                               & \textbf{65.02}$_{(+4.67)}$\textcolor{blue}{$\uparrow$} & 56.09           & \textbf{34.55} & 21.52          & \textbf{59.48} & \textbf{42.91}$_{(+2.30)}$\textcolor{blue}{$\uparrow$} \\
\midrule
Zero-Shot$_\text{ ICML'21}$\cite{radford2021learning}      & \multirow{7}{*}{ResNet-101}   & 62.66                                                  & 54.86          & 38.69          & 28.01          & 64.44          & 46.50 \\
Rand. Init LP$_\text{ ICML'21}$\cite{radford2021learning}  &                               & 57.51$_{(-5.15)}$\textcolor{red}{$\downarrow$}         & 44.96          & 22.61          & 16.00          & 40.20          & 30.94$_{(-15.56)}$\textcolor{red}{$\downarrow$} \\
CLIP-Adapter$_\text{ IJCV'23}$\cite{gao2021clip}           &                               & 61.87$_{(-0.79)}$\textcolor{red}{$\downarrow$}         & 52.87          & 32.49          & 21.74          & 54.91          & 40.50$_{(-6.00)}$\textcolor{red}{$\downarrow$} \\
TIP-Adapter$_\text{ ECCV'22}$\cite{zhang2021tip}           &                               & 60.83$_{(-1.83)}$\textcolor{red}{$\downarrow$}         & 53.24          & 38.64          & 28.88          & 65.07          & 46.46$_{(+0.04)}$\textcolor{red}{$\downarrow$} \\
TIP-Adapter(f)$_\text{ ECCV'22}$\cite{zhang2021tip}        &                               & 65.13$_{(+2.47)}$\textcolor{blue}{$\uparrow$}          & 56.48          & 38.64          & 26.48          & 64.57          & 46.54$_{(+0.04)}$\textcolor{blue}{$\uparrow$} \\
TaskRes(e)$_\text{ CVPR'23}$\cite{yu2023task}              &                               & 66.10$_{(+3.44)}$\textcolor{blue}{$\uparrow$}          & 56.56          & 36.76          & 24.75          & 61.52          & 44.90$_{(-1.60)}$\textcolor{red}{$\downarrow$} \\
\rowcolor{Gray}ZS-LP                                       &                               & 63.79$_{(+1.13)}$\textcolor{blue}{$\uparrow$}          & 53.74          & 33.64          & 22.89          & 58.07          & 42.09$_{(-4.41)}$\textcolor{red}{$\downarrow$} \\
\rowcolor{Gray}CLAP                                        &                               & \textbf{67.93}$_{(+5.27)}$\textcolor{blue}{$\uparrow$} & \textbf{58.98} & \textbf{40.68} & \textbf{28.35} & \textbf{67.10} & \textbf{48.78}$_{(+2.28)}$\textcolor{blue}{$\uparrow$} \\
\midrule
Zero-Shot$_\text{ ICML'21}$\cite{radford2021learning}      & \multirow{7}{*}{ViT-B/32}     & 63.74                                                  & 54.81          & 40.84          & 29.64          & 66.03          & 47.83 \\
Rand. Init LP$_\text{ ICML'21}$\cite{radford2021learning}  &                               & 56.87$_{(-6.87)}$\textcolor{red}{$\downarrow$}         & 46.50          & 23.44          & 16.64          & 41.13          & 31.93$_{(-15.90)}$\textcolor{red}{$\downarrow$} \\ 
CLIP-Adapter$_\text{ IJCV'23}$\cite{gao2021clip}           &                               & 62.70$_{(-1.04)}$\textcolor{red}{$\downarrow$}         & 52.93          & 34.27          & 23.58          & 57.58          & 42.09$_{(-5.74)}$\textcolor{red}{$\downarrow$} \\ 
TIP-Adapter$_\text{ ECCV'22}$\cite{zhang2021tip}           &                               & 47.71$_{(-16.03)}$\textcolor{red}{$\downarrow$}        & 39.99          & 23.31          & 20.02          & 44.47          & 31.95$_{(-15.88)}$\textcolor{red}{$\downarrow$} \\ 
TIP-Adapter(f)$_\text{ ECCV'22}$\cite{zhang2021tip}        &                               & 45.65$_{(-18.09)}$\textcolor{red}{$\downarrow$}        & 38.00          & 22.47          & 12.40          & 27.44          & 25.08$_{(-22.75)}$\textcolor{red}{$\downarrow$} \\ 
TaskRes(e)$_\text{ CVPR'23}$\cite{yu2023task}              &                               & 65.18$_{(+1.44)}$\textcolor{blue}{$\uparrow$}          & 55.39          & 36.54          & 25.97          & 61.93          & 44.96$_{(-2.87)}$\textcolor{red}{$\downarrow$} \\ 
\rowcolor{Gray}ZS-LP                                       &                               & 64.02$_{(+0.28)}$\textcolor{blue}{$\uparrow$}          & 53.61          & 34.93          & 24.06          & 60.72          & 43.33$_{(-4.50)}$\textcolor{red}{$\downarrow$} \\ 
\rowcolor{Gray}CLAP                                        &                               & \textbf{68.33}$_{(+4.59)}$\textcolor{blue}{$\uparrow$} & \textbf{58.38} & \textbf{41.27} & \textbf{29.91} & \textbf{68.61} & \textbf{49.54}$_{(+1.71)}$\textcolor{blue}{$\uparrow$} \\ 
\midrule
Zero-Shot$_\text{ ICML'21}$\cite{radford2021learning}      & \multirow{7}{*}{ViT-B/16}     & 68.71                                                   & 60.76           & 46.18         & 47.76          & 73.98          & 57.17 \\
Rand. Init LP$_\text{ ICML'21}$\cite{radford2021learning}  &                               & 62.95$_{(-5.76)}$\textcolor{red}{$\downarrow$}          & 52.48           & 29.22         & 29.40          & 50.54          & 40.41$_{(-16.76)}$\textcolor{red}{$\downarrow$} \\ 
CLIP-Adapter$_\text{ IJCV'23}$\cite{gao2021clip}           &                               & 68.46$_{(-0.25)}$\textcolor{red}{$\downarrow$}          & 59.55           & 39.88         & 38.83          & 64.62          & 50.72$_{(-6.45})$\textcolor{red}{$\downarrow$} \\
TIP-Adapter$_\text{ ECCV'22}$\cite{zhang2021tip}           &                               & 53.81$_{(-14.90)}$\textcolor{red}{$\downarrow$}         & 45.69           & 29.21         & 36.04          & 55.26          & 41.55$_{(-15.62)}$\textcolor{red}{$\downarrow$} \\ 
TIP-Adapter(f)$_\text{ ECCV'22}$\cite{zhang2021tip}        &                               & 51.71$_{(-17.00)}$\textcolor{red}{$\downarrow$}         & 43.07           & 27.13         & 27.04          & 45.07          & 35.58$_{(-21.59)}$\textcolor{red}{$\downarrow$} \\  
TaskRes(e)$_\text{ CVPR'23}$\cite{yu2023task}              &                               & 70.84$_{(+2.13)}$\textcolor{blue}{$\uparrow$}           & 62.15           & 43.76         & 43.91          & 71.59          & 55.35$_{(-1.82)}$\textcolor{red}{$\downarrow$} \\
\rowcolor{Gray}ZS-LP                                       &                               & 69.73$_{(+1.02)}$\textcolor{blue}{$\uparrow$}           & 60.40           & 41.63         & 41.94          & 70.64          & 53.65$_{(-3.52)}$\textcolor{red}{$\downarrow$} \\ 
\rowcolor{Gray}CLAP                                        &                               & \textbf{73.38}$_{(+4.67)}$\textcolor{blue}{$\uparrow$}  & \textbf{65.00}  &\textbf{48.35} & \textbf{49.53} & \textbf{77.26} & \textbf{60.04}$_{(+2.87)}$\textcolor{blue}{$\uparrow$} \\ 
\bottomrule
\end{tabular}
\end{table*}

\begin{table*}[h!]
\caption{\textbf{Exploring the proper constraint value in CLAP.} Full numerical performance for the ablation experiment regarding the initial configuration of the Lagrangian multipliers in the class-adaptive Linear Probing. Results using ResNet-50 as the backbone averaged across $3$ random seeds.}
\label{numerical_ablation_constraint}
\centering
\resizebox{1\textwidth}{!}{
\begin{tabular}{lccccccccccccc}
\toprule
Method  & Setting & ImageNet & Caltech101 & OxfordPets & StanfordCars & Flowers102 & Food101 & FGVCAAircraft & SUN397 & DTD & EuroSAT & UCF101 & Average \\
\midrule
ZS-LP                               & \multirow{5}{*}{1-shot}  & 57.91$_{\pm0.25}$ & 87.98$_{\pm0.13}$ & 77.96$_{\pm2.57}$ & 55.24$_{\pm0.46}$ & 79.62$_{\pm0.47}$ & 70.60$_{\pm0.35}$ & 20.30$_{\pm0.81}$ & 61.00$_{\pm0.19}$ & 46.59$_{\pm1.51}$ & 55.57$_{\pm1.71}$ & 61.34$_{\pm0.73}$ & 61.28$_{\pm0.83}$ \\
CLAP(Constant-w=1)               &                          & 59.74$_{\pm0.18}$ & 88.68$_{\pm0.43}$ & 84.23$_{\pm0.69}$ & 58.39$_{\pm0.29}$ & 75.77$_{\pm0.55}$ & 74.44$_{\pm0.36}$ & 20.64$_{\pm0.09}$ & 61.47$_{\pm0.02}$ & 49.45$_{\pm1.49}$ & 59.21$_{\pm1.82}$ & 64.64$_{\pm0.57}$ & 63.33$_{\pm0.59}$ \\
CLAP(ClassWise   - avgCorrected) &                          & 59.02$_{\pm0.24}$ & 88.44$_{\pm0.20}$ & 84.29$_{\pm0.87}$ & 57.75$_{\pm0.43}$ & 79.36$_{\pm0.52}$ & 73.59$_{\pm0.08}$ & 20.76$_{\pm0.20}$ & 61.19$_{\pm0.08}$ & 48.09$_{\pm1.26}$ & 59.85$_{\pm2.61}$ & 63.02$_{\pm1.09}$ & 63.21$_{\pm0.69}$ \\
CLAP(Constant-w=ZS)              &                          & 58.80$_{\pm0.21}$ & 88.65$_{\pm0.34}$ & 83.65$_{\pm0.90}$ & 56.61$_{\pm0.28}$ & 78.17$_{\pm0.30}$ & 73.82$_{\pm0.35}$ & 20.67$_{\pm0.70}$ & 61.24$_{\pm0.16}$ & 48.96$_{\pm1.49}$ & 59.77$_{\pm1.48}$ & 63.83$_{\pm0.53}$ & 63.11$_{\pm0.61}$ \\
CLAP(ClassWise)                  &                          & 58.50$_{\pm0.24}$ & 88.38$_{\pm0.25}$ & 83.64$_{\pm1.18}$ & 56.35$_{\pm0.40}$ & 79.90$_{\pm0.46}$ & 73.00$_{\pm0.14}$ & 20.62$_{\pm0.55}$ & 61.15$_{\pm0.18}$ & 47.46$_{\pm1.15}$ & 59.21$_{\pm0.82}$ & 62.48$_{\pm0.99}$ & 62.79$_{\pm0.58}$ \\
\midrule
ZS-LP                               & \multirow{5}{*}{2-shot}  & 57.85$_{\pm0.04}$ & 89.26$_{\pm0.21}$ & 80.56$_{\pm1.58}$ & 60.69$_{\pm0.42}$ & 84.46$_{\pm0.19}$ & 72.94$_{\pm0.35}$ & 23.18$_{\pm0.36}$ & 62.61$_{\pm0.20}$ & 51.79$_{\pm1.77}$ & 63.06$_{\pm3.14}$ & 67.23$_{\pm0.54}$ & 64.88$_{\pm0.80}$ \\
CLAP(Constant-w=1)               &                          & 61.29$_{\pm0.07}$ & 89.74$_{\pm0.07}$ & 85.26$_{\pm0.59}$ & 62.02$_{\pm0.40}$ & 77.60$_{\pm0.25}$ & 75.77$_{\pm0.23}$ & 22.29$_{\pm0.49}$ & 64.03$_{\pm0.15}$ & 52.36$_{\pm1.37}$ & 63.72$_{\pm0.88}$ & 67.96$_{\pm0.41}$ & 65.64$_{\pm0.45}$ \\
CLAP(ClassWise   - avgCorrected) &                          & 60.42$_{\pm0.16}$ & 89.70$_{\pm0.13}$ & 85.18$_{\pm0.69}$ & 61.87$_{\pm0.21}$ & 83.37$_{\pm0.32}$ & 75.46$_{\pm0.28}$ & 22.61$_{\pm0.39}$ & 63.67$_{\pm0.09}$ & 53.45$_{\pm0.88}$ & 64.07$_{\pm0.58}$ & 68.27$_{\pm0.67}$ & 66.19$_{\pm0.40}$ \\
CLAP(Constant-w=ZS)              &                          & 59.94$_{\pm0.09}$ & 89.99$_{\pm0.19}$ & 85.04$_{\pm0.45}$ & 61.58$_{\pm0.40}$ & 81.88$_{\pm0.30}$ & 75.37$_{\pm0.24}$ & 23.09$_{\pm0.38}$ & 63.71$_{\pm0.17}$ & 53.96$_{\pm1.64}$ & 65.85$_{\pm1.28}$ & 68.19$_{\pm0.54}$ & 66.24$_{\pm0.52}$ \\
CLAP(ClassWise)                  &                          & 58.50$_{\pm0.24}$ & 89.79$_{\pm0.15}$ & 84.93$_{\pm0.66}$ & 61.40$_{\pm0.38}$ & 84.22$_{\pm0.35}$ & 74.94$_{\pm0.24}$ & 23.21$_{\pm0.32}$ & 63.31$_{\pm0.13}$ & 53.05$_{\pm1.03}$ & 65.63$_{\pm1.49}$ & 67.77$_{\pm0.53}$ & 66.07$_{\pm0.50}$ \\
\midrule
ZS-LP                               & \multirow{5}{*}{4-shot}  & 57.88$_{\pm0.18}$ & 90.36$_{\pm0.35}$ & 82.76$_{\pm1.02}$ & 63.73$_{\pm0.37}$ & 88.47$_{\pm0.65}$ & 74.41$_{\pm0.33}$ & 25.57$_{\pm0.45}$ & 64.70$_{\pm0.38}$ & 57.58$_{\pm0.20}$ & 72.78$_{\pm3.44}$ & 69.55$_{\pm0.26}$ & 67.98$_{\pm0.69}$ \\
CLAP(Constant-w=1)               &                          & 62.51$_{\pm0.11}$ & 90.51$_{\pm0.29}$ & 86.51$_{\pm0.23}$ & 64.84$_{\pm0.41}$ & 79.81$_{\pm0.35}$ & 76.36$_{\pm0.10}$ & 23.30$_{\pm0.48}$ & 65.83$_{\pm0.28}$ & 55.79$_{\pm0.88}$ & 68.37$_{\pm1.37}$ & 68.60$_{\pm0.49}$ & 67.49$_{\pm0.45}$ \\
CLAP(ClassWise   - avgCorrected) &                          & 61.96$_{\pm0.15}$ & 90.41$_{\pm0.33}$ & 86.67$_{\pm0.18}$ & 65.67$_{\pm0.19}$ & 86.36$_{\pm0.86}$ & 76.27$_{\pm0.17}$ & 24.67$_{\pm0.42}$ & 66.12$_{\pm0.32}$ & 57.98$_{\pm0.46}$ & 69.10$_{\pm1.36}$ & 69.56$_{\pm0.44}$ & 68.62$_{\pm0.44}$ \\
CLAP(Constant-w=ZS)              &                          & 61.35$_{\pm0.21}$ & 90.62$_{\pm0.33}$ & 86.31$_{\pm0.25}$ & 65.61$_{\pm0.39}$ & 85.05$_{\pm0.75}$ & 76.04$_{\pm0.17}$ & 25.46$_{\pm0.54}$ & 66.38$_{\pm0.29}$ & 58.51$_{\pm0.51}$ & 73.37$_{\pm2.25}$ & 69.71$_{\pm0.60}$ & 68.95$_{\pm0.57}$ \\
CLAP(ClassWise)                  &                          & 60.73$_{\pm0.20}$ & 90.62$_{\pm0.46}$ & 86.51$_{\pm0.32}$ & 65.50$_{\pm0.26}$ & 87.66$_{\pm0.85}$ & 75.92$_{\pm0.16}$ & 25.65$_{\pm0.67}$ & 65.99$_{\pm0.31}$ & 58.85$_{\pm0.06}$ & 73.15$_{\pm2.34}$ & 69.88$_{\pm0.26}$ & 69.13$_{\pm0.54}$ \\
\midrule
ZS-LP                               & \multirow{5}{*}{8-shot}  & 59.10$_{\pm0.19}$ & 91.62$_{\pm0.29}$ & 85.80$_{\pm0.40}$ & 69.29$_{\pm0.12}$ & 93.94$_{\pm0.29}$ & 74.51$_{\pm0.28}$ & 29.59$_{\pm0.82}$ & 67.08$_{\pm0.04}$ & 63.18$_{\pm0.99}$ & 78.55$_{\pm3.04}$ & 73.05$_{\pm0.88}$ & 71.43$_{\pm0.67}$ \\
CLAP(Constant-w=1)               &                          & 63.83$_{\pm0.07}$ & 90.87$_{\pm0.18}$ & 87.15$_{\pm0.39}$ & 66.95$_{\pm0.24}$ & 81.65$_{\pm0.14}$ & 77.42$_{\pm0.22}$ & 23.60$_{\pm0.24}$ & 67.24$_{\pm0.31}$ & 58.18$_{\pm0.36}$ & 69.06$_{\pm1.41}$ & 70.42$_{\pm0.27}$ & 68.76$_{\pm0.35}$ \\
CLAP(ClassWise   - avgCorrected) &                          & 63.69$_{\pm0.12}$ & 91.01$_{\pm0.05}$ & 87.50$_{\pm0.44}$ & 68.95$_{\pm0.08}$ & 90.26$_{\pm0.13}$ & 77.49$_{\pm0.29}$ & 25.70$_{\pm0.55}$ & 67.93$_{\pm0.17}$ & 61.21$_{\pm0.35}$ & 70.02$_{\pm1.44}$ & 72.01$_{\pm0.26}$ & 70.52$_{\pm0.35}$ \\
CLAP(Constant-w=ZS)              &                          & 63.41$_{\pm0.11}$ & 91.21$_{\pm0.07}$ & 87.49$_{\pm0.41}$ & 69.99$_{\pm0.16}$ & 88.20$_{\pm0.19}$ & 77.46$_{\pm0.26}$ & 29.20$_{\pm0.56}$ & 68.66$_{\pm0.24}$ & 62.79$_{\pm0.79}$ & 76.51$_{\pm2.80}$ & 72.79$_{\pm0.68}$ & 71.61$_{\pm0.57}$ \\
CLAP(ClassWise)                  &                          & 62.98$_{\pm0.13}$ & 91.45$_{\pm0.05}$ & 87.75$_{\pm0.40}$ & 70.35$_{\pm0.30}$ & 92.06$_{\pm0.43}$ & 77.42$_{\pm0.31}$ & 28.97$_{\pm0.89}$ & 68.61$_{\pm0.20}$ & 63.24$_{\pm0.65}$ & 76.66$_{\pm2.78}$ & 73.34$_{\pm0.49}$ & 72.08$_{\pm0.60}$ \\
\midrule
ZS-LP                               & \multirow{5}{*}{16-shot} & 61.00$_{\pm0.11}$ & 92.98$_{\pm0.09}$ & 86.27$_{\pm0.33}$ & 75.49$_{\pm0.09}$ & 95.82$_{\pm0.29}$ & 75.86$_{\pm0.04}$ & 34.82$_{\pm0.86}$ & 69.72$_{\pm0.04}$ & 66.43$_{\pm0.30}$ & 83.16$_{\pm0.13}$ & 76.54$_{\pm0.42}$ & 74.37$_{\pm0.25}$ \\
CLAP(Constant-w=1)               &                          & 64.76$_{\pm0.03}$ & 91.18$_{\pm0.16}$ & 87.64$_{\pm0.11}$ & 68.97$_{\pm0.31}$ & 82.45$_{\pm0.17}$ & 78.06$_{\pm0.08}$ & 24.43$_{\pm0.40}$ & 68.13$_{\pm0.04}$ & 59.06$_{\pm0.46}$ & 71.46$_{\pm0.68}$ & 71.00$_{\pm0.15}$ & 69.74$_{\pm0.24}$ \\
CLAP(ClassWise   - avgCorrected) &                          & 65.02$_{\pm0.04}$ & 91.51$_{\pm0.20}$ & 88.11$_{\pm0.26}$ & 71.55$_{\pm0.19}$ & 92.08$_{\pm0.03}$ & 78.28$_{\pm0.07}$ & 27.80$_{\pm0.18}$ & 69.32$_{\pm0.13}$ & 63.20$_{\pm0.63}$ & 72.56$_{\pm0.16}$ & 74.20$_{\pm0.23}$ & 72.15$_{\pm0.19}$ \\
CLAP(Constant-w=ZS)              &                          & 65.29$_{\pm0.04}$ & 91.81$_{\pm0.20}$ & 88.07$_{\pm0.05}$ & 73.99$_{\pm0.27}$ & 89.61$_{\pm0.15}$ & 78.46$_{\pm0.09}$ & 33.50$_{\pm0.77}$ & 70.46$_{\pm0.11}$ & 65.48$_{\pm0.52}$ & 79.95$_{\pm0.22}$ & 74.61$_{\pm0.39}$ & 73.75$_{\pm0.26}$ \\
CLAP(ClassWise)                  &                          & 65.02$_{\pm0.06}$ & 91.93$_{\pm0.18}$ & 88.51$_{\pm0.16}$ & 75.12$_{\pm0.21}$ & 94.21$_{\pm0.13}$ & 78.55$_{\pm0.07}$ & 33.59$_{\pm0.86}$ & 70.78$_{\pm0.05}$ & 66.41$_{\pm0.74}$ & 80.07$_{\pm0.38}$ & 76.07$_{\pm0.21}$ & 74.57$_{\pm0.28}$ \\
\bottomrule
\end{tabular}
}
\end{table*}

\end{document}